\documentclass{article} 
\usepackage{iclr2020_conference,times}

\usepackage{hyperref}
\usepackage{url}
\usepackage{booktabs} 
\usepackage{amsfonts} 
\usepackage{nicefrac} 
\usepackage{microtype} 

\usepackage{wrapfig,booktabs}

\usepackage{wrapfig} 
\usepackage{graphicx}

\usepackage{amsmath}
\usepackage{amssymb}
\usepackage{algorithm} 
\usepackage{algorithmic} 
\usepackage{enumerate} 
\usepackage{multirow} 
\usepackage{amsmath}

\usepackage{subfigure}

\usepackage{helvet} 
\usepackage{courier} 
\usepackage{color}

\title{Variational Hetero-Encoder Randomized\\GANs for Joint Image-Text Modeling}
\iclrfinalcopy

\author{Hao Zhang,~~Bo Chen\thanks{Corresponding author},~~Long Tian,~~Zhengjue Wang\\
National Laboratory of Radar Signal Processing\\
Xidian University, Xi’an, China\\
\texttt{zhanghao\_xidian@163.com}~~~~\texttt{bchen@mail.xidian.edu.cn}\\
\texttt{tianlong\_xidian@163.com}~~~~\texttt{zhengjuewang@163.com} \\
\AND
Mingyuan Zhou\\
McCombs School of Business\\
The University of Texas at Austin, Austin, TX 78712, USA\\
\texttt{mingyuan.zhou@mccombs.utexas.edu}
}

\newcommand{\Wmat}[0]{\ensuremath{{\bf W}} }

\newcommand{\bv}[0]{\ensuremath{\boldsymbol{b}} }

\newcommand{\dv}[0]{\ensuremath{\boldsymbol{d}} }
\newcommand{\ev}[0]{\ensuremath{\boldsymbol{e}} }

\newcommand{\gv}[0]{\ensuremath{\boldsymbol{g}} }

\newcommand{\kv}[0]{\ensuremath{\boldsymbol{k}} }

\newcommand{\rv}[0]{\ensuremath{\boldsymbol{r}} }
\newcommand{\sv}[0]{\ensuremath{\boldsymbol{s}} }
\newcommand{\tv}[0]{\ensuremath{\boldsymbol{t}} }

\newcommand{\xv}[0]{\ensuremath{\boldsymbol{x}} }

\newcommand{\zv}[0]{\ensuremath{\boldsymbol{z}} }

\newcommand{\Phimat}[0]{\ensuremath{\boldsymbol{\Phi}}}

\newcommand{\Omegamat}[0]{\ensuremath{\boldsymbol{\Omega}}}

\newcommand{\thetav}[0]{\ensuremath{\boldsymbol{\theta}} }

\newcommand{\lambdav}[0]{\ensuremath{\boldsymbol{\lambda}} }

\newcommand{\phiv}[0]{\ensuremath{\boldsymbol{\phi}} }

\newcommand{\varepsilonv}[0]{\ensuremath{\boldsymbol{\varepsilon}} }

\newcommand{\mc}{\multicolumn}

\newcommand{\KL}{\mbox{KL}}

\newcommand{\given}{\,|\,}
\newcommand{\E}{\mathbb{E}}

\makeatletter
\newcommand\figcaption{\def\@captype{figure}\caption}
\newcommand\tabcaption{\def\@captype{table}\caption}
\makeatother

\begin{document}

\maketitle

\begin{abstract}
For bidirectional joint image-text modeling, we develop variational hetero-encoder (VHE) randomized generative adversarial network (GAN), a versatile deep generative model that integrates a probabilistic text decoder, probabilistic image encoder, and GAN into a coherent end-to-end multi-modality learning framework. VHE randomized GAN (VHE-GAN) encodes an image to decode its associated text, and feeds the variational posterior as the source of randomness into the GAN image generator. We plug three off-the-shelf modules, including a deep topic model, a ladder-structured image encoder, and StackGAN++, into VHE-GAN, which already achieves competitive performance. This further motivates the development of VHE-raster-scan-GAN that generates photo-realistic images in not only a multi-scale low-to-high-resolution manner, but also a hierarchical-semantic coarse-to-fine fashion. By capturing and relating hierarchical semantic and visual concepts with end-to-end training, VHE-raster-scan-GAN achieves state-of-the-art performance in a wide variety of image-text multi-modality learning and generation tasks. 
\end{abstract}

\section{Introduction}

Images and texts commonly occur together in the real world. 
There exists a wide variety of deep neural network based unidirectional methods that model images (texts) given texts (images) 
\citep{gomez2017self-supervised,kiros2012deep,reed2016generative,xu2018attngan,Zhang2016StackGAN}. 
There also exist probabilistic graphic model based bidirectional methods \citep{srivastava2012multimodalDBN,srivastava2012multimodal,wang2018MPGBN} that capture the joint distribution of images and texts. 
These bidirectional methods, however, often make restrictive parametric assumptions that limit their image generation ability.
Exploiting recent progress on deep probabilistic models and variational inference \citep{kingma2014stochastic,GBN,Zhang2018WHAI,goodfellow2014generative,Zhang2017StackGAN++}, we propose an end-to-end learning framework to construct multi-modality deep generative models that can not only generate vivid image-text pairs, but also achieve state-of-the-art results on various unidirectional tasks \citep{srivastava2012multimodalDBN,srivastava2012multimodal,wang2018MPGBN,gomez2017self-supervised,xu2018attngan,Zhang2016StackGAN,Zhang2017StackGAN++,verma2018generalized,zhang2018photographic}, such as generating photo-realistic images given texts and performing
text-based zero-shot learning.

To extract and relate semantic and visual concepts, we first introduce variational hetero-encoder (VHE) that encodes an image to decode its textual description ($e.g.$, tags, sentences, binary attributes, and long documents), where the probabilistic encoder and decoder are jointly optimized using variational inference \citep{blei2017variational,hoffman2013stochastic,kingma2014stochastic,rezende2014stochastic}.
The latent representation of VHE can be sampled from either the variational posterior provided by the image encoder given an image input, or the posterior of the text decoder via MCMC given a text input. 
VHE by construction has the ability to generate texts given images. To further enhance its text generation performance and allow synthesizing photo-realistic images given an image, text, or random noise, we feed the variational posterior of VHE in lieu of random noise as the source of randomness into the image generator of a generative adversarial network (GAN) \citep{goodfellow2014generative}. We refer to this new modeling framework as VHE randomized GAN (VHE-GAN).

Off-the-shelf text decoders, image encoders, and GANs can be directly plugged into the VHE-GAN framework for end-to-end multi-modality learning. To begin with, as shown in Figs.~\ref{Fig: VHE} and \ref{Fig: StackGAN++}, we construct VHE-StackGAN++ by using the Poisson gamma belief network (PGBN) \citep{GBN} as the VHE text decoder, using the Weibull upward-downward variational encoder \citep{Zhang2018WHAI} as the VHE image encoder, and feeding the concatenation of the multi-stochastic-layer latent representation of the VHE as the source of randomness into the image generator of StackGAN++ \citep{Zhang2017StackGAN++}.
While VHE-StackGAN++ already achieves very attractive performance, we find that its performance can be clearly boosted by better exploiting the multi-stochastic-layer semantically meaningful hierarchical latent structure of the PGBN text decoder.
To this end, as shown in Figs.~\ref{Fig: VHE} and \ref{Fig: raster-scan-GAN}, we develop VHE-raster-scan-GAN to perform image generation in not only a multi-scale low-to-high-resolution manner in each layer, as done by StackGAN++, but also a hierarchical-semantic coarse-to-fine fashion across layers,
a unique feature distinguishing it from existing methods. Consequently, not only can VHE-raster-scan-GAN generate vivid high-resolution images with better details, but also build interpretable hierarchical semantic-visual relationships between the generated images and texts.

Our main contributions include: 1) VHE-GAN that provides a plug-and-play framework to integrate off-the-shelf probabilistic decoders, variational encoders, and GANs for end-to-end bidirectional multi-modality learning; the shared latent space can be inferred either by image encoder $q(\zv\given \xv)$, if given images, or by Gibbs sampling from the conditional posterior of text decoder $p(\tv\given \zv)$, if given texts; 2) VHE-raster-scan-GAN that captures and relates hierarchical semantic and visual concepts to achieve state-of-the-art results in {various unidirectional and bidirectional} 
image-text modeling tasks.
\vspace{-1mm}

\section{Variational hetero-encoder randomized GANs}
\vspace{-0.5mm}

VAEs and GANs are two distinct types of deep generative models.
Consisting of a generator (decoder) $p(\xv \given \zv)$, a prior  $p(\zv)$, and an inference network (encoder) $q(\zv\given \xv)$ that is used to approximate the  posterior  $p(\zv\given \xv)$, VAEs \citep{kingma2014stochastic,rezende2014stochastic} are optimized by maximizing the evidence lower bound (ELBO) as
\begin{equation}\label{Eq: VAE_ELBO}
    \mbox{ELBO}= \E_{\xv\sim p_{\text{data}}(\xv)}[ \mathcal{L}(\xv)],
    ~~~\mathcal{L}(\xv) := \mathbb{E}_{\zv\sim q(\zv\given \xv)} \left[ \ln p(\xv\given \zv) \right] - \KL \left[ q(\zv\given \xv)||p(\zv) \right], \vspace{-0.5mm}
\end{equation}
where $p_{\text{data}}(\xv)=\sum_{i=1}^N \frac{1}{N} \delta_{\xv_i}$ represents the empirical data distribution. 
Distinct from VAEs that make parametric assumptions on data distribution and perform posterior inference, 
GANs in general use implicit data distribution and do not provide meaningful latent representations \citep{goodfellow2014generative};
they learn both a generator $G$ and a discriminator $D$ by optimizing a mini-max objective as
\vspace{-1.5mm}
\begin{equation}\label{Eq:Gan}\textstyle
    \min_{G} \max_{D}
    \{
    \mathbb{E}_{\xv \sim p_{\text{data}}(\xv)} \left[ \ln D(\xv) \right] + \mathbb{E}_{\zv \sim p(\zv)} \left[ \ln(1- D(G(\zv))) \right]
    \}, \vspace{-1mm}
\end{equation}
where $p(\zv)$ is a random noise distribution that acts as the source of randomness for data generation.
\vspace{-0.5mm}

\subsection{VHE-GAN objective function for end-to-end multi-modality learning}

Below we show how to construct VHE-GAN to jointly model images $\xv$ and their associated texts $\tv$, capturing and relating hierarchical semantic and visual concepts.
First, we modify the usual VAE into VHE, optimizing a lower bound of the text log-marginal-likelihood $\E_{\tv\sim p_{\text{data}}(\tv)}[\ln p(\tv)]$ as
\begin{equation}\label{Eq: VHE_ELBO}
\mbox{ELBO}_{\text{vhe}}= \E_{%
	p_{\text{data}}(\tv,\xv)}
[\mathcal{L}_{\text{vhe}}(\tv,\xv)],~~\mathcal{L}_{\text{vhe}}(\tv,\xv):=
\mathbb{E}_{\zv\sim q(\zv\given \xv)} \left[ \ln p(\tv\given \zv) \right] - \KL \left[ q(\zv\given \xv)||p(\zv) \right],\!
\end{equation}
where
$p(\tv\given \zv)$ is the text decoder, $p(\zv)$ is the prior, $p(\tv)=\E_{\zv\sim p(\zv)}[p(\tv\given \zv)]$, and
$\mathcal{L}_{\text{vhe}}(\tv,\xv) \le \ln \E_{\zv\sim q(\zv\given\xv)} [\frac{p(\tv\given \zv) p(\zv)}{ q(\zv\given \xv)}] =%
\ln p(\tv)$. Second, the image encoder $q(\zv\given \xv)$, which encodes image $\xv$ into its latent representation $\zv$, is used to approximate the posterior $p(\zv\given \tv)=p(\tv\given \zv)p(\zv)/p(\tv)$.
Third, variational posterior $q(\zv\given \xv)$ in lieu of random noise $p(\zv)$ %
is fed as the source of randomness into the GAN image generator.
Combing these three steps, with the parameters of the image encoder $q(\zv\given \xv)$, text decoder $p(\tv\given \zv)$, and GAN generator denoted by $E$, $G_{\text{vae}}$, and $G_{\text{gan}}$, respectively, we express the objective function of VHE-GAN for joint image-text end-to-end learning as
\begin{align}\label{Eq:vrAEGan}\textstyle
&~~~~~~~~~~~~~~~~~~~~~~~~~~~~~~~~~~~~~~~~~~~~
\min_{E, G_{\text{vae}},G_{\text{gan}}} \max_{D} 
\E_{p_{\text{data}}(\tv,\xv)} [\mathcal{L}(\tv,\xv)],\notag\\
&%
{\mathcal{L}(\tv,\xv):= %
	\ln D(\xv)+
	\KL \left[ q(\zv\given \xv)||p(\zv) \right]+ \E_{\zv\sim q(\zv\given \xv)}\big[ \ln (1-D(G_{\text{gan}}(\zv)) ) - \ln p(\tv\given\zv)\big].}
\end{align}
Note the objective function in \eqref{Eq:vrAEGan} implies a data-triple-reuse training strategy, which uses the same data mini-batch in each stochastic gradient update iteration to jointly train the VHE, GAN discriminator, and GAN generator; see a related objective function, shown in \eqref{Eq:vrAEGan_naive} of Appendix A, that is resulted from naively combining the VHE and GAN training objectives.
In VHE-GAN, the optimization of the encoder parameter $E$ is related to not only the VHE's ELBO, but also the GAN mini-max objective function, forcing the variational posterior $q(\zv\given \xv)$ to serve as a bridge between VHE and GAN, allowing them to help each other.
Although there are some models \citep{mescheder2017adversarial, makhzani2015adversarial, Tolstikhin2017WVAE, dumoulin2017adversarially, donahue2017adversarial, che2017mode,srivastava2017veegan,grover2018flow-gan, larsen2016autoencoding, huang2018introvae} combining VAEs and GANs in various ways, they focus on single-modality tasks while the VHE-GAN on two different modalities. 
In Appendix A, we analyze the properties of the VHE-GAN objective function and discuss related works.
Below we develop two different VHE-GANs, one integrates off-the-shelf modules, while the other introduces new interpretable hierarchical latent structure. 
\begin{figure}
	\centering
	\subfigure[]{\includegraphics[scale=0.124]{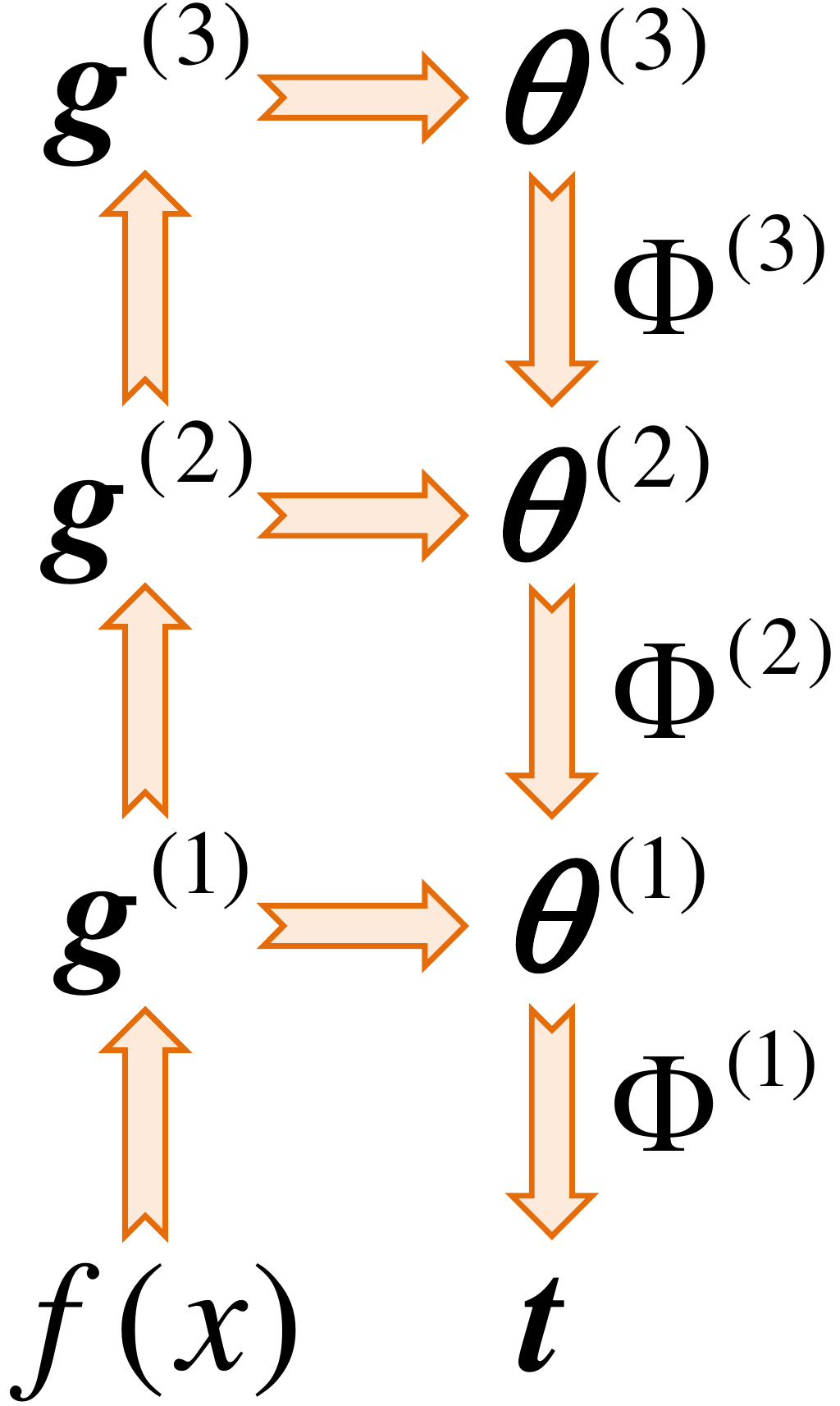}\label{Fig: VHE}}
	\quad	
	\subfigure[]{\includegraphics[scale=0.124]{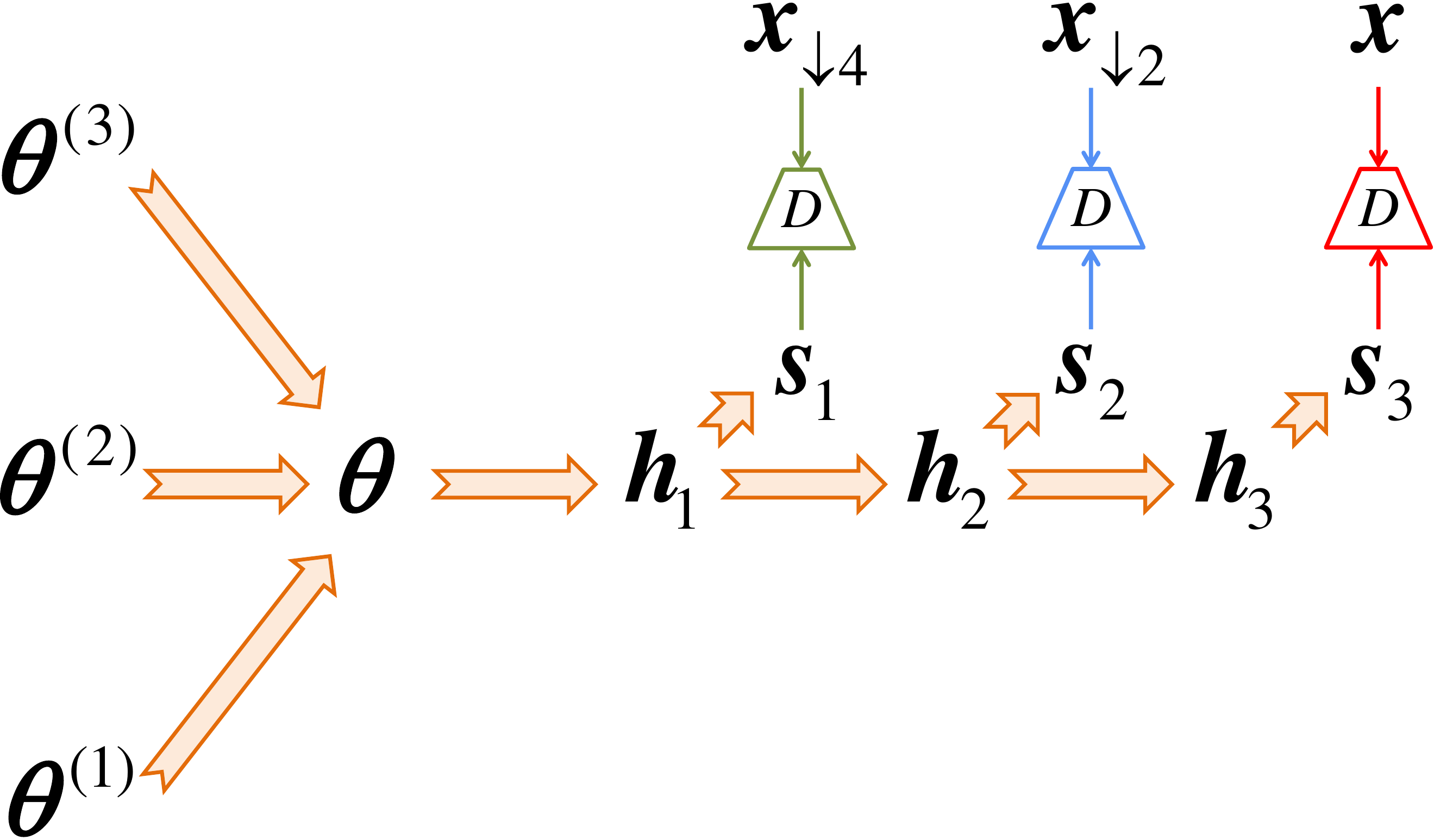}\label{Fig: StackGAN++}}	
	\quad
	\subfigure[]{\includegraphics[scale=0.124]{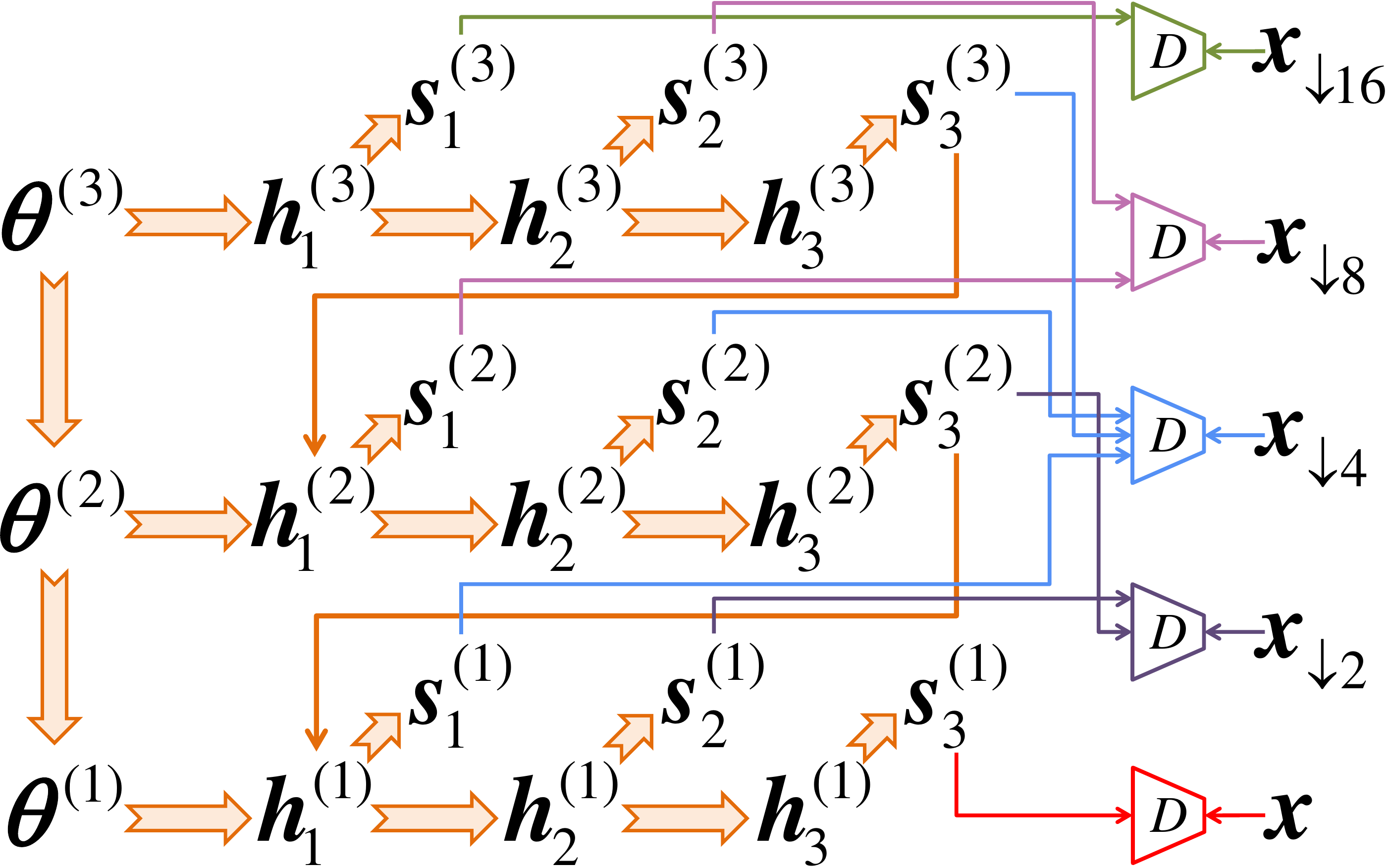}\label{Fig: raster-scan-GAN}}
	\quad
	\subfigure[]{\includegraphics[scale=0.124]{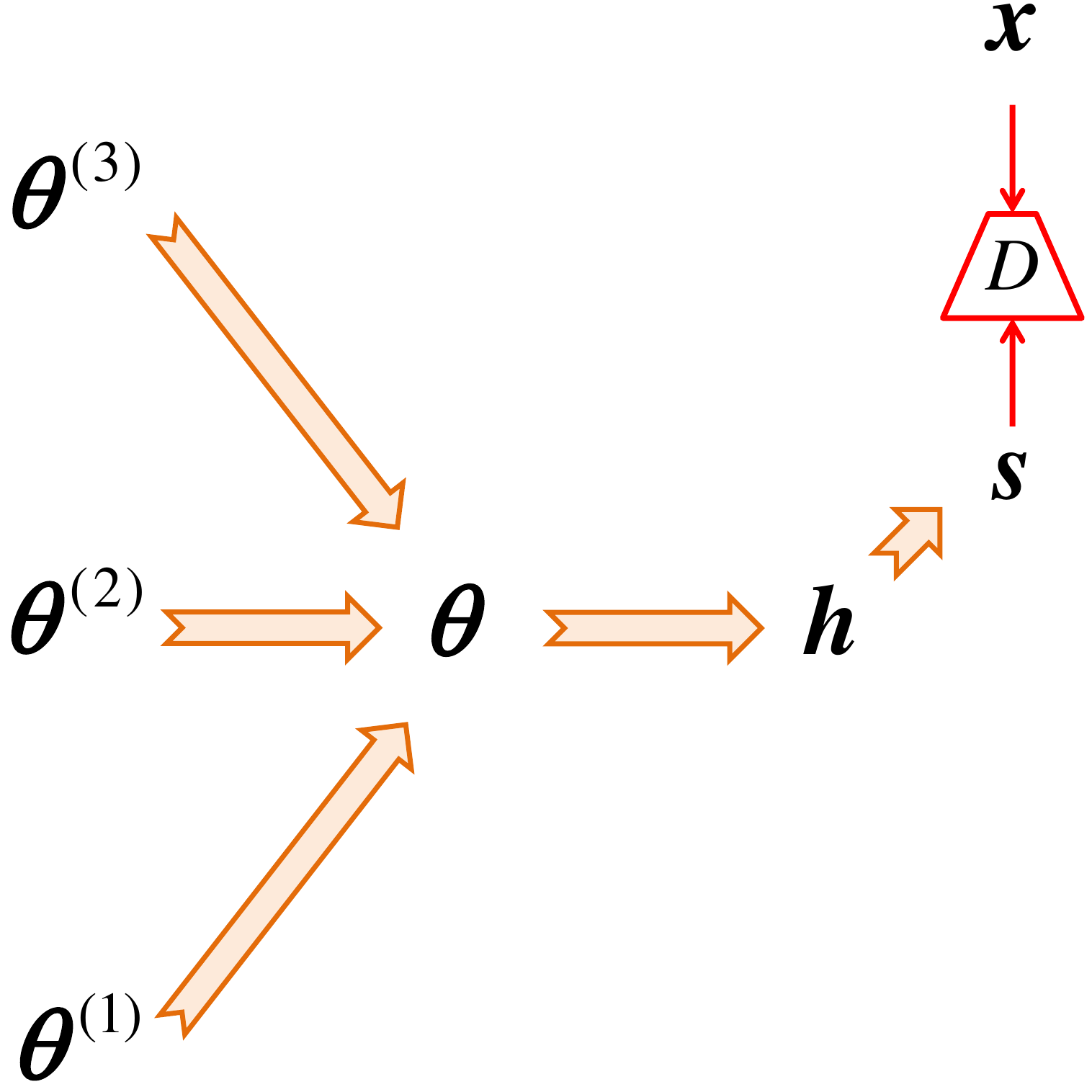}\label{Fig: GAN}}
	\vspace{-4mm}
	\quad
	\subfigure[]{\includegraphics[scale=0.124]{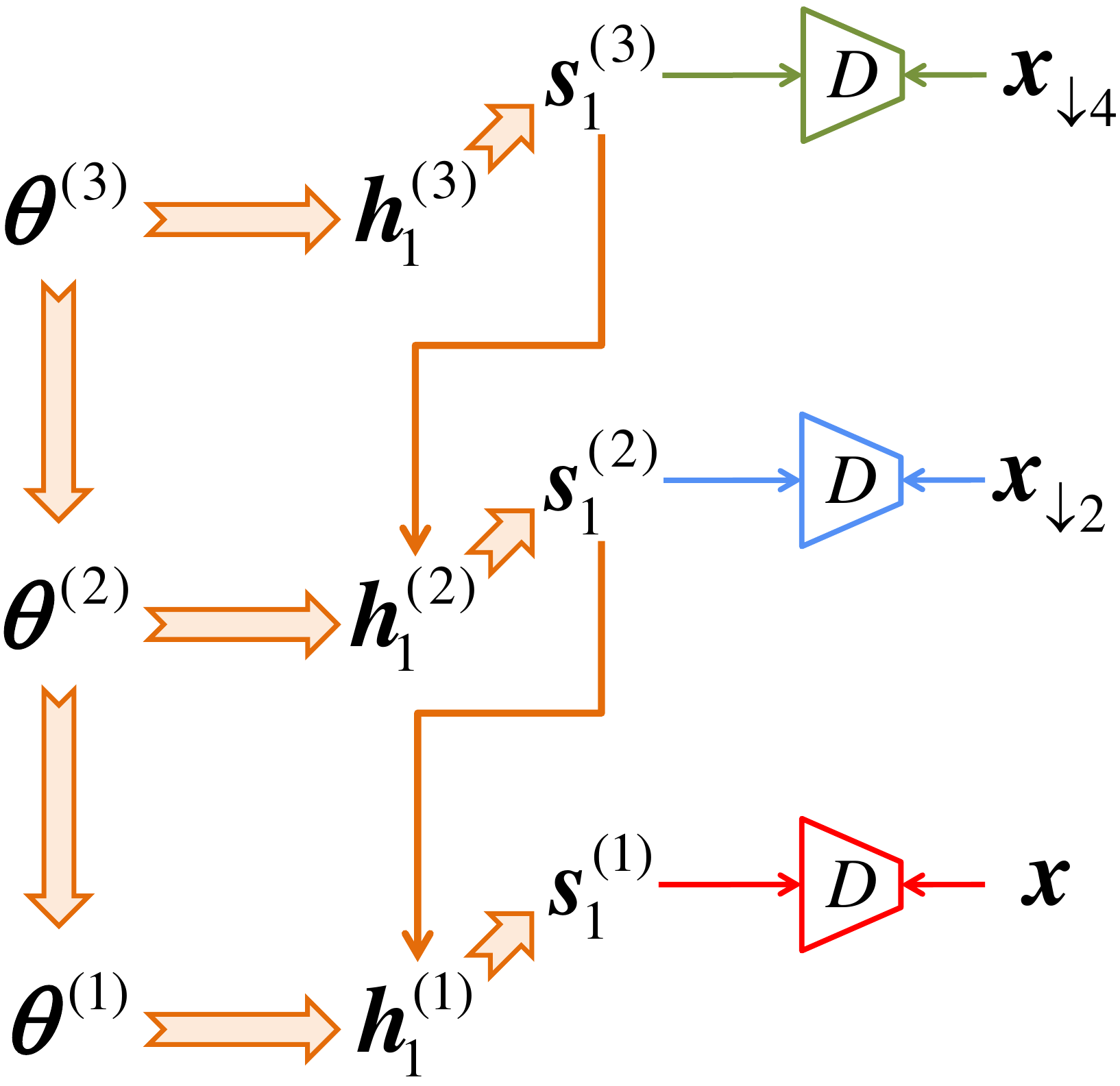}\label{Fig: raster-GAN}}
	\caption{\small
		Illustration of (a) VHE, (b) StackGAN++, (c) raster-scan-GAN, {(d) vanilla-GAN, and (e) simple-raster-scan-GAN}. VHE-raster-scan-GAN consists of (a) and (c).		
		$\xv_{\downarrow d}$ is down-sampled from $\xv$ with scaling factor~$d$.
    	{VHE-StackGAN++,  consisting of (a) and (b), 
	VHE-vanilla-GAN, consisting of (a) and (d), and VHE-simple-raster-scan-GAN, consisting of (a) and (e), are all used 
	for ablation studies.}} 
	
	\vspace{-5mm}
	\label{Fig: VHEGAN}
\end{figure}

\subsection{VHE-StackGAN++ with off-the-shelf modules}\label{sec:2.2}

As shown in Figs.~\ref{Fig: VHE} and \ref{Fig: StackGAN++}, we first construct VHE-StackGAN++ by plugging into VHE-GAN three off-the-shelf modules, including a deep topic model \citep{GBN}, a ladder-structured encoder \citep{Zhang2018WHAI}, and StackGAN++ \citep{Zhang2017StackGAN++}.
For text analysis, both sequence models and topic models are widely used. Sequence models \citep{bengio2003a} often represent each document as a sequence of word embedding vectors, capturing local dependency structures with some type of recurrent neural networks (RNNs), such as long short-term memory (LSTM) \citep{hochreiter1997long}. 
Topic models such as latent Dirichlet allocation (LDA) \citep{blei2003latent} often represent each document as a bag of words (BoW), capturing global word cooccurrence patterns into latent topics.
Suitable for capturing local dependency structure, existing sequence models often have difficulty in capturing long-range word dependencies and hence macro-level information, such as global word cooccurrence patterns ($i.e.$, topics), especially for long documents. 
By contrast, while topic models ignore word order, they are very effective in capturing
latent topics, which are often directly related to macro-level visual information \citep{gomez2017self-supervised,dieng2017topicrnn,lau2017topically}. 
Moreover, topic models can be applied to not only sequential texts, such as few sentences \citep{wang2009multi-document,jin2015aligning} and long documents \citep{GBN}, but also non-sequential ones, such as textual tags \citep{srivastava2012multimodal,srivastava2014multimodal,wang2018MPGBN} and binary attributes \citep{elhoseiny2017link,zhu2018a}. 
For this reason, for the VHE text decoder, we choose PGBN \citep{GBN}, a state-of-the-art topic model that can also be represented as a multi-stochastic-layer deep generalization of LDA \citep{cong2017deep}. We complete VHE-StackGAN++ by choosing the Weibull upward-downward variational encoder \citep{Zhang2018WHAI} as the VHE image encoder, and feeding the concatenation of all the hidden layers of PGBN as the source of randomness to the image generator of StackGAN++ \citep{Zhang2017StackGAN++}. 

As in Fig.~\ref{Fig: VHEGAN}, we use a VHE that encodes an image into a deterministic-upward--stochastic-downward ladder-structured latent representation, which is used to decode the corresponding text. 
Specifically, we represent each document as a BoW high-dimensional sparse count vector $\tv_n\in \mathbb{Z}^{K_0}$, where $\mathbb{Z}=\{0,1,\cdots \}$ and $K_0$ is the vocabulary size.
For the VHE text decoder, we choose to use PGBN to extract hierarchical latent representation from $\tv_n$.
PGBN consists of multiple gamma distributed stochastic hidden layers, generalizing the ``shallow'' Poisson factor analysis \citep{BNBP_PFA_AISTATS2012,NBP2012} into a deep setting.
PGBN with $L$ hidden layers, from top to bottom, is expressed as
\begin{align}\label{PGBN_generate}
&\thetav_n^{(L)} \sim \mbox{Gam}\left(\rv,1/s_n^{(L+1)} \right), ~\rv \sim \mbox{Gam} (\gamma_0/K_L,1/s_0),
\nonumber \\
& \small \thetav_n^{(l)} \sim \mbox{Gam}\left(\Phimat^{(l+1)} \thetav_n^{(l+1)} ,1/s_n^{(l+1)} \right), l=L-1, \cdots, 2,1,~~~ \tv_n \sim \mbox{Pois} \left(\Phimat^{(1)} \thetav_n^{(1)} \right),
\end{align}
where the hidden units $\thetav_n^{(l)} \in \mathbb{R}_+^{K_l}$ of layer $l$ are factorized under the gamma likelihood into the product of topics $\Phimat^{(l)} \in \mathbb{R}_+^{K_{l-1} \times K_{l}}$ and hidden units of the next layer, $\mathbb{R}_+=\{x,x\ge0\}$, $s_n^{(l)}>0$, and $K_l$ is the number of topics of layer $l$. If the texts are represented as binary attribute vectors $\bv_n$,
we can add a Bernoulli-Poisson link layer as $\bv_n=\mathbf{1}(\tv_n\ge 1)$ \citep{EPM_AISTATS2015,GBN}.
We place a Dirichlet prior on each column of $\Phimat^{(l)}$. 
The topics can be organized into a directed acyclic graph (DAG), whose node $k$ at layer $l$ can be visualized with the top words of $\big[ \prod_{t=1}^{l-1} \Phimat^{(t)} \big] \phiv_k^{(l)} $; the topics tend to be very general in the top layer and become increasingly more specific when moving downwards. 
This semantically meaningful latent hierarchy provides unique opportunities to build a better image generator by coupling the semantic hierarchical structures with visual ones.
\vspace{-0.5mm}

Let us denote $\Phimat=\{\Phimat^{(1)},\ldots,\Phimat^{(L)},\rv\}$ as the set of global parameters of PGBN shown in \eqref{PGBN_generate}.
Given $\Phimat$, we adopt the inference in \citet{Zhang2018WHAI} to build an Weibull upward-downward variational image encoder as 
$\prod _{n=1}^N \prod _{l=1}^L q ( \thetav_n^{(l)} \given \xv_n, \Phimat^{(l+1)}, \thetav_n^{(l+1)} )$, 
where $\Phimat^{(L+1)}:=\rv$, $\thetav_n^{(L+1)}:=\emptyset$, and
\begin{equation}\label{Eq: posterior-completely}
    q ( \thetav_n^{(l)} \given \xv_n, 
    \Phimat^{(l+1)}, \thetav_n^{(l+1)} )= \mbox{Weibull}(\kv_n^{(l)}+\Phimat^{(l+1)} \thetav_n^{(l+1)},\lambdav_n^{(l)}).\vspace{-0.2mm}
\end{equation}
The Weibull distribution is used to approximate the gamma distributed conditional posterior, and its parameters $\kv_n^{(l)}, \lambdav_n^{(l)}\in \mathbb{R}^{K_l}$ are deterministically transformed from the convolutional neural network (CNN) image features $f(\xv_n)$ \citep{szegedy2016rethinking},
as shown in Fig.~\ref{Fig: VHE} and described in Appendix D.1.
We denote $\Omegamat$ as the set of encoder parameters.
We refer to \citet{Zhang2018WHAI} for more details about this deterministic-upward--stochastic-downward ladder-structured inference network, which is distinct from a usual inference network that has a pure bottom-up structure and only interacts with the generative model via the ELBO \citep{kingma2014stochastic,gulrajani2017pixelvae}.

The multi-stochastic-layer latent representation $\zv = \{ \thetav^{(l)} \}_{l=1}^L$ is the bridge between two modalities.
As shown in Fig.~\ref{Fig: StackGAN++}, VHE-StackGAN++ simply randomizes the image generator of
StackGAN++ \citep{Zhang2017StackGAN++} with the concatenated vector 
$\thetav = \big[ \thetav^{(1)}, \cdots, \thetav^{(L)} \big]$.
We provide the overall objective function in \eqref{Eq: mvrARGAN-V1} of Appendix D.2. 
Note that existing neural-network-based models \citep{gomez2017self-supervised,xu2018attngan,Zhang2016StackGAN,Zhang2017StackGAN++,verma2018generalized,zhang2018photographic} are often able to perform unidirectional but not bidirectional transforms between images $\xv$ and texts $\tv$.
However, bidirectional transforms are straightforward for the proposed model, as $\zv$ can be either drawn from the image encoder $q(\zv\given \xv)$ in \eqref{Eq: posterior-completely}, or drawn with an upward-downward Gibbs sampler \citep{GBN} from the conditional posteriors $p(\zv\given \tv) $ of the PGBN text decoder $p(\tv\given \zv)$ in \eqref{PGBN_generate}.

\subsection{VHE-raster-scan-GAN with a hierarchical-semantic multi-resolution image generator}

While we find that VHE-StackGAN++ has already achieved impressive results, its simple concatenation of $\thetav^{(l)}$ does not fully exploit the semantically-meaningful hierarchical latent representation of the PGBN-based text decoder.
For three DAG subnets inferred from three different datasets, as shown in Figs.~\ref{fig: flower_text_topic}
-\ref{fig: coco_text_topic} of Appendix C.7, the higher-layer PGBN topics match general visual concepts, such as those on shapes, colors, and backgrounds, while the lower-layer ones provide finer details. 
This motivates us to develop an image generator to exploit the semantic structure, which matches coarse-to-fine visual concepts, to gradually refine its generation.
To this end, as shown in Fig. \ref{Fig: raster-scan-GAN}, we develop ``raster-scan'' GAN that performs generation not only in a multi-scale low-to-high-resolution manner in each layer, but also a hierarchical-semantic coarse-to-fine fashion across layers.
\vspace{-0.5mm}

Suppose we are building a three-layer raster-scan GAN to generate an image of size $256^2$. We randomly select an image $\xv_n$ and then sample $\{\thetav_n^{(l)}\}_{l=1}^3 $ from the variational posterior $\prod _{l=1}^3 q ( \thetav_n^{(l)} \given \xv_n, \Phimat^{(l+1)}, \thetav_n^{(l+1)} )$.
First, the top-layer latent variable $\thetav^{(3)}$, often capturing general semantic information, is transformed to hidden features $h^{(3)}_i$ for the $i^{th}$ branch:
\begin{equation}
 h^{(3)}_1 = F^{(3)}_1(\thetav^{(3)});~~ h^{(3)}_i = F^{(3)}_i(h^{(3)}_{i-1}, \thetav^{(3)}),~~ i=2,3, 
\end{equation}
where $F_i^{(l)}$ is a CNN.
Second, having obtained $\{h_{i}^{(3)}\}_{i=1}^3$, generators $\{G_i^{(3)}\}_{i=1}^{3}$ synthesize low-to-high-resolution image samples
$\{\sv_{i}^{(3)}=G^{(3)}_i(h_i^{(3)})\}_{i=1}^3$, where $\sv_{1}^{(3)}$, $\sv_{2}^{(3)}$, and $\sv_{3}^{(3)}$ are of $16^2$, $32^2$, and $ 64^2$, respectively. Third, $\sv_{3}^{(3)}$ is down-sampled to $\hat{\sv}_{3}^{(3)}$ of size $32^2$ and combined with the information from $\thetav^{(2)}$ to provide the hidden features at layer two:
\begin{equation}
	h^{(2)}_1 = C( F^{(2)}_1(\thetav^{(2)}), \hat{\sv}_{3}^{(3)}); ~~h^{(2)}_i = F^{(2)}_i(h^{(2)}_{i-1}, \thetav^{(2)}), ~~i=2,3, 
\end{equation}
where $C$ denotes concatenation along the channel.
Fourth, the generators synthesize image samples $\{\sv_{i}^{(2)}=G^{(2)}_i(h_i^{(2)})\}_{i=1}^3$,
where $\sv_{1}^{(2)}$, $\sv_{2}^{(2)}$, and $\sv_{3}^{(2)}$ are of $32^2$, $64^2$, and $ 128^2$, respectively.
The same process is then replicated at layer one to generate
$\{\sv_{i}^{(1)}=G^{(1)}_i(h_i^{(1)})\}_{i=1}^3$,
where $\sv_{1}^{(1)}$, $\sv_{2}^{(1)}$, and $\sv_{3}^{(1)}$ are of size $64^2$, $128^2$, and $ 256^2$, respectively, and $\sv_{3}^{(1)}$ becomes a desired high-resolution synthesized image with fine details.
The detailed structure of raster-scan-GAN is described in Fig.~\ref{Fig: appendix_multi-stackGAN} of Appendix D.3. 
PyTorch code is provided to aid the understanding and help reproduce the results.

Different from many existing methods \citep{gomez2017self-supervised,reed2016generative,xu2018attngan,Zhang2017StackGAN++} whose textual feature extraction is separated from the end task, VHE-raster-scan-GAN performs joint optimization. 
As detailedly described in the Algorithm in Appendix E, at each mini-batch based iteration, after updating $\Phimat$ by the topic-layer-adaptive stochastic gradient Riemannian (TLASGR) MCMC of \citet{cong2017deep}, a Weibull distribution based reparameterization gradient \citep{Zhang2018WHAI} is used to end-to-end optimize the following objective:
\begin{align}\label{Eq: mvrARGAN}
&\textstyle \min_{\{G_i^{(l)}\}_{i,l},~\Omegamat} \max_{\{D_i^{(l)}\}_{i,l}}
\mathbb{E}_{p_{\text{data}}(\xv_n, \tv_n)}
\mathbb{E}_
{\prod _{l=1}^3 q ( \thetav_n^{(l)} \given \xv_n, %
	\Phimat^{(l+1)}, \thetav_n^{(l+1)} )}
\big\{-\log p(\tv_n\given \Phimat^{(1)}, \thetav_n^{(1)})
\nonumber\\
&~~~~~~~~~~~~~~ ~~~~~~~~~~~~~~\textstyle + \sum_{l=1}^{3} \mbox{KL}[q(\thetav_n^{(l)}\given \xv_n, \Phimat^{(1+1)}, \thetav_n^{(l+1)})\, ||\, p(\thetav_n^{(l)}\given\Phimat^{(1+1)}, \thetav_n^{(l+1)})] \nonumber\\
&~~~~~~~~~~~~~~ ~~~~~~~~~~~~~~ \textstyle + \sum_{l=1}^3 \sum_{i=1}^3 [\log D_i^{(l)} (\xv_{n,i}^{(l)}, \thetav_n^{(l)}) + \log (1-D_i^{(l)}(G_i^{(l)}(\thetav_n^{(l)}),\thetav_n^{(l)}))] \big \},\vspace{-0.5mm}
\end{align}
where $\{\xv_{n,i}^{(l)}\}_{i=1,l=1}^{3,3}$ denote different resolutions of $\xv_n$, corresponding to $\{\sv_{n,i}^{(l)}\}_{i=1,l=1}^{3,3}$.

\subsection{Related work on joint image-text learning}\label{Sec: Related work}

\citet{gomez2017self-supervised} develop a CNN to learn a transformation from images to textual features pre-extracted by LDA.
GANs have been exploited to generate images given pre-learned textual features extracted by RNNs \citep{denton2015deep,reed2016generative,Zhang2016StackGAN,xu2018attngan,zhang2018photographic,li2019object}.
All these works need a pre-trained linguistic model based on large-scale extra text data and the transformations between images and texts are only unidirectional.
The recently proposed Obj-GAN \citep{li2019object} 
 needs even more side information 
such as the locations and labels of objects inside images, which could be difficult and costly to acquire in practice.
On the other hand, probabilistic graphical model based methods \citep{srivastava2012multimodalDBN,srivastava2012multimodal,wang2018MPGBN} are proposed to learn a joint latent space for images and texts to realize bidirectional transformations, but their image generators are often limited to generating low-level image features.
By contrast, VHE-raster-scan-GAN performs bidirectional end-to-end learning to capture and relate hierarchical visual and semantic concepts across multiple stochastic layers, capable of a wide variety of joint image-text learning and generation tasks, as described below.

\section{Experimental results}

For joint image-text  learning, following previous work, we evaluate the proposed VHE-StackGAN++ and VHE-raster-scan-GAN on three datasets: CUB \citep{WahCUB_200_2011}, Flower \citep{nilsback2008automated}, and COCO \citep{lin2014microsoft}, as described in Appendix F.
Besides the usual text-to-image generation task, due to the distinct bidirectional inference capability of the proposed models, we can perform a rich set of additional tasks such as image-to-text, image-to-image, and noise-to-image-text-pair generations. Due to space constraint, we present below some representative results, and defer additional ones to the Appendix. 
We provide the details of our experimental settings in Appendix F.  PyTorch code is provided at \href{https://github.com/BoChenGroup/VHE-GAN}{https://github.com/BoChenGroup/VHE-GAN}.

\subsection{Text-to-image learning} \label{Sec3_1}

Although the proposed VHE-GANs do not have a text encoder to directly project a document to the shared latent space, given a document and a set of topics inferred during training, we use the upward-downward Gibbs sampler of \citet{GBN} to draw $\{ \thetav^{(l)} \}_{l=1}^L$ from its conditional posterior under PGBN, which are then fed into the GAN image generator to synthesize random images.

\begin{figure*}[t]
	\centering
	\includegraphics[scale=0.38]{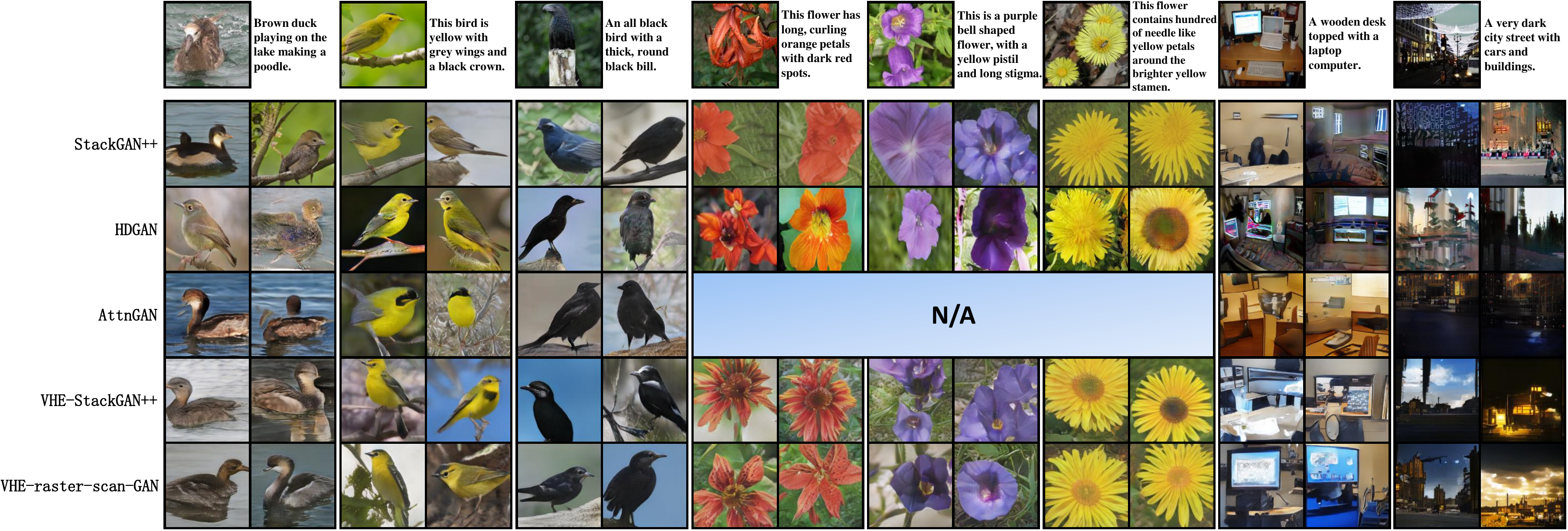}
	\vspace{-6mm}
	\caption{\small Comparison %
		on image generation given texts from CUB, Flower, and COCO. Shown in the top row are the textual descriptions and their associated real images; see Appendix C.2 for higher-resolution images. {Note AttnGAN did not perform experiments on Flower and hence its results on Flower are not shown, and since Obj-GAN only performed experiment on COCO, we defer its visual results to Appendix C.3.}}
	\label{Fig: text_to_image}
	\vspace{-3mm}
\end{figure*}

{\bf{Text-to-image generation: }}
In Tab. \ref{Tab: IS}, with inception score (IS) \citep{salimans2016improved} and Frechet inception distance (FID) \citep{heusel2017gans}, we compare our models with three state-of-the-art GANs in text-to-image generation.
For visualization, we show in the top row of Fig. \ref{Fig: text_to_image} different test textual descriptions and the real images associated with them, and in the other rows random images generated conditioning on these textual descriptions by different algorithms.
{Higher-resolution images are shown in Appendix C.2.}
We also provide example results on COCO, a much more challenging dataset, in Fig. \ref{Fig: coco-another-text-to-image} of Appendix C.3.

\begin{table}[t]
	\caption{\small Inception score (IS, larger is better) and Frechet inception distance (FID, smaller is better)
		of StackGAN++ \citep{Zhang2017StackGAN++}, HDGAN \citep{zhang2018photographic}, AttGAN \citep{xu2018attngan}, {Obj-GAN \citep{li2019object}, and the proposed VHE-raster-scan-GAN};
		the values labeled with $^*$ are calculated by the provided well-trained models and the others are quoted from the original publications; see Tab. \ref{Tab: IS with error bar} in Appendix C.1 for the error bars of IS. {Note that while the FID of Obj-GAN 
		is the lowest, it does not necessarily imply it produces high-quality images, 
		as shown in Figs. \ref{Fig: coco-another-text-to-image} and \ref{Fig: Obj-GAN}; this is because FID only measures the similarity on the image feature space, 
		but ignores the shapes of objects and diversity of generated images. More discussions can be found in Section 3.1 and Appendix G.}
	}
	\centering
	\footnotesize
	\resizebox{0.95\textwidth}{!}{
		\begin{tabular}{c|c|c|c|c|c|c|c|c|c|c}
			\hline
			Method & \mc{2}{|c|}{StackGAN++} & \mc{2}{|c|}{HDGAN} & \mc{2}{|c|}{AttnGAN} & \mc{2}{|c}{{Obj-GAN}} &  \mc{2}{|c}{VHE-raster-scan-GAN} \\ \hline
			Criterion& IS & FID & IS & FID & IS & FID & IS & FID & IS & FID \\ \hline
			Flower & 3.26 & 48.68 & 3.45 & 40.12$^*$ & -- & -- & - & - & {\bf{3.72}} & {\bf{35.13 }} \\ \hline
			CUB & 3.84 & 15.30 & 4.15 & 13.48$^*$ & 4.36 & 13.02$^*$ & - & -  & {\bf{4.41 }} & {\bf{12.02}} \\ \hline
			COCO & 8.30 & 81.59 &11.86 & 78.16$^*$ &25.89 & 77.01$^*$ &{26.58$^*$} & {\bf{36.98$^*$}} & {\bf{27.16 }} & 75.88\\ \hline
	\end{tabular}}
	\label{Tab: IS}
	\caption{\small {Ablation study for image-to-text learning, where the structures of different variations of raster-scan-GAN are illustrated in 
	Figs. \ref{Fig: StackGAN++}, \ref{Fig: GAN}, and  \ref{Fig: raster-GAN}.}}
	\centering
	\footnotesize
	\resizebox{0.95\textwidth}{!}{
		\begin{tabular}{c|c|c|c|c|c|c|c|c}
			\hline
			Method & \mc{2}{|c|}{PGBN+StackGAN++} & \mc{2}{|c|}{{VHE-vanilla-GAN}} & \mc{2}{|c|}{VHE-StackGAN++} & \mc{2}{|c}{{VHE-simple-raster-scan-GAN}}  \\ \hline
			Criterion& IS & FID & IS & FID & IS & FID & IS & FID  \\ \hline
			Flower & 3.29 & 41.04 & {3.01} & {52.15} & 3.56 & 38.66 & {3.62} & {36.18}  \\ \hline
			CUB & 3.92 & 13.79 & {3.52} & {21.24} & 4.20 & 12.93 & {4.31} & {12.35}   \\ \hline
			COCO & 10.63 & 79.65 & {6.36} & {97.15} & 12.63 & 78.02 & {20.13} & {77.18}   \\ \hline
	\end{tabular}}
	\label{Tab: Ablation}
\end{table}

It is clear from Fig. \ref{Fig: text_to_image} that although both StackGAN++ \citep{Zhang2017StackGAN++} and HDGAN \citep{zhang2018photographic} generate photo-realistic images nicely matched to the given texts, they often misrepresent or ignore some key textual information, such as ``black crown'' for the 2nd test text, ``yellow pistil'' for 5th, ``yellow stamen'' for 6th, and ``computer'' for 7th. {These observations also apply to AttnGAN \citep{xu2018attngan}}.
By contrast, both the proposed VHE-StackGAN++ and VHE-raster-scan-GAN do a better job in capturing and faithfully representing these key textual information into their generated images.
Fig. \ref{Fig: coco-another-text-to-image} for COCO further shows the advantages of VHE-raster-scan-GAN in better representing the given textual information in its generated images.
{Note Obj-GAN, which learns a bounding box generator that restricts object locations, obtains the lowest FID on COCO.
However, it appears that this type of restriction significantly improves FID at the expense of sacrificing  the diversity of generated images given text, as shown in Fig. \ref{Fig: Obj-GAN} of Appendix G.
From the results in Fig. \ref{Fig: coco-another-text-to-image}, it also appears that Obj-GAN overly emphasizes correctly arranging the spatial locations of different visual features,  which is important to achieve low FID, but does not do well in generating correct object shapes, which is important to visual effect.
Besides, the training of Obj-GAN requires more side information including the locations and labels of objects in the images, which are often not  provided in practice (e.g., neither CUB nor Flower comes with this type of side information). While the proposed VHE-GAN models do not need these additional side information, they could be further improved by following Obj-GAN to take them into consideration.
}

\begin{figure*}[t]
	\centering
	\subfigure[]{\includegraphics[scale=0.33]{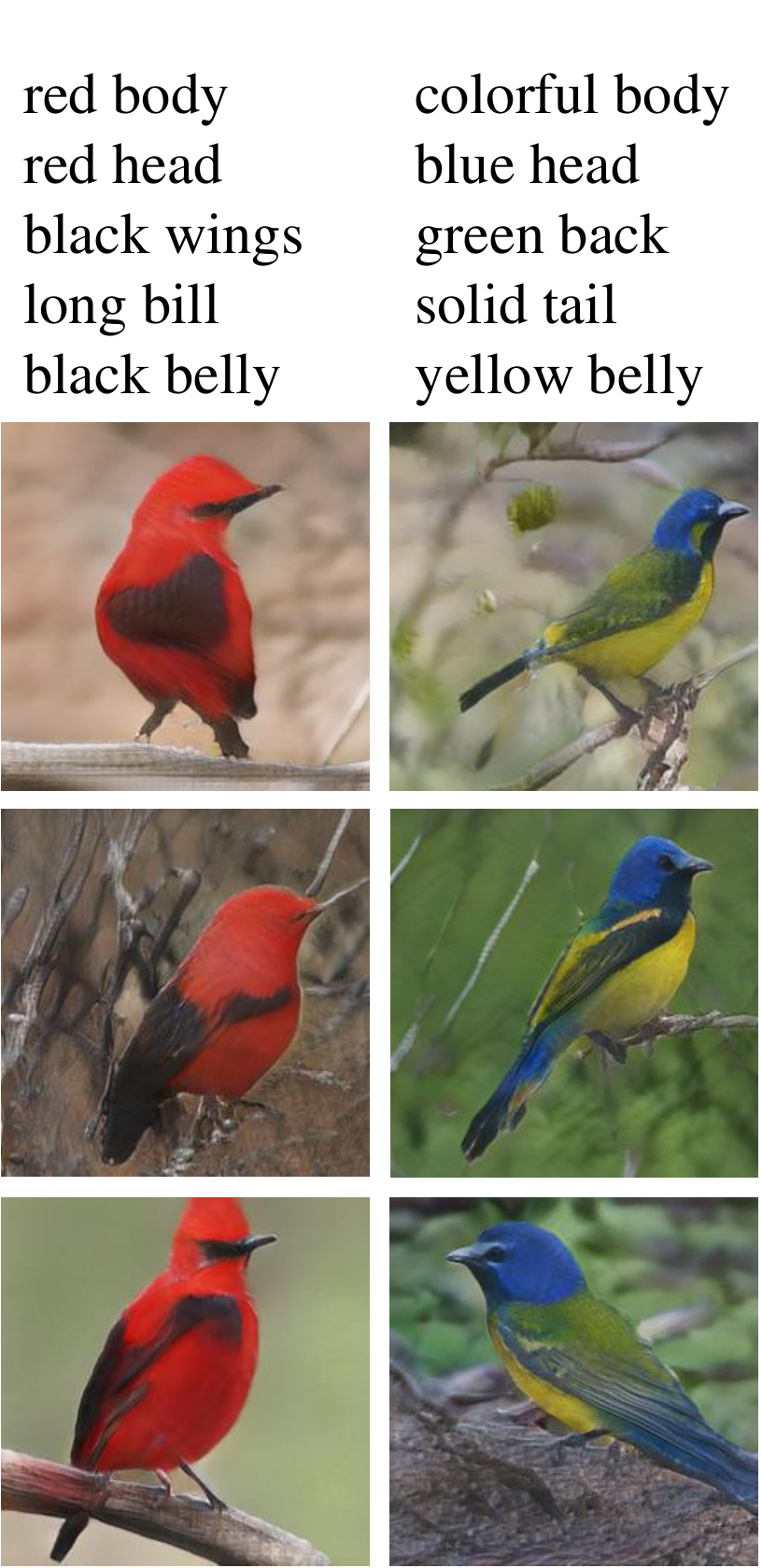}\label{Fig: attribute_generate}}
	\hfil
	\subfigure[]{\includegraphics[scale=0.33]{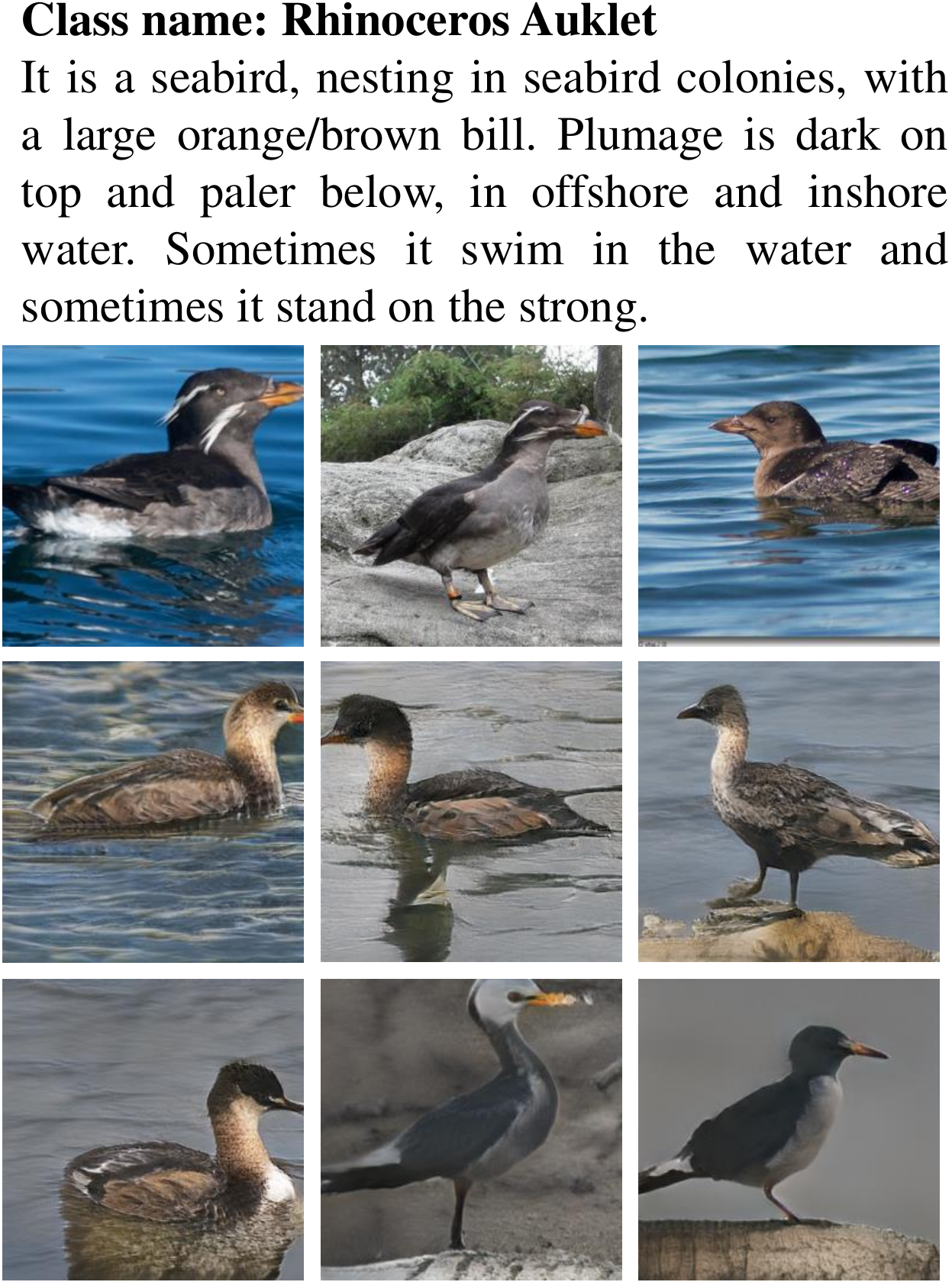}\label{Fig: document_generate}}
	\hfil
	\subfigure[]{\includegraphics[scale=0.33]{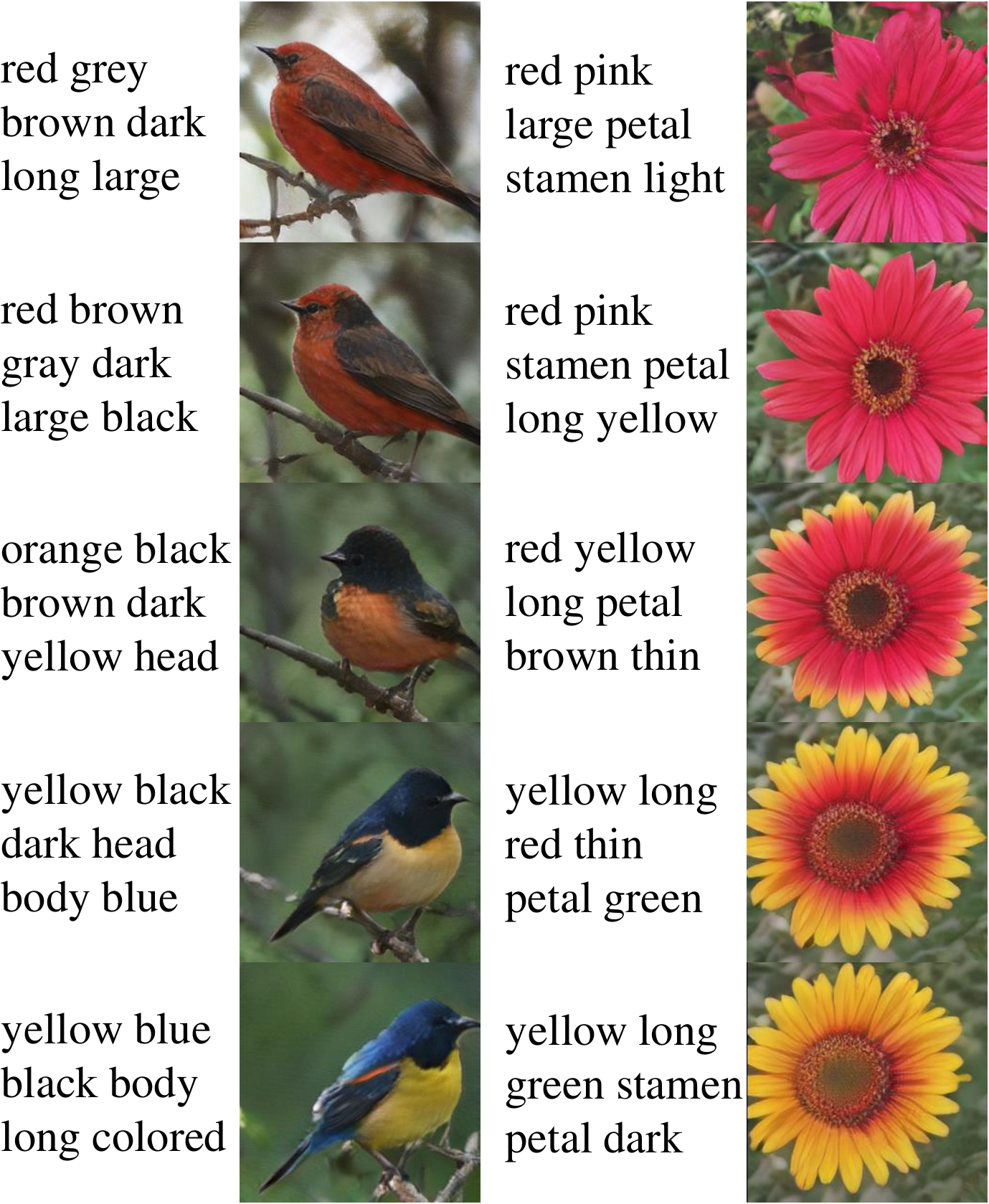}\label{Fig: interpolate}}
	\vspace{-5mm}
	\caption{\small Example results of VHE-raster-scan-GAN on three different tasks:
		(a) image generation given five textual attributes; 
		(b) image generation given a long class-specific document (showing three representative sentences for brevity) from CUB; 
		and (c) latent space interpolation for joint image-text generation on CUB (left column) and Flower (right column), where the texts in the first and last row are given.}
	\vspace{-5mm}
\end{figure*}

As discussed in Section \ref{sec:2.2}, compared with sequence models, topic models can be applied to more diverse textual descriptions, including textual attributes and long documents. 
For illustration, we show in Figs. \ref{Fig: attribute_generate} and \ref{Fig: document_generate}
example images generated conditioning on a set of textual attributes and an encyclopedia document, respectively.
These synthesized  images are photo-realistic and their visual contents well match the semantics of the given texts. 
{Trained on CelebA \citep{liu2015faceattributes}, we provide  in Fig. \ref{Fig: face} examples of facial image generation given attributes; see Appendix B for details. 
}

{{\bf{Ablation studies: }}}We also consider {several ablation studies} for text-to-image generation, as shown in Tab.~\ref{Tab: Ablation}.
{\bf{First}}, we modify  StackGAN++ \citep{Zhang2017StackGAN++}, using the text features extracted by PGBN to replace the original ones by RNN, referred to as PGBN+StackGAN++. 
It is clear that PGBN+StackGAN++ outperforms the original StackGAN++, but underperforms VHE-StackGAN++, which can be explained by that 
1) the PGBN deep topic model is more effective in extracting macro-level textual information, such as key words, than RNNs; 
and 2) jointly end-to-end training the textual feature extractor and image encoder, discriminator, and generator helps better capture and relate the visual and semantical concepts.
{\bf{Second}}, note that VHE-StackGAN++ has the same structured image generator as both StackGAN++ and HDGAN do, but performs better than them.
We attribute its performance gain to 
1) its PGBN deep topic model helps better capture key semantic information from the textual descriptions;
and 
2) it performs end-to-end joint image-text learning via the VHE-GAN framework,
rather than separating the extraction of textual features from text-to-image generation.
{{\bf{Third}}, VHE-vanilla-GAN underperforms VHE-StackGAN++, suggesting that the stacking structure is helpful for generating high resolution images, as previously verified in \citet{Zhang2016StackGAN}.
VHE-simple-raster-scan-GAN outperforms VHE-StackGAN++ but underperforms VHE-raster-scan-GAN, confirming  the benefits of combining the stacking and raster-scan structures. 
More visual results for ablation studies can be found in Appendix C.2.}
Below we focus on illustrating the outstanding performance of 
VHE-raster-scan-GAN.

{\bf{Latent space interpolation: }}
In order to understand the jointly learned image and text manifolds, given texts $\tv_1$ and $\tv_2$, we draw $\thetav_1$ and $\thetav_2$ and use the interpolated variables between them to generate both images via the GAN's image generator and texts via the PGBN text decoder.
As in Fig. \ref{Fig: interpolate}, the first row shows the true texts $\tv_1$ and images generated with $\thetav_1$, the last row shows $\tv_2$ and images generated with $\thetav_2$, and the second to fourth rows show the generated texts and images with the interpolations from $\thetav_1$ to $\thetav_2$. 
The strong correspondences between the generated images and texts, with smooth changes in colors, object positions, and backgrounds between adjacent rows, suggest that the latent space of VHE-raster-scan-GAN is both visually and semantically meaningful. Additional more fine-gridded latent space interpolation results are shown in Figs. \ref{Fig: appendix_interpolate_bird1}-\ref{Fig: appendix_interpolate_flower2} of Appendix C.4.

\begin{figure}[t]
	\centering
	\subfigure[]{\includegraphics[scale=0.32]{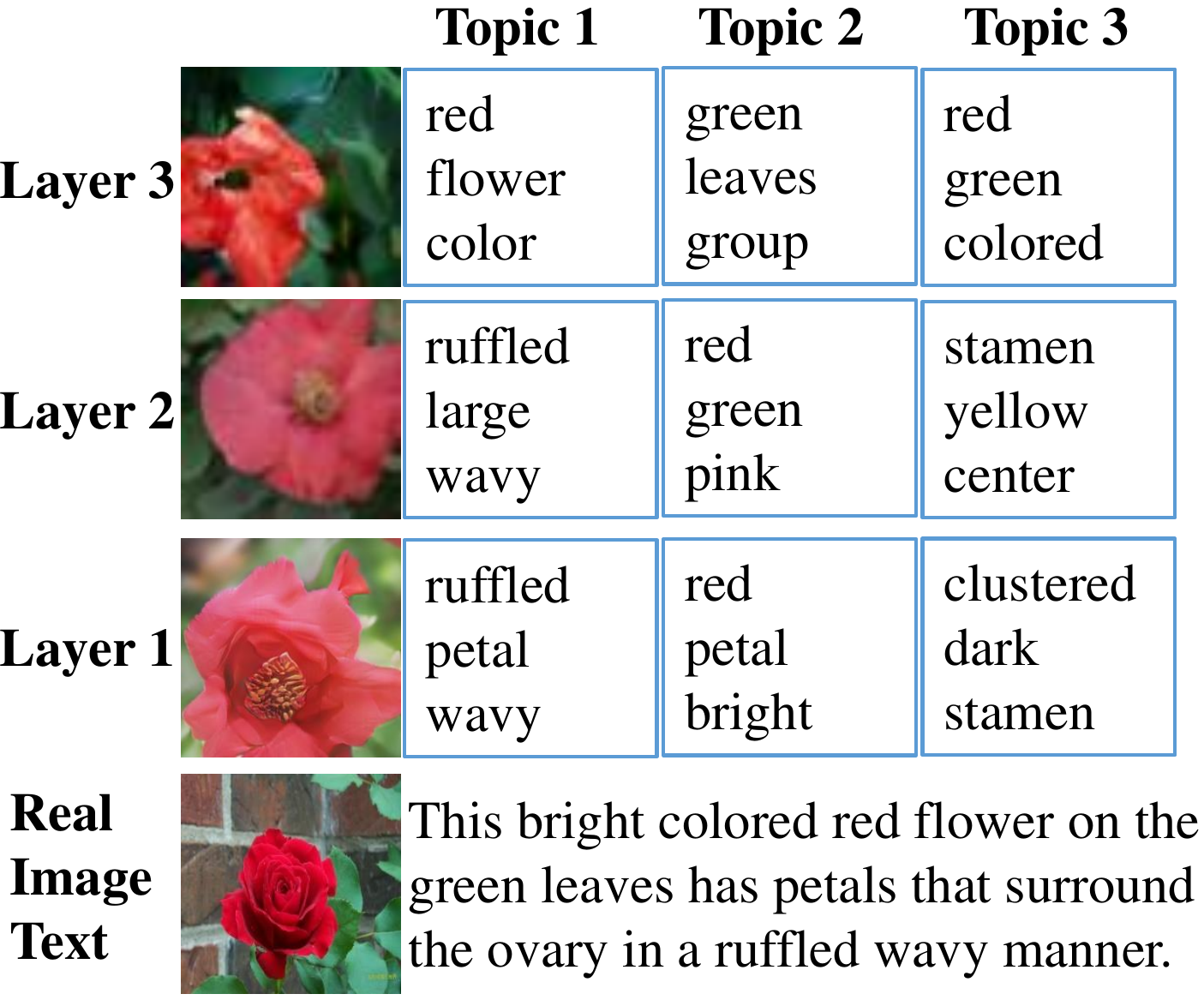}\label{Fig: hierarchical_images_flower}}\hfil
	\subfigure[]{\includegraphics[scale=0.32]{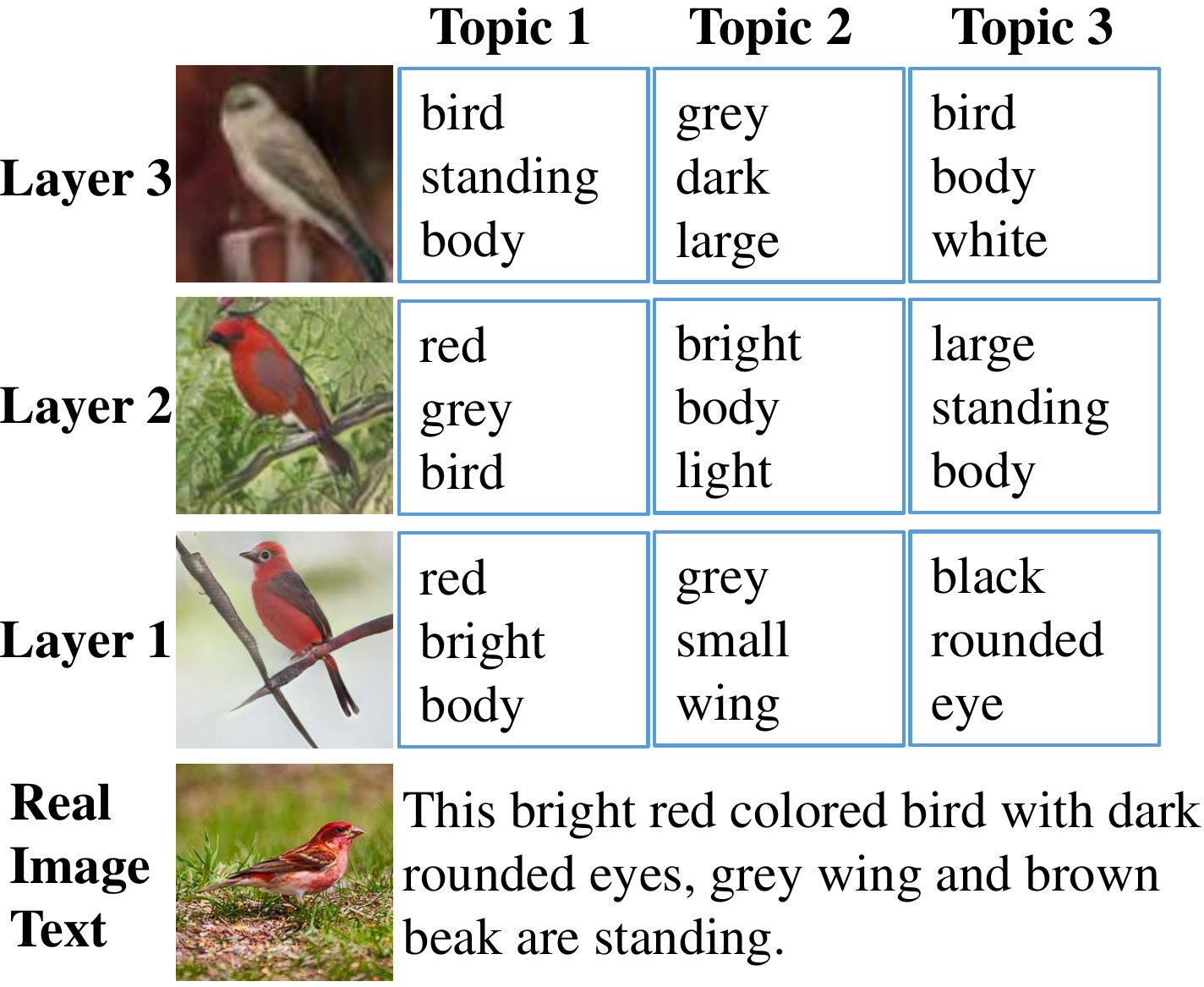}\label{Fig: hierarchical_images_bird}}\hfil
	\subfigure[]{\includegraphics[scale=0.32]{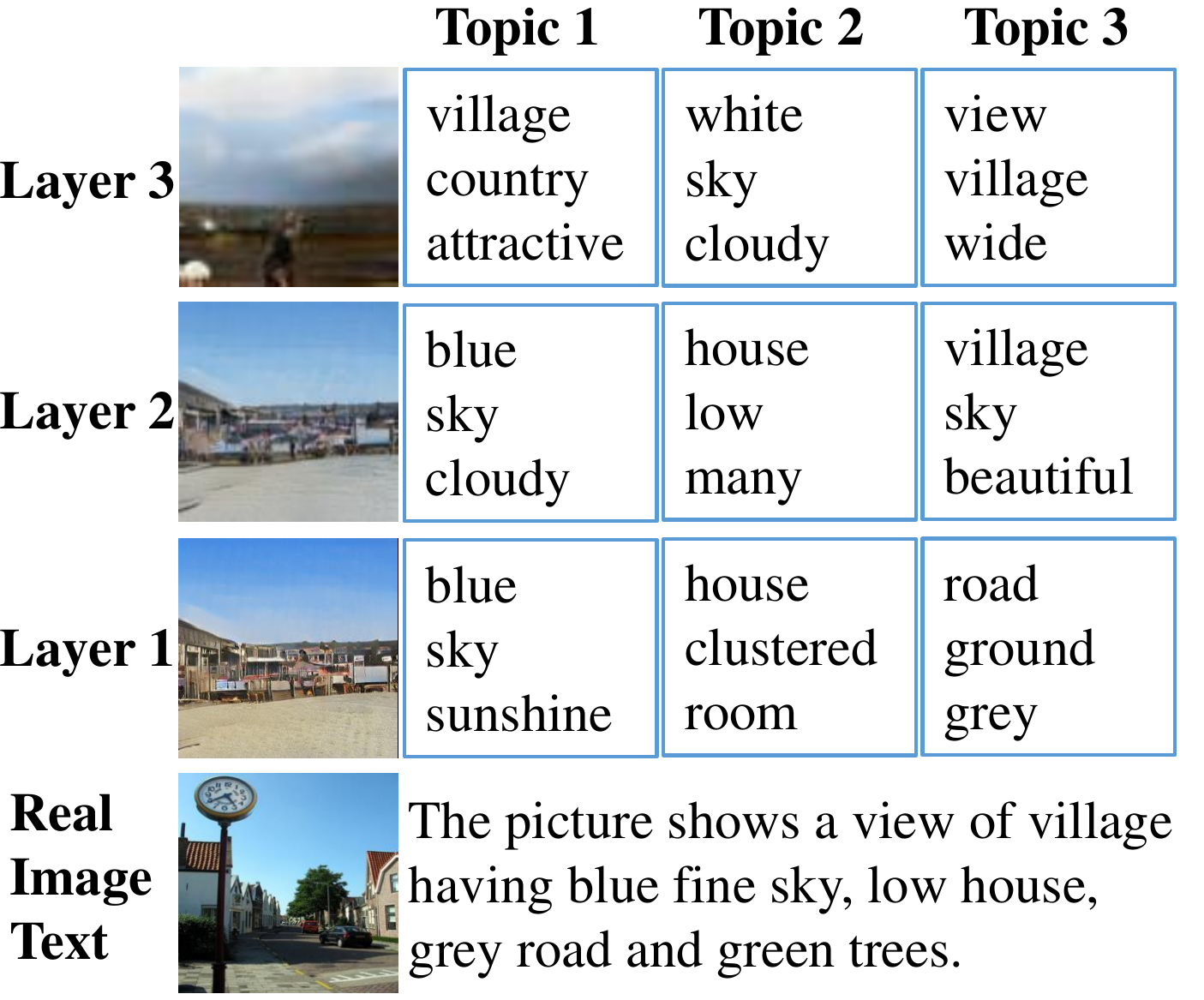}\label{Fig: hierarchical_images_coco}}
	\vspace{-5mm}
	\caption{\small Visualization of example semantic and visual concepts captured by a three-stochastic-hidden-layer VHE-raster-scan-GAN from
		(a) Flower, (b) Bird, and (c) COCO. In each subplot, given the real text $\tv_n$ shown at the bottom, we draw $\{\thetav_n^{(l)}\}_{l=1}^{3}$ via Gibbs sampling; we show the three most active topics in $\Phimat^{(l)}$ (ranked by the weights of $\thetav_n^{(l)}$) at layer $l=3,2,1$, where each topic is visualized by its top three words; and we feed $\{\thetav_n^{(l)}\}_{l=1}^3$ into raster-scan-GAN to generate three random images (one per layer, coarse to fine from layers 3 to 1). }
	\label{Fig: hierarchical_images}
	\vspace{-3mm}
\end{figure}

{\bf{Visualization of captured semantic and visual concepts: }} 
\citet{GBN} show that the semantic concepts extracted by PGBN and their hierarchical relationships can be represented as a DAG, only a subnet of which will be activated given a specific text input. 
In each subplot of Fig. \ref{Fig: hierarchical_images}, we visualize example topic nodes of the DAG subnet activated by the given text input, and show the corresponding images generated at different hidden layers.
There is a good match at each layer between the visual contents of the generated images and semantics of the top activated topics,
which are mainly about general shapes, colors, or backgrounds at the top layer, and become more and more fine-grained when moving downward.
In Fig.~\ref{Fig: Hierarchcial topics_on_COCO}, for the DAG learned on COCO, we show a representative subnet that is rooted at a top-layer node about ``rooms and objects at home,'' and provide both semantic and visual representations for each node.
Being able to capture and relate hierarchical semantic and visual concepts helps explain the state-of-the-art performance of VHE-raster-scan-GAN.

\subsection{Image-to-text learning}

VHE-raster-scan-GAN can perform a wide variety of extra tasks, such as image-to-text generation, text-based zero-shot learning (ZSL), and image retrieval given a text query. In particular, given image $\xv_n$, we draw $\hat{\tv}_n$ as 
$
\hat{\tv}_n\given \thetav_n\sim p(\tv\given \Phimat,\thetav_n),~\thetav_n\given \xv_n\sim q_{\Omegamat}(\thetav\given \Phimat,\xv_n)
$
and use it for downstream tasks.

{\bf{Image-to-text generation: }}
Given an image, we may generate some key words, as shown in Fig. \ref{Fig: image_to_document}, where the true and generated ones are displayed on the left and right of the input image, respectively.
It is clear that VHE-raster-scan-GAN successfully captures the object colors, shapes, locations, and backgrounds to predict relevant key words.

\textbf{Text-based ZSL: }  Text-based  ZSL is a specific task  that learns a relationship between images and texts on the seen classes and transfer it to the unseen ones \citep{Fu2017Recent}.
We follow the the same settings on CUB and Flower as existing text-based ZSL methods summarized in Tab. \ref{Tab: ZSL}.
There are two default splits for CUB---the hard (CUB-H) and easy one (CUB-E)---and one split setting for Flower, as described in Appendix F.
Note that except for our models that infer a shared semantically meaningful latent space between two modalities, none of the other methods have generative models for both modalities, regardless of whether they learn a classifier or a distance metric in a latent space for ZSL.
Tab. \ref{Tab: ZSL} shows that VHE-raster-scan-GAN clearly outperforms the state of the art in terms of the Top-1 accuracy on both the CUB-H and Flower, and is comparable to the second best on CUB-E (it is
the best among all methods that have reported their Top-5 accuracies on CUB-E).
Note for CUB-E, every unseen class has some corresponding seen classes under the same super-category, which makes the classification of surface or distance metric learned on the seen classes easier to generalize to the unseen ones.
We also note that both GAZSL and ZSLPP rely on visual part detection to extract image features, making their performance sensitive to the quality of the visual part detector that often has to be elaborately
tuned for different classes and hence limiting their generalization ability,
for example, the visual part detector for birds is not suitable for flowers.
Tab. \ref{Tab: ZSL} also includes the results of ZSL using VHE, which show that given the same structured text decoder and image encoder, VHE consistently underperforms both VHE-StackGAN++ and VHE-raster-scan-GAN.
This suggests 1) the advantage of a joint generation of two modalities, and 2) the ability of GAN in helping VHE achieve better data representation.
The results in Tab. \ref{Tab: ZSL} also show that the ZSL performance of VHE-raster-scan-GAN has a clear trend of improvement as PGBN becomes
deeper, suggesting the advantage of having a multi-stochastic-hidden-layer deep topic model for text generation.
We also collect the ZSL results of the last 1000 mini-batch based stochastic gradient update iterations to calculate the error bars. For existing methods, since there are no error bars provided in published paper, we only provide the text error bars of the methods that have publicly accessible code.

\begin{table}
\centering
		\tabcaption{\small Accuracy (\%) of ZSL on CUB and Flower. Note that some of them are attribute-based methods but applicable in our setting by replacing attribute vectors with text features (labeled by $^*$), as discussed in \citep{elhoseiny2017link}.}
	    \vspace{3mm}
    	\resizebox{0.9\textwidth}{!}{
		\begin{tabular}{c|c|c|c|c}
			\hline
			Text-ZSL dataset                                    & CUB-H  &\mc{2}{c|}{CUB-E}       & Flower       \\ \hline
			Accuracy criterion                                 & top-1          & top-1          & top-5          & top-1         \\ \hline \hline
			WAC-Kernel \citep{elhoseiny2017write}      & 7.7 $\pm$ 0.28         & 33.5 $\pm$ 0.22         & 64.3 $\pm$ 0.20           & 9.1 $\pm$ 2.77 \\ \hline
			ZSLNS \citep{qiao2016less}                 & 7.3  $\pm$ 0.36         & 29.1 $\pm$ 0.28 & 61.8 $\pm$ 0.22 & 8.7 $\pm$ 2.46 \\ \hline
			ESZSL$^*$ \citep{romeraparedes2015an}      & 7.4 $\pm$ 0.31          & 28.5 $\pm$ 0.26         & 59.9 $\pm$ 0.20         & 8.6 $\pm$ 2.53 \\ \hline
			SynC$^*$ \citep{changpinyo2016synthesized} & 8.6           & 28.0          & 61.3            & 8.2 \\ \hline
			ZSLPP \citep{elhoseiny2017link}            & 9.7           & 37.2          & --            & --           \\ \hline
			GAZSL \citep{zhu2018a}                     & 10.3 $\pm$ 0.26         & {\bf{43.7}} $\pm$ 0.28  & 67.61 $\pm$ 0.24           & --           \\ \hline
			VHE-L3                              & 14.0 $\pm$ 0.24 & 34.6 $\pm$ 0.25 & 64.6 $\pm$ 0.20 & 8.9 $\pm$ 1.57 \\ \hline
			VHE-StackGAN++-L3                              & 16.1 & 38.5 & 68.2 & 10.6 \\ \hline
			VHE-raster-scan-GAN-L1                              & 11.7 $\pm$ 0.31 & 32.1 $\pm$ 0.32 & 62.6 $\pm$ 0.33 & 9.4 $\pm$ 1.68 \\ \hline
			VHE-raster-scan-GAN-L2                              & 14.9 $\pm$ 0.26 & 37.1 $\pm$ 0.24 & 64.6 $\pm$ 0.25 & 11.0 $\pm$ 1.54 \\ \hline
			VHE-raster-scan-GAN-L3                              & {\bf{16.7 $\pm$ 0.24}}  & 39.6 $\pm$ 0.20 & {\bf{70.3 $\pm$ 0.18}} & {\bf{12.1 $\pm$ 1.47}} \\ \hline
		\end{tabular}
    }
    \label{Tab: ZSL}

	\caption{\small {Comparison of the image-to-text retrieval performance, measured by Top-1 accuracy, and text-to-image retrieval performance, measured by AP@50, between different methods on 
	CUB-E. }}
	\centering
	\footnotesize
	\resizebox{1.0\textwidth}{!}{
		\begin{tabular}{c|c|c|c|c|c}
			\hline
			Method & CNN-LSTM & AttnGAN & TA-GAN& VHE-StackGAN++ & VHE-raster-scan-GAN\\
			& {\citep{li2017identity} }& {\citep{xu2018attngan}} & {\citep{nam2018text}} &  & 
			  \\ \hline
			Top1-ACC($\%$) & 61.5 & 55.1 & 61.3 & 60.2 & \bf{61.7}  \\ \hline
			AP@50($\%$) & 57.6 & 51.0 & \bf{62.8} & 61.3 & 62.6  \\ \hline
	\end{tabular}}\vspace{-4mm}
	\label{Tab: Retrieval task}
\end{table}

\vspace{-2mm}
\subsection{{Image/text retrieval}}
\vspace{-1mm}
{As discussed in Section \ref{Sec: Related work}, 
the proposed models are able to infer the shared latent space given either an image or text. We test both VHE-StackGAN++ and VHE-raster-scan-GAN on the same 
image/text retrieval tasks as in TA-GAN \citep{nam2018text}, where we use the cosine distance between the inferred latent space given  images ($q(\thetav \given \xv)$, image encoder) and these given texts ($p(\thetav \given \tv)$, Gibbs sampling) to compute the similarity scores.
Similar with TA-GAN, the top-1 image-to-text retrieval accuracy (Top-1 Acc) and the percentage of matching images in top-50 text-to-image retrieval results (AP@50) on CUB-E dataset are used to measure the performance. As shown in Table \ref{Tab: Retrieval task},
 VHE-raster-scan-GAN clearly outperforms AttnGAN \citep{xu2018attngan}  and is comparable with TA-GAN.
Note TA-GAN needs to extract its text features based on the fastText model  \citep{bojanowski2017enriching} pre-trained on a large corpus, while VHE-raster-scan-GAN learns everything directly from the current dataset in an end-to-end manner.
Also, VHE-raster-scan-GAN outperforms VHE-StackGAN++, which further confirms the benefits of combining both the stacking and raster scan structures. 
} 

\subsection{Generation of random text-image pairs}

\begin{figure}[t]
\centering
		\includegraphics[width=.75\textwidth]{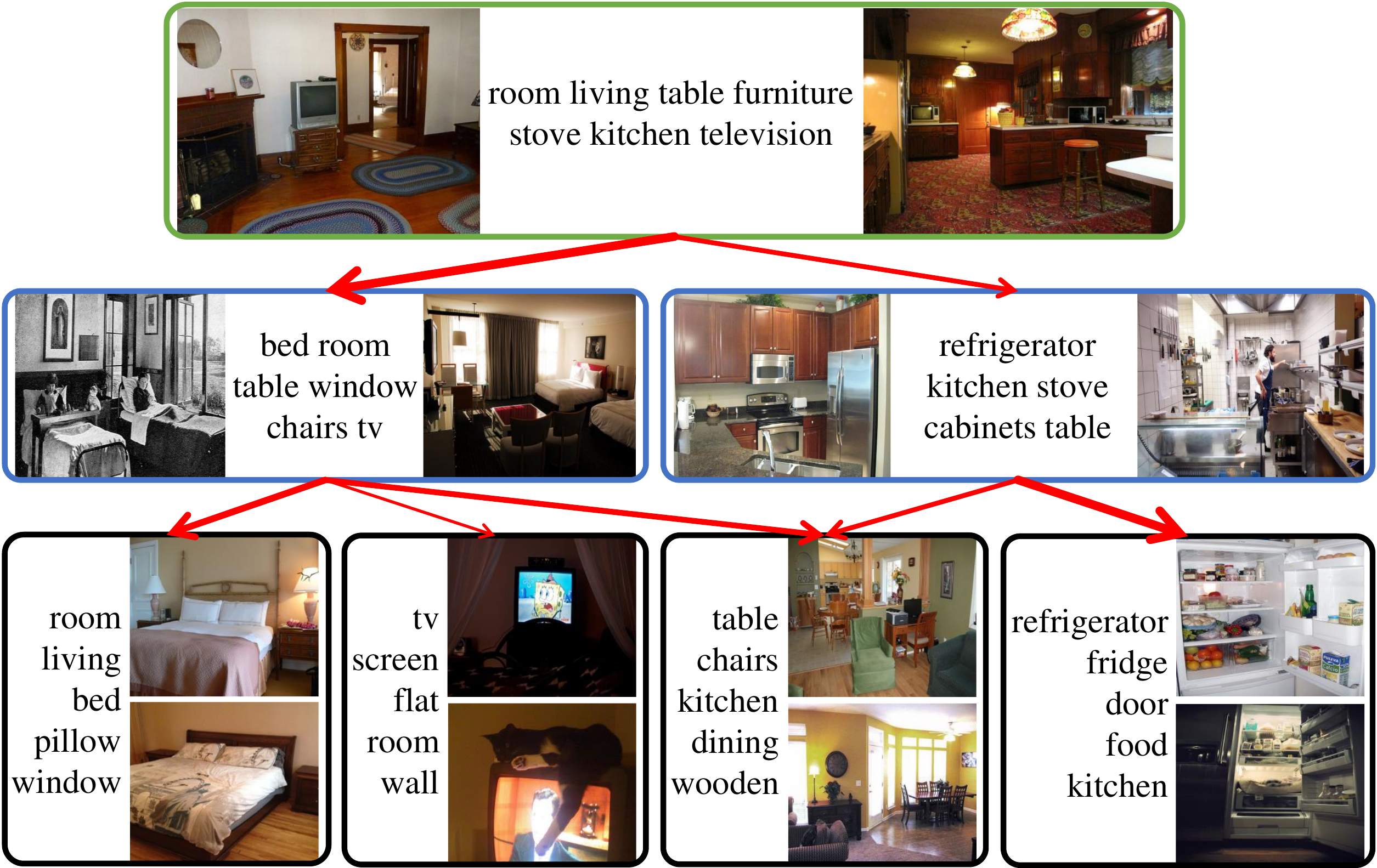}
		\caption{\small An example topic hierarchy learned on COCO and its visual representation. We sample $\thetav_n^{(1:3)} \sim q(\thetav_n^{(1:3)}\given \Phimat, \xv_n)$ for all $n$; for topic node $k$ of layer $l$, we show both its top words and the top two images ranked by their activations $\theta_{nk}^{(l)}$.
		\label{Fig: Hierarchcial topics_on_COCO}}

	\centering
	\subfigure[]{\includegraphics[scale=0.39]{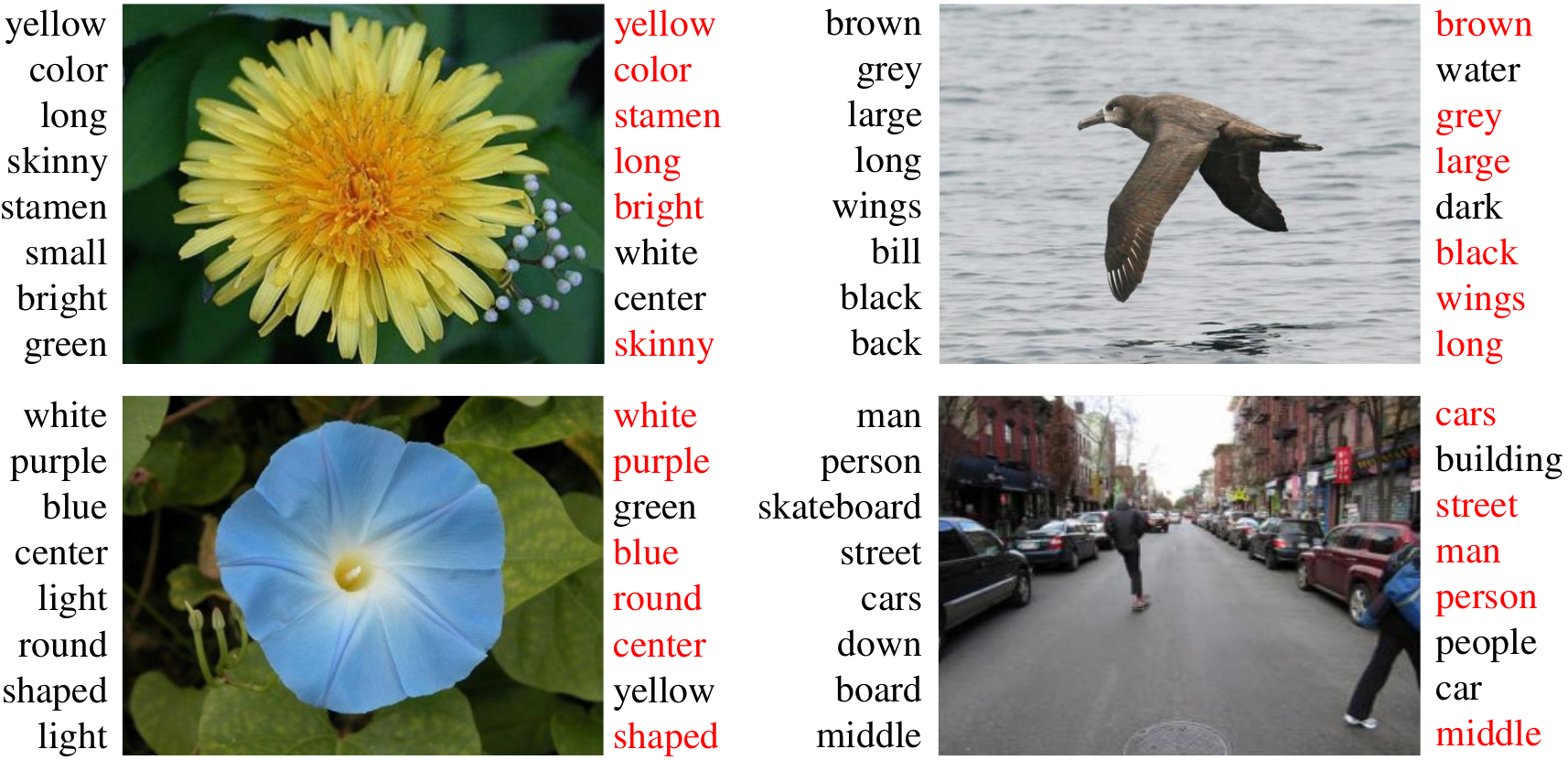}\label{Fig: image_to_document}}\,\,\,\,\,
	\subfigure[]{\includegraphics[scale=0.39]{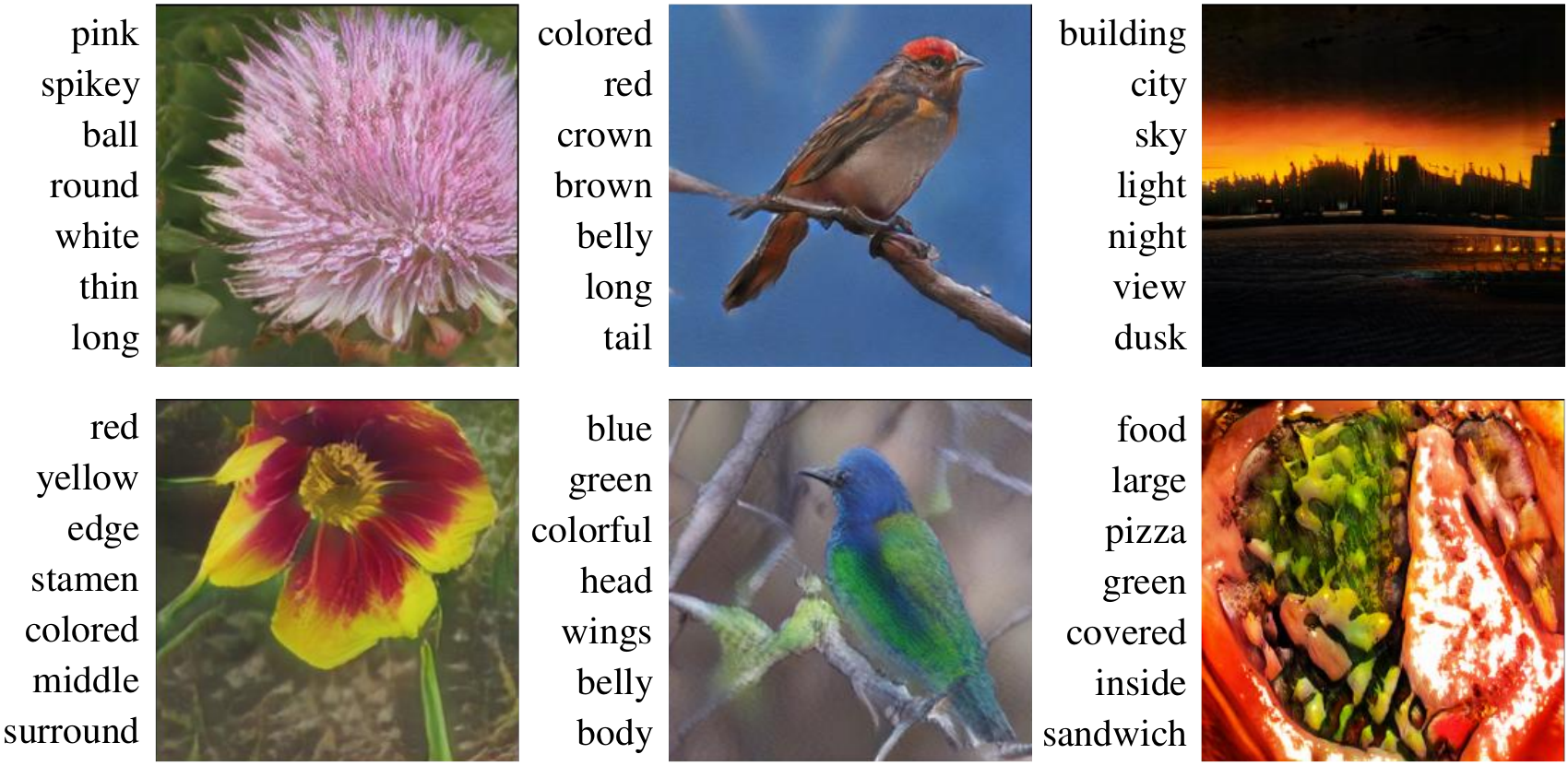}\label{random_noise_generate}}
	\vspace{-5mm}
	\caption{\small Example results of using VHE-raster-scan-GAN for (a) image-to-textual-tags generation, where the generated tags highlighted in red are included in the original ones; (b) image-text-pair generations (columns from left to right are based on Flower, CUB, and COCO, respectively).}\vspace{-2mm}
\end{figure}

Below we show how to generate data samples that contain both modalities.
After training a three-stochastic-hidden-layer VHE-raster-scan-GAN, following the data generation process of the PGBN text decoder,
given $\{ \Phimat^{(l)} \}_{l=1}^{3}$ and $\rv$, we first generate $\thetav^{(3)} \sim \mbox{Gam}\left( \rv, 1/s^{(4)} \right)$ and then downward propagate it through the PGBN as in \eqref{PGBN_generate} to calculate the Poisson rates for all words using $\Phimat^{(1)} \thetav^{(1)}$.
Given a random draw, $\{\thetav^{(l)}\}_{l=1}^{3}$ is fed into the raster-scan-GAN image generator to generate a corresponding image.
Shown in Fig. \ref{random_noise_generate} are six random draws, for each of which we show its top seven words and generated image, whose relationships are clearly interpretable, suggesting
that VHE-raster-scan-GAN is able to recode the key information of both modalities and the relationships between them.
In addition to the tasks shown above, VHE-raster-scan-GAN can also be used to perform image retrieval given a text query, and image regeneration; see Appendices C.5 and C.6 for example results on these additional tasks.

\section{Conclusion}

We develop variational hetero-encoder randomized generative adversarial network (VHE-GAN) to provide a plug-and-play joint image-text modeling framework. VHE-GAN {is a versatile deep generative model} that integrates off-the-shelf image encoders, text decoders, and GAN image discriminators and generators into a coherent end-to-end learning objective. 
It couples its VHE and GAN components by feeding the VHE variational posterior in lieu of noise as the source of randomness of the GAN generator. 
We show VHE-StackGAN++ that combines the Poisson gamma belief network, a deep topic model, and StackGAN++ achieves competitive performance, and VHE-raster-scan-GAN, which further improves VHE-StackGAN++ by exploiting the semantically-meaningful hierarchical structure of the deep topic model, generates photo-realistic images not only in a multi-scale low-to-high-resolution manner, but also in a hierarchical-semantic coarse-to-fine fashion, achieving outstanding results in many challenging image-to-text, text-to-image, and joint text-image learning and generation tasks.

\subsection*{Acknowledgements}
B. Chen acknowledges the support of the Program for Young
Thousand Talent by Chinese Central Government, the 111
Project (No. B18039), NSFC (61771361), NSFC for Distinguished
Young Scholars (61525105), Shaanxi Innovation Team Project, and the Innovation Fund of Xidian University. M. Zhou acknowledges the support of the U.S. National Science Foundation under Grant IIS-1812699.

\bibliography{iclr2020_conference}

\begin{thebibliography}{63}
\providecommand{\natexlab}[1]{#1}
\providecommand{\url}[1]{\texttt{#1}}
\expandafter\ifx\csname urlstyle\endcsname\relax
  \providecommand{\doi}[1]{doi: #1}\else
  \providecommand{\doi}{doi: \begingroup \urlstyle{rm}\Url}\fi

\bibitem[Akata et~al.(2015)Akata, Reed, Walter, Lee, and
  Schiele]{akata2015evaluation}
Zeynep Akata, Scott Reed, Daniel Walter, Honglak Lee, and Bernt Schiele.
\newblock Evaluation of output embeddings for fine-grained image
  classification.
\newblock In \emph{CVPR}, pp.\  2927--2936, 2015.

\bibitem[Bengio et~al.(2003)Bengio, Ducharme, Vincent, and Janvin]{bengio2003a}
Yoshua Bengio, Rejean Ducharme, Pascal Vincent, and Christian Janvin.
\newblock A neural probabilistic language model.
\newblock \emph{Journal of Machine Learning Research}, 3\penalty0 (6):\penalty0
  1137--1155, 2003.

\bibitem[Blei et~al.(2003)Blei, Ng, and Jordan]{blei2003latent}
David~M Blei, Andrew~Y Ng, and Michael~I Jordan.
\newblock Latent {D}irichlet allocation.
\newblock \emph{Journal of Machine Learning Research}, 3:\penalty0 993--1022,
  2003.

\bibitem[Blei et~al.(2017)Blei, Kucukelbir, and McAuliffe]{blei2017variational}
David~M. Blei, Alp Kucukelbir, and Jon~D. McAuliffe.
\newblock Variational inference: {A} review for statisticians.
\newblock \emph{Journal of the American Statistical Association}, 112\penalty0
  (518):\penalty0 859--877, 2017.

\bibitem[Bojanowski et~al.(2017)Bojanowski, Grave, Joulin, and
  Mikolov]{bojanowski2017enriching}
Piotr Bojanowski, Edouard Grave, Armand Joulin, and Tomas Mikolov.
\newblock Enriching word vectors with subword information.
\newblock \emph{Transactions of the Association for Computational Linguistics},
  5:\penalty0 135--146, 2017.

\bibitem[Changpinyo et~al.(2016)Changpinyo, Chao, Gong, and
  Sha]{changpinyo2016synthesized}
Soravit Changpinyo, Wei-Lun Chao, Boqing Gong, and Fei Sha.
\newblock Synthesized classifiers for zero-shot learning.
\newblock In \emph{CVPR}, pp.\  5327--5336, 2016.

\bibitem[Che et~al.(2017)Che, Li, Jacob, Bengio, and Li]{che2017mode}
Tong Che, Yanran Li, Athul~Paul Jacob, Yoshua Bengio, and Wenjie Li.
\newblock Mode regularized generative adversarial networks.
\newblock In \emph{ICLR}, 2017.

\bibitem[Cong et~al.(2017)Cong, Chen, Liu, and Zhou]{cong2017deep}
Yulai Cong, Bo~Chen, Hongwei Liu, and Mingyuan Zhou.
\newblock Deep latent {D}irichlet allocation with topic-layer-adaptive
  stochastic gradient {Riemannian MCMC}.
\newblock In \emph{ICML}, 2017.

\bibitem[Denton et~al.(2015)Denton, Chintala, Szlam, and
  Fergus]{denton2015deep}
Emily~L Denton, Soumith Chintala, Arthur Szlam, and Rob Fergus.
\newblock Deep generative image models using a {L}aplacian pyramid of
  adversarial networks.
\newblock In \emph{NIPS}, pp.\  1486--1494, 2015.

\bibitem[Dieng et~al.(2017)Dieng, Wang, Gao, and Paisley]{dieng2017topicrnn}
Adji~B Dieng, Chong Wang, Jianfeng Gao, and John Paisley.
\newblock {TopicRNN: A} recurrent neural network with long-range semantic
  dependency.
\newblock In \emph{ICLR}, 2017.

\bibitem[Donahue et~al.(2017)Donahue, Krahenbuhl, and
  Darrell]{donahue2017adversarial}
Jeff Donahue, Philipp Krahenbuhl, and Trevor Darrell.
\newblock Adversarial feature learning.
\newblock In \emph{ICLR}, 2017.

\bibitem[Dumoulin et~al.(2017)Dumoulin, Belghazi, Poole, Lamb, Arjovsky,
  Mastropietro, and Courville]{dumoulin2017adversarially}
Vincent Dumoulin, Ishmael Belghazi, Ben Poole, Alex Lamb, Martin Arjovsky,
  Olivier Mastropietro, and Aaron~C Courville.
\newblock Adversarially learned inference.
\newblock In \emph{ICLR}, 2017.

\bibitem[Elhoseiny et~al.(2017{\natexlab{a}})Elhoseiny, Elgammal, and
  Saleh]{elhoseiny2017write}
Mohamed Elhoseiny, Ahmed~M Elgammal, and Babak Saleh.
\newblock Write a classifier: Predicting visual classifiers from unstructured
  text.
\newblock \emph{IEEE Transactions on Pattern Analysis and Machine
  Intelligence}, 39\penalty0 (12):\penalty0 2539--2553, 2017{\natexlab{a}}.

\bibitem[Elhoseiny et~al.(2017{\natexlab{b}})Elhoseiny, Zhu, Zhang, and
  Elgammal]{elhoseiny2017link}
Mohamed Elhoseiny, Yizhe Zhu, Han Zhang, and Ahmed~M Elgammal.
\newblock Link the head to the "beak": Zero shot learning from noisy text
  description at part precision.
\newblock In \emph{CVPR}, pp.\  6288--6297, 2017{\natexlab{b}}.

\bibitem[Fu et~al.(2018)Fu, Xiang, Jiang, Xue, Sigal, and Gong]{Fu2017Recent}
Yanwei Fu, Tao Xiang, Yu~Gang Jiang, Xiangyang Xue, Leonid Sigal, and Shaogang
  Gong.
\newblock Recent advances in zero-shot recognition.
\newblock \emph{IEEE Signal Processing Magazine}, 35, 2018.

\bibitem[Gomez et~al.(2017)Gomez, Patel, Rusinol, Karatzas, and
  Jawahar]{gomez2017self-supervised}
Lluis Gomez, Yash Patel, Marcal Rusinol, Dimosthenis Karatzas, and C~V Jawahar.
\newblock Self-supervised learning of visual features through embedding images
  into text topic spaces.
\newblock In \emph{CVPR}, pp.\  2017--2026, 2017.

\bibitem[Goodfellow et~al.(2014)Goodfellow, Pougetabadie, Mirza, Xu,
  Wardefarley, Ozair, Courville, and Bengio]{goodfellow2014generative}
Ian~J Goodfellow, Jean Pougetabadie, Mehdi Mirza, Bing Xu, David Wardefarley,
  Sherjil Ozair, Aaron~C Courville, and Yoshua Bengio.
\newblock Generative adversarial nets.
\newblock In \emph{NIPS}, pp.\  2672--2680, 2014.

\bibitem[Grover et~al.(2018)Grover, Dhar, and Ermon]{grover2018flow-gan}
Aditya Grover, Manik Dhar, and Stefano Ermon.
\newblock Flow-{GAN}: Combining maximum likelihood and adversarial learning in
  generative models.
\newblock In \emph{AAAI}, pp.\  3069--3076, 2018.

\bibitem[Gulrajani et~al.(2017)Gulrajani, Kumar, Ahmed, Taiga, Visin, Vazquez,
  and Courville]{gulrajani2017pixelvae}
Ishaan Gulrajani, Kundan Kumar, Faruk Ahmed, Adrien~Ali Taiga, Francesco Visin,
  David Vazquez, and Aaron~C Courville.
\newblock Pixel{VAE}: A latent variable model for natural images.
\newblock In \emph{ICLR}, 2017.

\bibitem[Heusel et~al.(2017)Heusel, Ramsauer, Unterthiner, Nessler, and
  Hochreiter]{heusel2017gans}
Martin Heusel, Hubert Ramsauer, Thomas Unterthiner, Bernhard Nessler, and Sepp
  Hochreiter.
\newblock {GAN}s trained by a two time-scale update rule converge to a local
  nash equilibrium.
\newblock In \emph{NIPS}, pp.\  6626--6637, 2017.

\bibitem[Hochreiter \& Schmidhuber(1997)Hochreiter and
  Schmidhuber]{hochreiter1997long}
Sepp Hochreiter and Jurgen Schmidhuber.
\newblock Long short-term memory.
\newblock \emph{Neural Computation}, 9\penalty0 (8):\penalty0 1735--1780, 1997.

\bibitem[Hoffman \& Johnson(2016)Hoffman and Johnson]{hoffman2016elbo}
Matthew~D Hoffman and Matthew~J Johnson.
\newblock {ELBO} surgery: {Yet} another way to carve up the variational
  evidence lower bound.
\newblock In \emph{Workshop in Advances in Approximate Bayesian Inference,
  NIPS}, 2016.

\bibitem[Hoffman et~al.(2013)Hoffman, Blei, Wang, and
  Paisley]{hoffman2013stochastic}
Matthew~D Hoffman, David~M Blei, Chong Wang, and John Paisley.
\newblock Stochastic variational inference.
\newblock \emph{The Journal of Machine Learning Research}, 14\penalty0
  (1):\penalty0 1303--1347, 2013.

\bibitem[Huang et~al.(2018)Huang, Li, He, Sun, and Tan]{huang2018introvae}
Huaibo Huang, Zhihang Li, Ran He, Zhenan Sun, and Tieniu Tan.
\newblock Intro{VAE}: Introspective variational autoencoders for photographic
  image synthesis.
\newblock In \emph{NeurIPS}, 2018.

\bibitem[Jin et~al.(2015)Jin, Fu, Cui, Sha, and Zhang]{jin2015aligning}
Junqi Jin, Kun Fu, Runpeng Cui, Fei Sha, and Changshui Zhang.
\newblock Aligning where to see and what to tell: {I}mage caption with
  region-based attention and scene factorization.
\newblock In \emph{CVPR}, 2015.

\bibitem[Kingma \& Ba(2014)Kingma and Ba]{kingma2014adam}
Diederik~P Kingma and Jimmy Ba.
\newblock {Adam: A} method for stochastic optimization.
\newblock \emph{arXiv preprint arXiv:1412.6980}, 2014.

\bibitem[Kingma \& Welling(2014)Kingma and Welling]{kingma2014stochastic}
Diederik~P Kingma and Max Welling.
\newblock Stochastic gradient {VB} and the variational auto-encoder.
\newblock In \emph{ICLR}, 2014.

\bibitem[Kiros \& Szepesvari(2012)Kiros and Szepesvari]{kiros2012deep}
Ryan Kiros and Csaba Szepesvari.
\newblock Deep representations and codes for image auto-annotation.
\newblock pp.\  908--916, 2012.

\bibitem[Larsen et~al.(2016)Larsen, Sonderby, Larochelle, and
  Winther]{larsen2016autoencoding}
Anders Boesen~Lindbo Larsen, Soren~Kaae Sonderby, Hugo Larochelle, and Ole
  Winther.
\newblock Autoencoding beyond pixels using a learned similarity metric.
\newblock In \emph{ICML}, pp.\  1558--1566, 2016.

\bibitem[Lau et~al.(2017)Lau, Baldwin, and Cohn]{lau2017topically}
Jey~Han Lau, Timothy Baldwin, and Trevor Cohn.
\newblock Topically driven neural language model.
\newblock In \emph{ACL}, pp.\  355--365, 2017.

\bibitem[Li et~al.(2017)Li, Xiao, Li, Yang, and Wang]{li2017identity}
Shuang Li, Tong Xiao, Hongsheng Li, Wei Yang, and Xiaogang Wang.
\newblock Identity-aware textual-visual matching with latent co-attention.
\newblock In \emph{Proceedings of the IEEE International Conference on Computer
  Vision}, pp.\  1890--1899, 2017.

\bibitem[Li et~al.(2019)Li, Zhang, Zhang, Huang, He, Lyu, and
  Gao]{li2019object}
Wenbo Li, Pengchuan Zhang, Lei Zhang, Qiuyuan Huang, Xiaodong He, Siwei Lyu,
  and Jianfeng Gao.
\newblock Object-driven text-to-image synthesis via adversarial training.
\newblock In \emph{Proceedings of the IEEE Conference on Computer Vision and
  Pattern Recognition}, pp.\  12174--12182, 2019.

\bibitem[Lin et~al.(2014)Lin, Maire, Belongie, Hays, Perona, Ramanan, Dollar,
  and Zitnick]{lin2014microsoft}
Tsungyi Lin, Michael Maire, Serge~J Belongie, James Hays, Pietro Perona, Deva
  Ramanan, Piotr Dollar, and C~Lawrence Zitnick.
\newblock Microsoft {COCO}: Common objects in context.
\newblock In \emph{ECCV}, pp.\  740--755, 2014.

\bibitem[Liu et~al.(2015)Liu, Luo, Wang, and Tang]{liu2015faceattributes}
Ziwei Liu, Ping Luo, Xiaogang Wang, and Xiaoou Tang.
\newblock Deep learning face attributes in the wild.
\newblock In \emph{Proceedings of International Conference on Computer Vision
  (ICCV)}, December 2015.

\bibitem[Makhzani et~al.(2015)Makhzani, Shlens, Jaitly, Goodfellow, and
  Frey]{makhzani2015adversarial}
Alireza Makhzani, Jonathon Shlens, Navdeep Jaitly, Ian Goodfellow, and Brendan
  Frey.
\newblock Adversarial autoencoders.
\newblock \emph{arXiv preprint arXiv:1511.05644}, 2015.

\bibitem[Mescheder et~al.(2017)Mescheder, Nowozin, and
  Geiger]{mescheder2017adversarial}
Lars Mescheder, S~Nowozin, and Andreas Geiger.
\newblock Adversarial variational {B}ayes: {U}nifying variational autoencoders
  and generative adversarial networks.
\newblock In \emph{ICML}, pp.\  2391--2400. PMLR, 2017.

\bibitem[Nam et~al.(2018)Nam, Kim, and Kim]{nam2018text}
Seonghyeon Nam, Yunji Kim, and Seon~Joo Kim.
\newblock Text-adaptive generative adversarial networks: {M}anipulating images
  with natural language.
\newblock In \emph{Advances in Neural Information Processing Systems}, pp.\
  42--51, 2018.

\bibitem[Nilsback \& Zisserman(2008)Nilsback and
  Zisserman]{nilsback2008automated}
Maria-Elena Nilsback and Andrew Zisserman.
\newblock Automated flower classification over a large number of classes.
\newblock In \emph{Computer Vision, Graphics \& Image Processing, 2008.
  ICVGIP'08. Sixth Indian Conference on}, pp.\  722--729. IEEE, 2008.

\bibitem[Qiao et~al.(2016)Qiao, Liu, Shen, and Den~Hengel]{qiao2016less}
Ruizhi Qiao, Lingqiao Liu, Chunhua Shen, and Anton~Van Den~Hengel.
\newblock Less is more: Zero-shot learning from online textual documents with
  noise suppression.
\newblock In \emph{CVPR}, pp.\  2249--2257, 2016.

\bibitem[Reed et~al.(2016)Reed, Akata, Yan, Logeswaran, Schiele, and
  Lee]{reed2016generative}
Scott~E Reed, Zeynep Akata, Xinchen Yan, Lajanugen Logeswaran, Bernt Schiele,
  and Honglak Lee.
\newblock Generative adversarial text to image synthesis.
\newblock In \emph{ICML}, pp.\  1060--1069, 2016.

\bibitem[Rezende et~al.(2014)Rezende, Mohamed, and
  Wierstra]{rezende2014stochastic}
Danilo~Jimenez Rezende, Shakir Mohamed, and Daan Wierstra.
\newblock Stochastic backpropagation and approximate inference in deep
  generative models.
\newblock In \emph{ICML}, pp.\  1278--1286, 2014.

\bibitem[Romeraparedes \& Torr(2015)Romeraparedes and
  Torr]{romeraparedes2015an}
Bernardino Romeraparedes and Philip H~S Torr.
\newblock An embarrassingly simple approach to zero-shot learning.
\newblock In \emph{ICML}, pp.\  2152--2161, 2015.

\bibitem[Salimans et~al.(2016)Salimans, Goodfellow, Zaremba, Cheung, Radford,
  and Chen]{salimans2016improved}
Tim Salimans, Ian~J Goodfellow, Wojciech Zaremba, Vicki Cheung, Alec Radford,
  and Xi~Chen.
\newblock Improved techniques for training {GAN}s.
\newblock In \emph{NIPS}, pp.\  2234--2242, 2016.

\bibitem[Srivastava et~al.(2017)Srivastava, Valkoz, Russell, Gutmann, and
  Sutton]{srivastava2017veegan}
Akash Srivastava, Lazar Valkoz, Chris Russell, Michael~U Gutmann, and Charles~A
  Sutton.
\newblock {VEEGAN}: Reducing mode collapse in {GAN}s using implicit variational
  learning.
\newblock In \emph{NIPS}, pp.\  3308--3318, 2017.

\bibitem[Srivastava \& Salakhutdinov(2012{\natexlab{a}})Srivastava and
  Salakhutdinov]{srivastava2012multimodal}
Nitish Srivastava and Ruslan Salakhutdinov.
\newblock Multimodal learning with deep {B}oltzmann machines.
\newblock In \emph{NIPS}, pp.\  2222--2230, 2012{\natexlab{a}}.

\bibitem[Srivastava \& Salakhutdinov(2012{\natexlab{b}})Srivastava and
  Salakhutdinov]{srivastava2012multimodalDBN}
Nitish Srivastava and Ruslan Salakhutdinov.
\newblock Learning representations for multimodal data with deep belief nets.
\newblock In \emph{NIPS workshop}, pp.\  2222--2230, 2012{\natexlab{b}}.

\bibitem[Srivastava \& Salakhutdinov(2014)Srivastava and
  Salakhutdinov]{srivastava2014multimodal}
Nitish Srivastava and Ruslan Salakhutdinov.
\newblock Multimodal learning with deep {B}oltzmann machines.
\newblock \emph{Journal of Machine Learning Research}, 15\penalty0
  (1):\penalty0 2949--2980, 2014.

\bibitem[Szegedy et~al.(2016)Szegedy, Vanhoucke, Ioffe, Shlens, and
  Wojna]{szegedy2016rethinking}
Christian Szegedy, Vincent Vanhoucke, Sergey Ioffe, Jon Shlens, and Zbigniew
  Wojna.
\newblock Rethinking the inception architecture for computer vision.
\newblock In \emph{CVPR}, pp.\  2818--2826, 2016.

\bibitem[Tolstikhin et~al.(2018)Tolstikhin, Bousquet, Gelly, and
  Schoelkopf]{Tolstikhin2017WVAE}
Ilya Tolstikhin, Olivier Bousquet, Sylvain Gelly, and Bernhard Schoelkopf.
\newblock Wasserstein auto-encoders.
\newblock In \emph{ICLR}, 2018.

\bibitem[Verma et~al.(2018)Verma, Arora, Mishra, and Rai]{verma2018generalized}
Vinay~Kumar Verma, Gundeep Arora, Ashish~Kumar Mishra, and Piyush Rai.
\newblock Generalized zero-shot learning via synthesized examples.
\newblock In \emph{CVPR}, pp.\  4281--4289, 2018.

\bibitem[Wah et~al.(2011)Wah, Branson, Welinder, Perona, and
  Belongie]{WahCUB_200_2011}
C.~Wah, S.~Branson, P.~Welinder, P.~Perona, and S.~Belongie.
\newblock {The Caltech-UCSD Birds-200-2011 Dataset}.
\newblock Technical report, 2011.

\bibitem[Wang et~al.(2018)Wang, Chen, and Zhou]{wang2018MPGBN}
Chaojie Wang, Bo~Chen, and Mingyuan Zhou.
\newblock Multimodal {P}oisson gamma belief network.
\newblock In \emph{AAAI}, 2018.

\bibitem[Wang et~al.(2009)Wang, Zhu, Li, and Gong]{wang2009multi-document}
Dingding Wang, Shenghuo Zhu, Tao Li, and Yihong Gong.
\newblock Multi-document summarization using sentence-based topic models.
\newblock In \emph{ACL}, pp.\  297--300, 2009.

\bibitem[Xu et~al.(2018)Xu, Zhang, Huang, Zhang, Gan, Huang, and
  He]{xu2018attngan}
Tao Xu, Pengchuan Zhang, Qiuyuan Huang, Han Zhang, Zhe Gan, Xiaolei Huang, and
  Xiaodong He.
\newblock {A}ttn{GAN}: Fine-grained text to image generation with attentional
  generative adversarial networks.
\newblock In \emph{CVPR}, pp.\  1316--1324, 2018.

\bibitem[Zhang et~al.(2017{\natexlab{a}})Zhang, Xu, and Li]{Zhang2016StackGAN}
Han Zhang, Tao Xu, and Hongsheng Li.
\newblock Stack{GAN}: {T}ext to photo-realistic image synthesis with stacked
  generative adversarial networks.
\newblock In \emph{CVPR}, 2017{\natexlab{a}}.

\bibitem[Zhang et~al.(2017{\natexlab{b}})Zhang, Xu, Li, Zhang, Wang, Huang, and
  Metaxas]{Zhang2017StackGAN++}
Han Zhang, Tao Xu, Hongsheng Li, Shaoting Zhang, Xiaogang Wang, Xiaolei Huang,
  and Dimitris Metaxas.
\newblock {S}tack{GAN}++: {R}ealistic image synthesis with stacked generative
  adversarial networks.
\newblock \emph{IEEE Transactions on Pattern Analysis and Machine
  Intelligence}, PP\penalty0 (99):\penalty0 1--1, 2017{\natexlab{b}}.

\bibitem[Zhang et~al.(2018{\natexlab{a}})Zhang, Chen, Guo, and
  Zhou]{Zhang2018WHAI}
Hao Zhang, Bo~Chen, Dandan Guo, and Mingyuan Zhou.
\newblock {WHAI: W}eibull hybrid autoencoding inference for deep topic
  modeling.
\newblock In \emph{ICLR}, 2018{\natexlab{a}}.

\bibitem[Zhang et~al.(2018{\natexlab{b}})Zhang, Xie, and
  Yang]{zhang2018photographic}
Zizhao Zhang, Yuanpu Xie, and Lin Yang.
\newblock Photographic text-to-image synthesis with a hierarchically-nested
  adversarial network.
\newblock In \emph{CVPR}, 2018{\natexlab{b}}.

\bibitem[Zhou(2015)]{EPM_AISTATS2015}
Mingyuan Zhou.
\newblock Infinite edge partition models for overlapping community detection
  and link prediction.
\newblock In \emph{AISTATS}, pp.\  1135--1143, 2015.

\bibitem[Zhou \& Carin(2015)Zhou and Carin]{NBP2012}
Mingyuan Zhou and Lawrence Carin.
\newblock Negative binomial process count and mixture modeling.
\newblock \emph{IEEE Trans. Pattern Anal. Mach. Intell.}, 37\penalty0
  (2):\penalty0 307--320, 2015.

\bibitem[Zhou et~al.(2012)Zhou, Hannah, Dunson, and
  Carin]{BNBP_PFA_AISTATS2012}
Mingyuan Zhou, Lauren Hannah, David Dunson, and Lawrence Carin.
\newblock Beta-negative binomial process and {P}oisson factor analysis.
\newblock In \emph{AISTATS}, pp.\  1462--1471, 2012.

\bibitem[Zhou et~al.(2016)Zhou, Cong, and Chen]{GBN}
Mingyuan Zhou, Yulai Cong, and Bo~Chen.
\newblock Augmentable gamma belief networks.
\newblock \emph{Journal of Machine Learning Research}, 17\penalty0
  (163):\penalty0 1--44, 2016.

\bibitem[Zhu et~al.(2018)Zhu, Elhoseiny, Liu, Peng, and Elgammal]{zhu2018a}
Yizhe Zhu, Mohamed Elhoseiny, Bingchen Liu, Xi~Peng, and Ahmed~M Elgammal.
\newblock A generative adversarial approach for zero-shot learning from noisy
  texts.
\newblock In \emph{CVPR}, 2018.

\end{thebibliography}
\bibliographystyle{iclr2020_conference}

\normalsize
\clearpage
\appendix

\section{Model property of VHE-GAN and related work}

Let us denote $ \textstyle q(\zv) = \E_{\xv\sim p_{\text{data}}(\xv)}[q(\zv\given \xv)] =\frac{1}{N} \sum_{n=1}^{N} {q(\zv\given \xv_n)}$ as the aggregated posterior \citep{hoffman2016elbo,makhzani2015adversarial}.
Removing the triple-data-reuse training strategy, we can re-express the VHE-GAN objective in \eqref{Eq:vrAEGan} as
\begin{align}\label{Eq:vrAEGan_naive}
&\min \limits_{E, G_{\text{vae}},G_{\text{gan}}}\max \limits_{D} [-\mbox{ELBO}_{\text{vhe}}+\mathcal{L}_{\text{gan}}],
{\mathcal{L}_{\text{gan}}:=
	\E_{\xv\sim p_{\text{data}}(\xv)} \ln D(\xv)+ \E_{\zv\sim q(\zv)} \ln (1-D(G_{\text{gan}}(\zv)) ), }    \!
\end{align}
which corresponds to a naive combination of the VHE and GAN training objectives, where the data samples used to train the VHE, GAN generator, and GAN discriminator in each gradient update iteration are not imposed to be the same. While the naive objective function in \eqref{Eq:vrAEGan_naive} differs from the true one in \eqref{Eq:vrAEGan} that is used to train VHE-GAN, it simplifies the analysis of its theoretical property, as described below.

Let us denote $q(\zv,\xv,\tv):=q(\zv\given \xv)p_{data}(\xv,\tv)$ as the joint distribution of $(\xv,\tv)$ and $\zv$ under the VHE variational posterior $q(\zv\given \xv)$, $I_{q}(\xv,\zv): = \E_{q(\zv,\xv)}\big[ \ln \frac{q(\zv, \xv) }{q(\zv)p_{data}(\xv)}\big] $
as the mutual information between $\xv\sim p_{data}(\xv)$ and $\zv\sim q(\zv)$, and $\mbox{JDS}(p_1||p_2):=\frac{1}{2}\KL [p_1|| (p_1+p_2)/2]+ \frac{1}{2}\KL [p_2|| (p_1+p_2)/2]$ as the Jensen--Shannon divergence between distributions $p_1$ and $p_2$.
Similar to the analysis in \citet{hoffman2016elbo}, the VHE's ELBO can be rewritten as $
\mbox{ELBO}_{\text{vhe}} =\mathbb{E}_{q(\zv,\xv,\tv)} \left[ \log p(\tv\given \zv) \right] - I_{q}(\xv,\zv)-\KL[q(\zv)||p(\zv)],
$
where the mutual information term can also be expressed as $I_{q}(\xv,\zv) = \E_{\xv\sim p_{data}(\xv)} \KL[q(\zv\given \xv)||q(\zv)]$.
Thus maximizing the ELBO encourages the mutual information term $I_{q}(\xv,\zv)$ to be minimized, which means while the data reconstruction term $\E_{q(\zv,\xv,\tv)} \left[ \log p(\tv\given \zv) \right] $ needs to be maximized, part of the VHE optimization objective penalizes a $\zv$ from carrying the information of the $\xv$ that it is encoded from. This mechanism helps
provide necessary regularization to prevent overfitting. %
As in  \citet{goodfellow2014generative}, with an optimal discriminator $D^*_{G}$ for generator $G$, we have
$
\min
\mathcal{L}_{\text{GAN}}(D^*_{G},G) = \ln 4+2 \mbox{JSD}(p_{data}(\xv)||p_{G_{\zv}}(\xv)),\notag
$
where $p_{G_{\zv}(\xv)}$ denotes the distribution of the generated data $G(\zv)$ that use $\zv\sim q(\zv)$ as the random source fed into the GAN generator. The JSD term is minimized when $p_{G_{\zv}}(\xv)=p_{data}(\xv)$.

With these analyses,
given an optimal GAN discriminator,
the naive VHE-GAN objective function in \eqref{Eq:vrAEGan_naive} reduces to
\begin{align}
& \min \limits_{E, G_{\text{gan}}, G_{\text{vae}}} %
- \E_{q(\zv,\xv,\tv)} \left[ \log p(\tv\given \zv) \right]+ {\KL}[q(\zv)||p(\zv)]+
I_{q}(\xv,\zv) + 2 {\mbox{JSD}}(p_{data}(\xv)||p_{G_{\zv}}(\xv)). \! \label{eq:PRGAN}
\end{align}

From the VHEs' point of view, examining \eqref{eq:PRGAN} shows that it alleviates the inherent conflict in VHE of maximizing the ELBO and maximizing the mutual information $I_{q}(\xv,\zv)$. This is because while the VHE part of VHE-GAN still relies on minimizing $I_{q}(\xv,\zv)$ to regularize the learning, the GAN part tries to transform $q(\zv)$ through the GAN generator to match the true data distribution $p_{data}(\xv)$. In other words, while its VHE part
penalizes a $\zv$ from carrying the information about the $\xv$ that it is encoded from, its GAN part encourages a $\zv$ to carry information about the true data distribution $p_{data}(\xv)$, but not necessarily the observed $\xv$ that it is encoded from.

From the GANs' point of view, examining \eqref{eq:PRGAN} shows that it provides GAN with a meaningful latent space, necessary for performing inference and data reconstruction (with the aid of the data-triple-use training strategy).
More specifically, this latent representation is also used by the VHE to maximize the data log-likelihood, a training procedure that tries to cover all modes of the empirical data distribution rather than dropping modes. %
For VHE-GAN \eqref{Eq:vrAEGan}, the source distribution is $q(\zv\given \xv)$, %
not only allowing GANs to participate in posterior inference and data reconstruction, but also helping GANs resist mode collapse.
In the following, we discuss some related works on combining VAEs and GANs.

\subsection{Related work on combining VAEs and GANs}
Examples in improving VAEs with adversarial learning include \citet{mescheder2017adversarial}, which allows the VAEs to take implicit encoder distribution, and adversarial auto-encoder \citep{makhzani2015adversarial} and Wasserstein auto-encoder \citep{Tolstikhin2017WVAE}, which drop the mutual information term from the ELBO and use adversarial learning to match the aggregated posterior and prior.
Examples in allowing GANs to perform inference include \citet{dumoulin2017adversarially}
and \citet{donahue2017adversarial}, which use GANs to match the joint distribution $q(\zv\given \xv)p_{data}(\xv)$ defined by the encoder and the one $p(\xv\given \zv)p(\zv)$ defined by the generator.
However, they often do not provide good data reconstruction.
Examples in using VAEs or maximum likelihood to help GANs resist mode collapse include \citet{che2017mode,srivastava2017veegan,grover2018flow-gan}.
Another example is VAEGAN \citep{larsen2016autoencoding} that combines unit-wise likelihood at hidden layer and adversarial loss at original space, but its update of the encoder is separated from the GAN mini-max objective.
On the contrary, IntroVAE \citep{huang2018introvae} retains the pixel-wise likelihood with an adversarial regularization on the latent space.
Sharing network between the VAE decoder and GAN generator in VAEGAN and IntroVAE, however, %
limit them to model a single modality.

\section{More discussion on sequence  models and topic models in text analysis.}
In Section 3.1, we have discussed two models to represent the text: sequence models and topic models.
Considering the versatility of  topic models \citep{wang2009multi-document, jin2015aligning, GBN, srivastava2012multimodal, srivastava2014multimodal, wang2018MPGBN, elhoseiny2017link, zhu2018a} in dealing with different types of textual information, and its
effectiveness in capturing latent topics that are often directly related to macro-level visual information \citep{gomez2017self-supervised,dieng2017topicrnn, lau2017topically}, we choose a state-of-the-art deep topic model, PGBN, to model the  textual descriptions in VHE.
Due to space constraint, we only provide simple illustrations in Figs.~\ref{Fig: attribute_generate} and \ref{Fig: document_generate}.
In this section, more insights and discussions are provided. %

\begin{figure}[ht]
	\centering
	\includegraphics[scale=0.46]{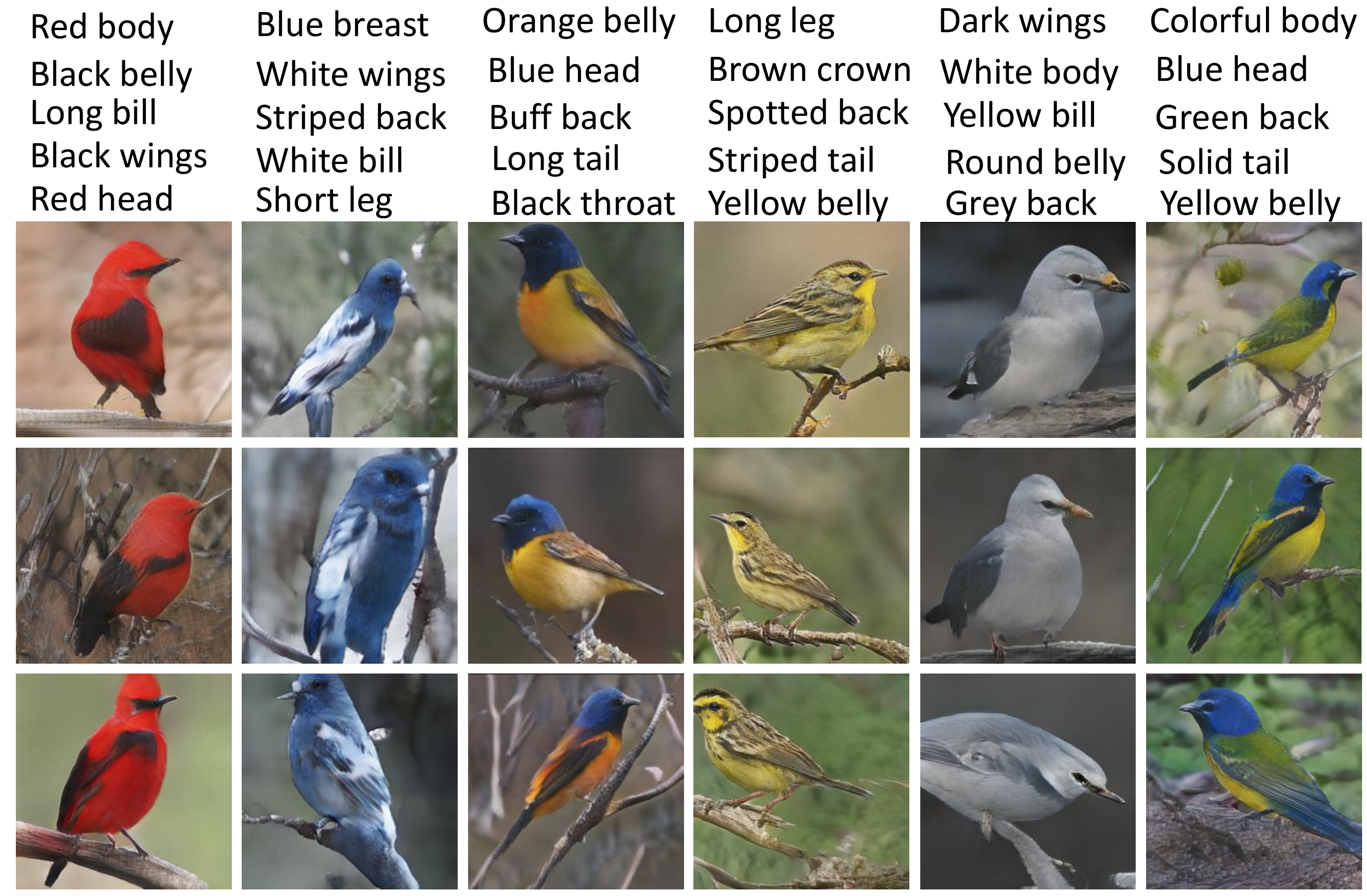}
	\caption{Generated random images by VHE-raster-scan-GAN conditioning on five binary attributes.}\label{Fig: generated by attributes}
\end{figure}

As discussed before, topic models are able to model non-sequential texts such as binary attributes.
The CUB dataset provides 312 binary attributes \citep{WahCUB_200_2011} for each images, such as whether ``crown color is blue'' and whether ``tail shape is solid'' to define the color or shape of different body parts of a bird.
We first transform these binary attributes for the $n$th image to a $312$-dimensional binary vector $\tv_n$, whose $i$th element  is 1 or 0 depending on whether the bird in this image owns the $i$th attribute or not. The binary attribute vectors $\tv_n$ are used together with the corresponding bird images $\xv_n$ to train VHE-raster-scan-GAN.
As shown in Fig.~\ref{Fig: generated by attributes}, we generate images given five binary attributes, which are formed into a $312$-dimensional binary vector $\tv$ (with five non-zero elements at these five attributes) that becomes the input to the PGBN text decoder. %
Clearly, these generated images are photo-realistic and faithfully represent the five provided attributes.

\begin{figure}[ht]
	\centering
	\subfigure[]{\includegraphics[scale=0.435]{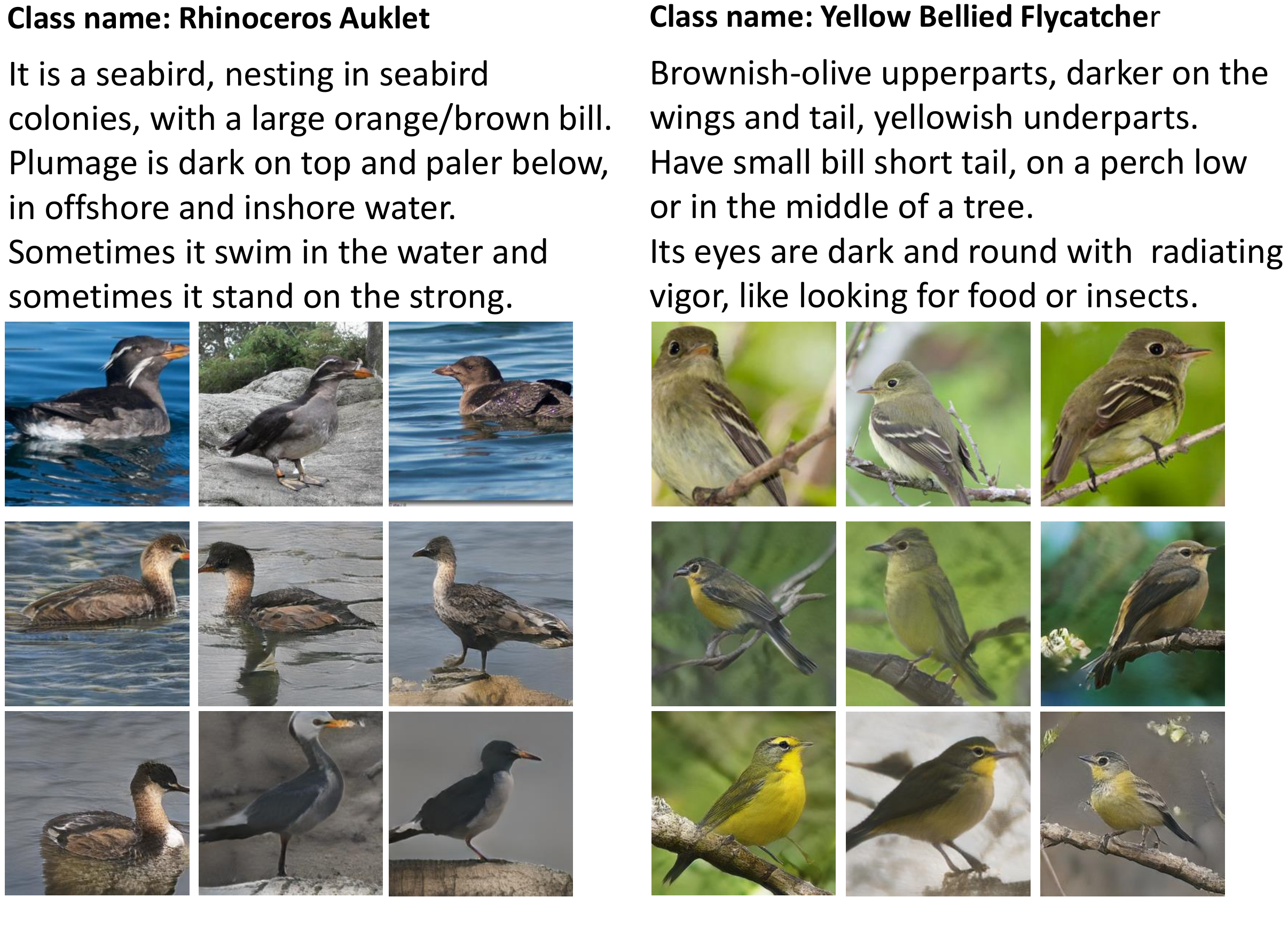}\label{Fig: Generated_by_documents_bird}}
	\quad
	\subfigure[]{\includegraphics[scale=0.435]{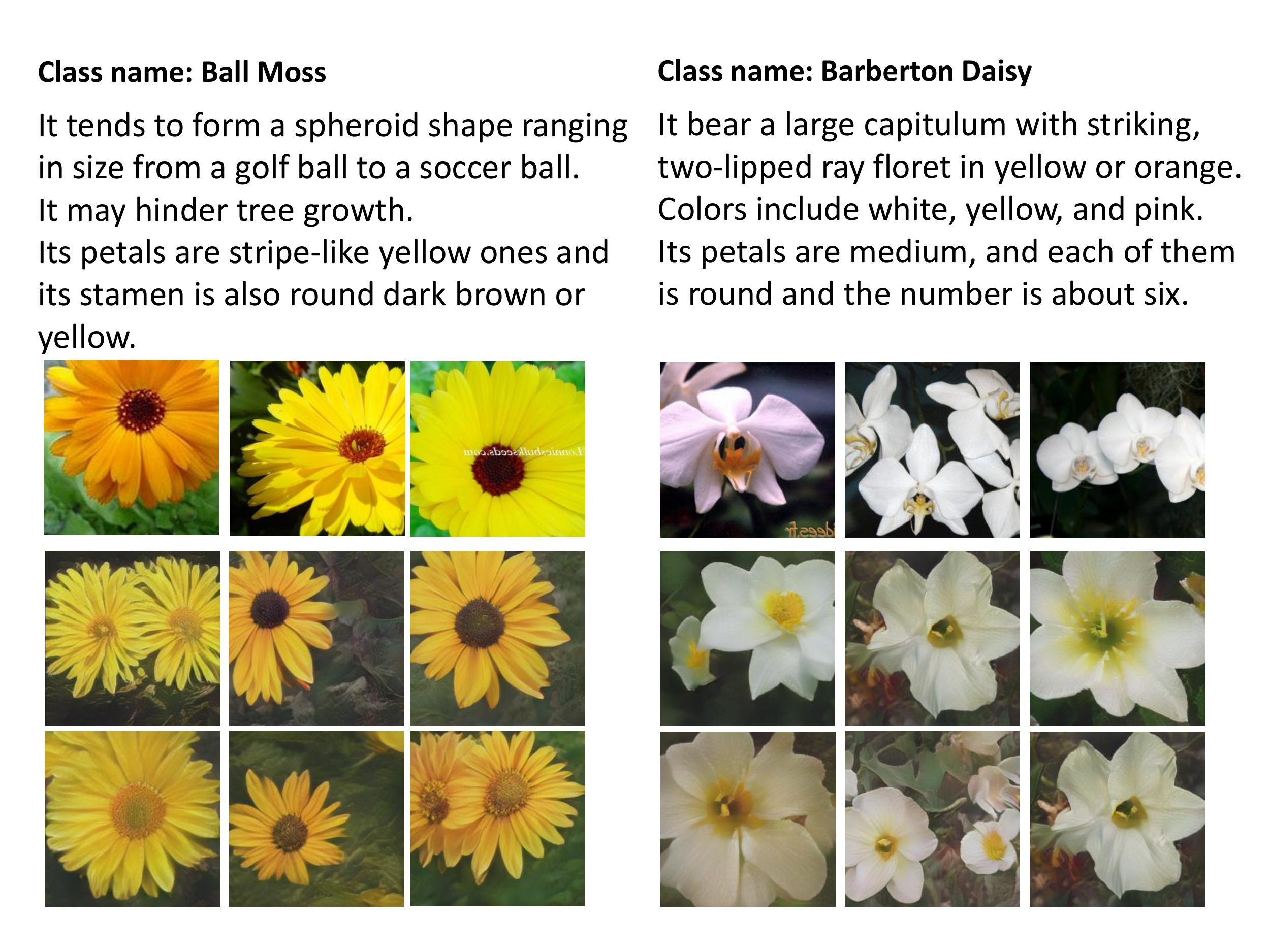}\label{Fig: Generated_by_documents_flower}}
	\caption{%
		Image generation conditioning on long encyclopedia documents using VHE-raster-scan-GAN trained on (a) CUB-E and (b) Flower. Shown in the top part of each subplot are representative sentences taken from the long document that describes an unseen class; for the three rows of images shown in the bottom part, the first row includes three real images from the corresponding unseen class, and the other two rows include a total of six randomly generated images conditioning on the long encyclopedia document of the corresponding unseen class.
	}\label{Fig:image_generate_from_document}
\end{figure}

The proposed VHE-GANs can also well model long documents.
In text-based ZSL discussed in Section 3.2, each class (not each image) is represented as a long encyclopedia document, %
whose global semantic structure is  hard to captured by existing sequence models.
Besides a good ZSL performance achieved by VHE-raster-scan-GAN, illustrating its advantages of text generation given images,
we show Fig.~\ref{Fig:image_generate_from_document} example results of image generation conditioning on long encyclopedia documents on the unseen classes of CUB-E \citep{qiao2016less, akata2015evaluation} and Flower \citep{elhoseiny2017write}.

\begin{figure}[t]
	\centering
	\includegraphics[scale=0.5]{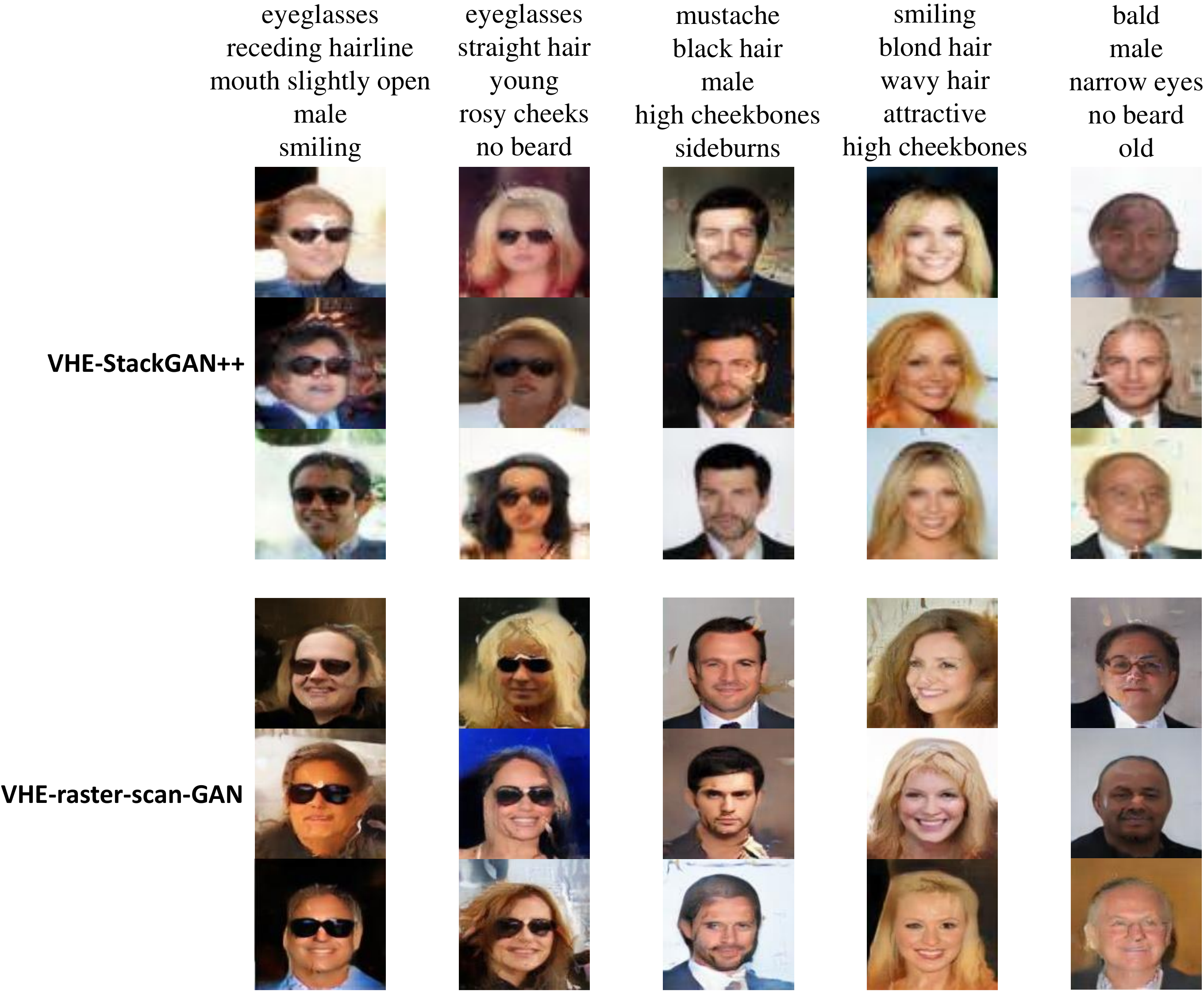}
	\caption{{Example results of facial image generation conditioning on five textual attributes, by VHEStackGAN++ and VHE-raster-scan-GAN trained on the CelebA dataset. Both models are trained with 20 epochs, with the output resolution set as $128 \times 128$. Note our current network architecture, designed mainly for natural images, has not yet been fine-tuned for facial images. 
	} }\label{Fig: face}
\end{figure}

{Analogous to how the Bird images are generated in Fig. \ref{Fig: generated by attributes}, we also perform facial image generation given a set of  textual attributes.
On CelebA dataset, given attributes, we train VHE-stackGAN++ and VHE-raster-scan-GAN to generate the facial images with resolution $128 \times 128$. 
As shown in Fig. \ref{Fig: face}, after the training of 20 epochs, we generate facial images given five attributes.
While the facial images generated by both models nicely match the given attributes, VHE-raster-scan-GAN provides higher visual quality and does a better job in 
representing the details.
}

\clearpage

\section{More experimental results on joint image-text learning}

\subsection{Tables \ref{Tab: IS}  and \ref{Tab: Ablation} with error bars.}
For text-to-image generation tasks, we use the official pre-defined training/testing split (illustrated in Appendix F) to train and test all the models.
Following the definition of error bar of IS in StackGAN++ \citep{Zhang2017StackGAN++}, HDGAN \citep{zhang2018photographic}, and AttnGAN \citep{xu2018attngan},
we provide the IS results with error bars for various methods in Table \ref{Tab: IS with error bar}, where the results of the StackGAN++ , HDGAN, and AttnGAN are quoted from the published papers. The FID error bar is not included as it has not been clearly defined.

\begin{table}[ht]
	\caption{Inception score (IS) results in Table \ref{Tab: IS} with error bars.}
	\centering
	\footnotesize
	\resizebox{1.\textwidth}{!}{
		\begin{tabular}{c|c|c|c|c|c}
			\hline
			Method & StackGAN++ & HDGAN & AttnGAN & {Obj-GAN}  & VHE-raster-scan-GAN \\ \hline
			Flower & 3.26 $\pm$ .01  & 3.45 $\pm$ .07 & --  & - & {\bf{3.72 $\pm$ .01}} \\ \hline
			CUB & 3.84 $\pm$ .06 & 4.15 $\pm$ .05 & 4.36 $\pm$ .03 & - & {\bf{4.41 $\pm$ .03}}  \\ \hline
			COCO & 8.30 $\pm$ .10 &11.86 $\pm$ .18 &25.89 $\pm$ .47 & {26.68 $\pm$ .52}  & {\bf{27.16 $\pm$ .23}}\\ \hline
	\end{tabular}}
	\label{Tab: IS with error bar}
\end{table}


\begin{table}[ht]
	\caption{{Inception score (IS) results in Table \ref{Tab: Ablation} with error bars.}}
	\centering
	\footnotesize
	\resizebox{1.\textwidth}{!}{
		\begin{tabular}{c|c|c|c|c}
			\hline
			Method & PGBN+StackGAN++ & {VHE-vanilla-GAN} & VHE-StackGAN++ &  {VHE-simple-raster-scan-GAN} \\ \hline
			Flower & 3.29 $\pm$ .02  & {3.01 $\pm$ .06} & 3.56 $\pm$ .03  & {3.62  $\pm$ .02}  \\ \hline
			CUB & 3.92 $\pm$ .06 & {3.52 $\pm$ .08} & 4.20 $\pm$ .04 & {4.31 $\pm$ .06} \\ \hline
			COCO & 10.63 $\pm$ .10 & {6.36 $\pm$ .20} & 12.63 $\pm$ .15 & {20.13 $\pm$ 22}  \\ \hline
	\end{tabular}}
	\label{Tab: IS with error bar_ablation}
\end{table}

\clearpage
\subsection{High-quality images of Figure \ref{Fig: text_to_image}}
Due to space constraint, we provide  relative small-size images in Fig.~\ref{Fig: text_to_image}.
Below we show the corresponding images with larger sizes.

\begin{figure}[ht]
	\centering
	\includegraphics[scale=0.75]{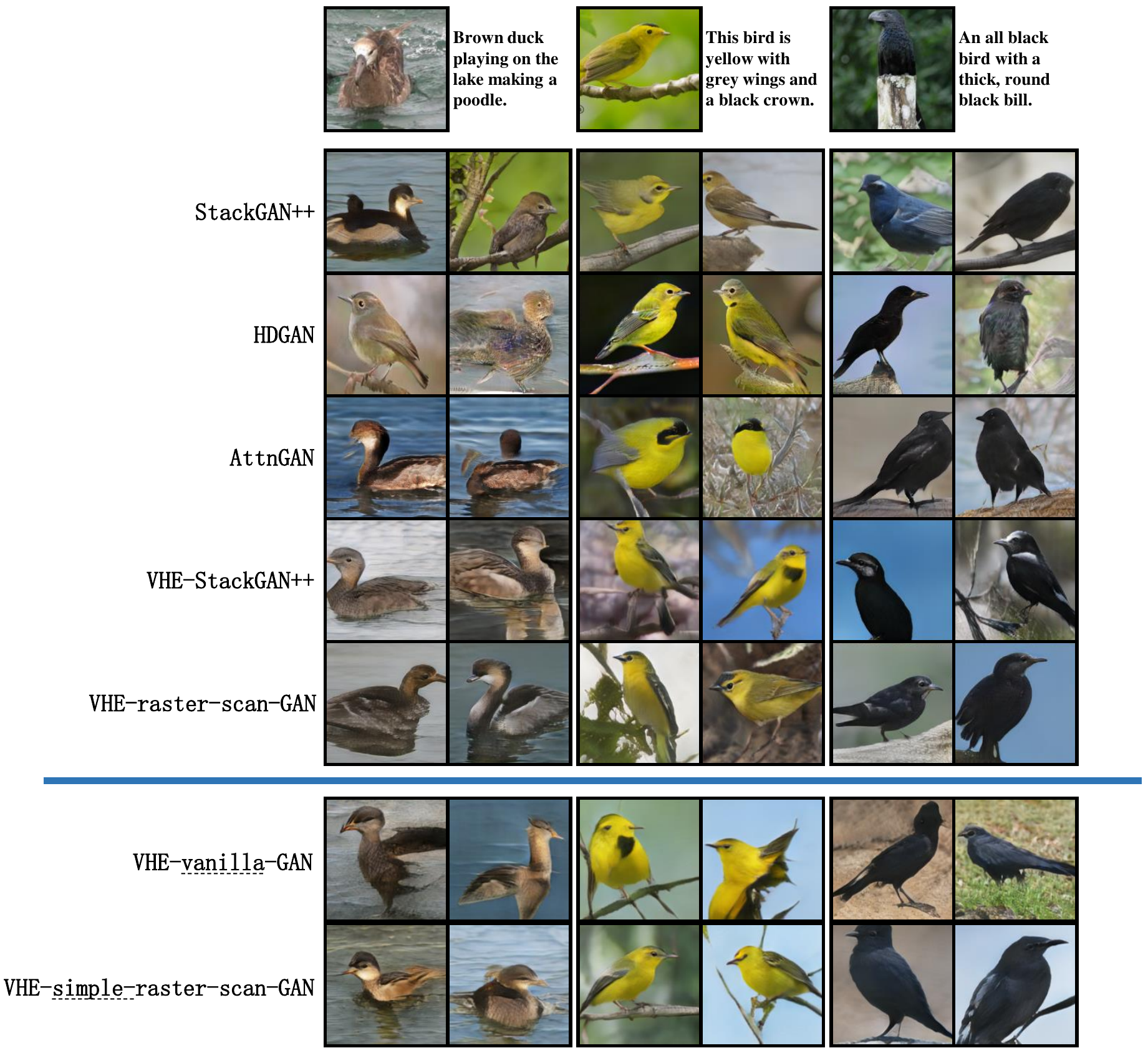}
	\caption{{The images above the blue line are the larger-size replots of CUB Bird images in Figure \ref{Fig: text_to_image}, while the images below the blue line are results for ablation study.} }\label{Fig: img-to-text-part1}
\end{figure}

\begin{figure}[ht]
	\centering
	\includegraphics[scale=0.76]{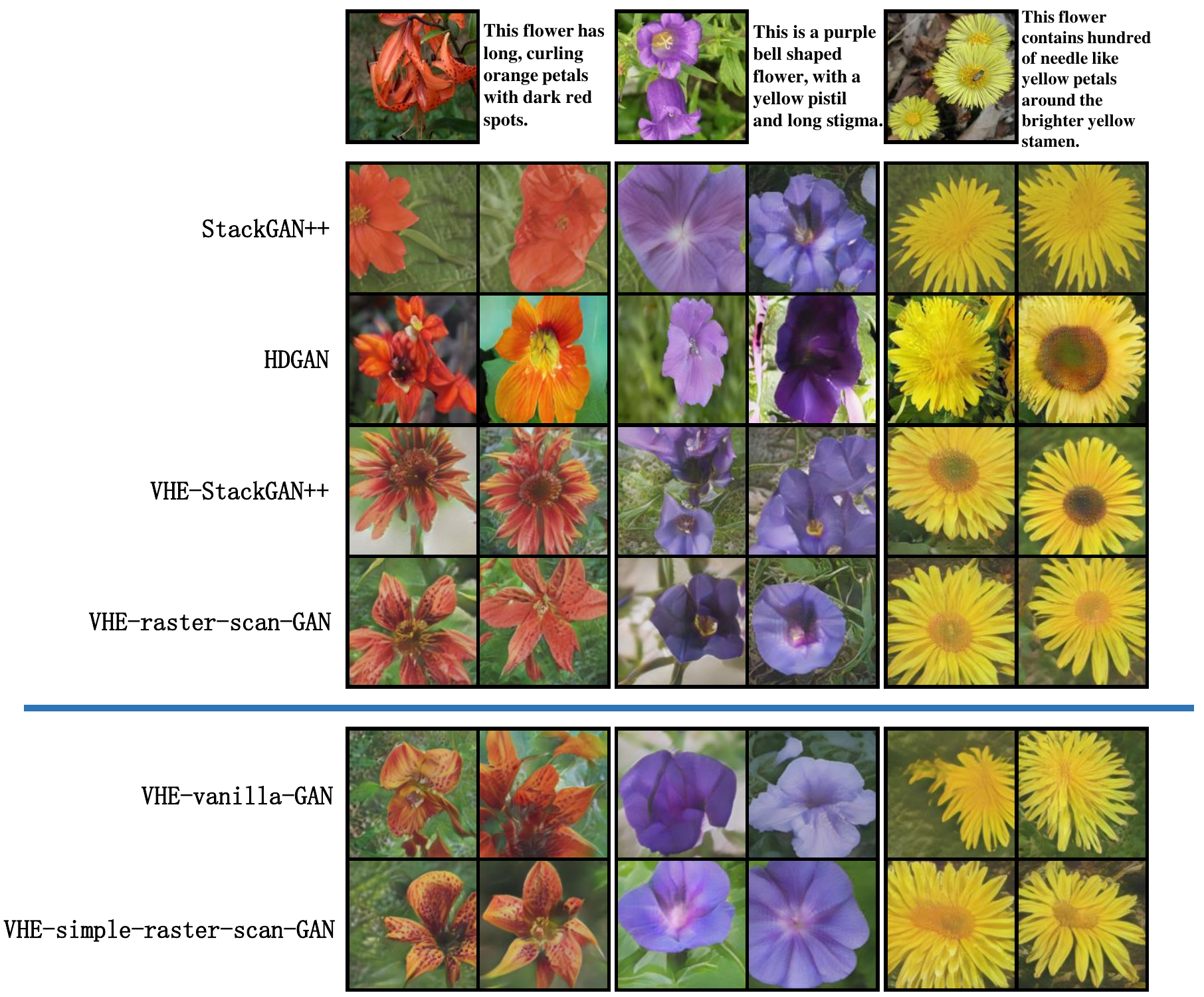}
	\caption{{The images above the blue line are the larger-size replots of Flower images in Figure \ref{Fig: text_to_image}, while the images below the blue line are results for ablation study.}}\label{Fig: img-to-text-part2}
\end{figure}

\begin{figure}[ht]
	\centering
	\includegraphics[scale=1.]{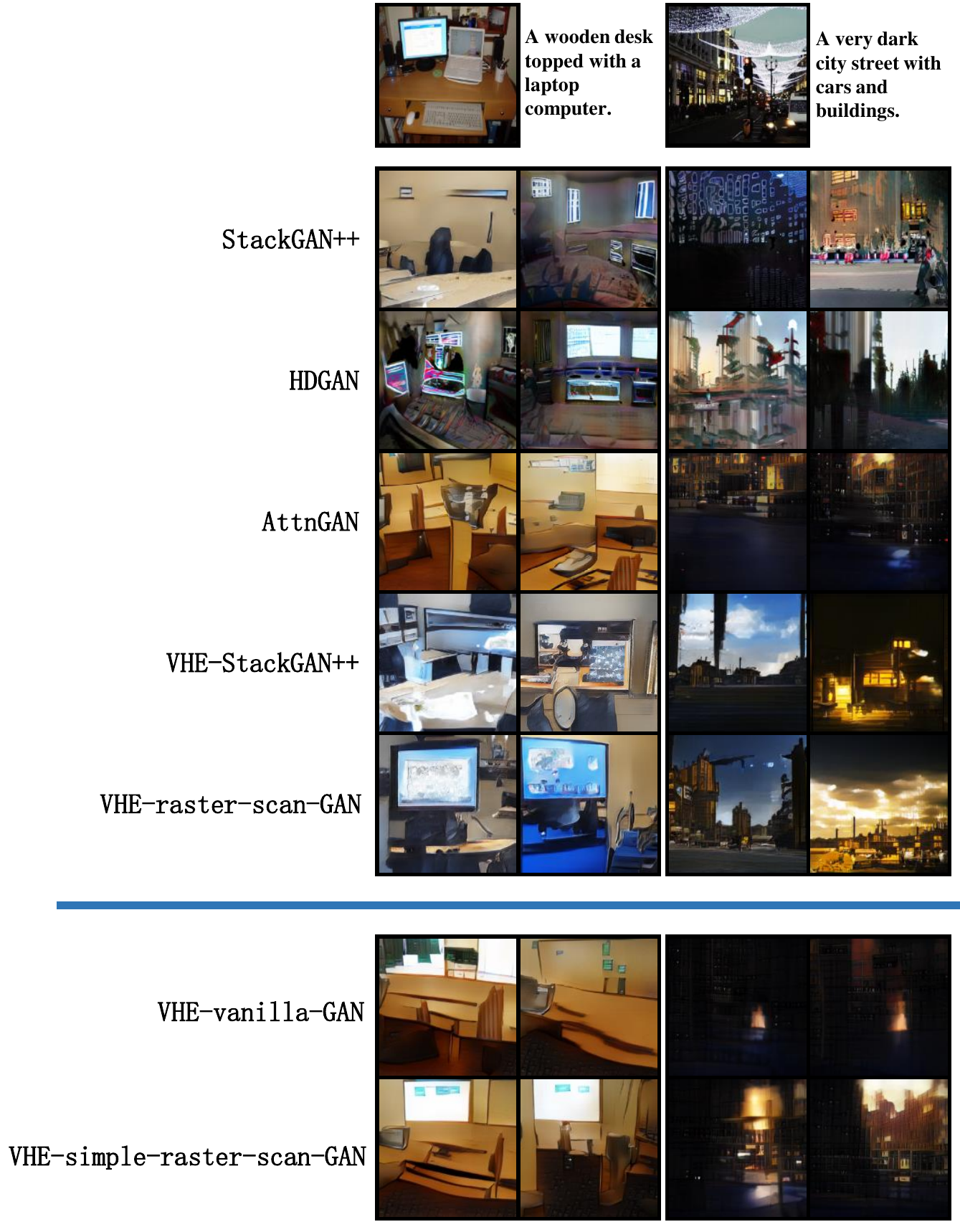}
	\caption{{The images above the blue line are the larger-size replots of COCO images in Figure \ref{Fig: text_to_image}, while the images below the blue line are results for ablation study.}}\label{Fig: img-to-text-part3}
\end{figure}

\clearpage
\subsection{More text-to-image generation results on COCO}
\begin{figure}[ht]
	\centering
	\includegraphics[scale=0.28]{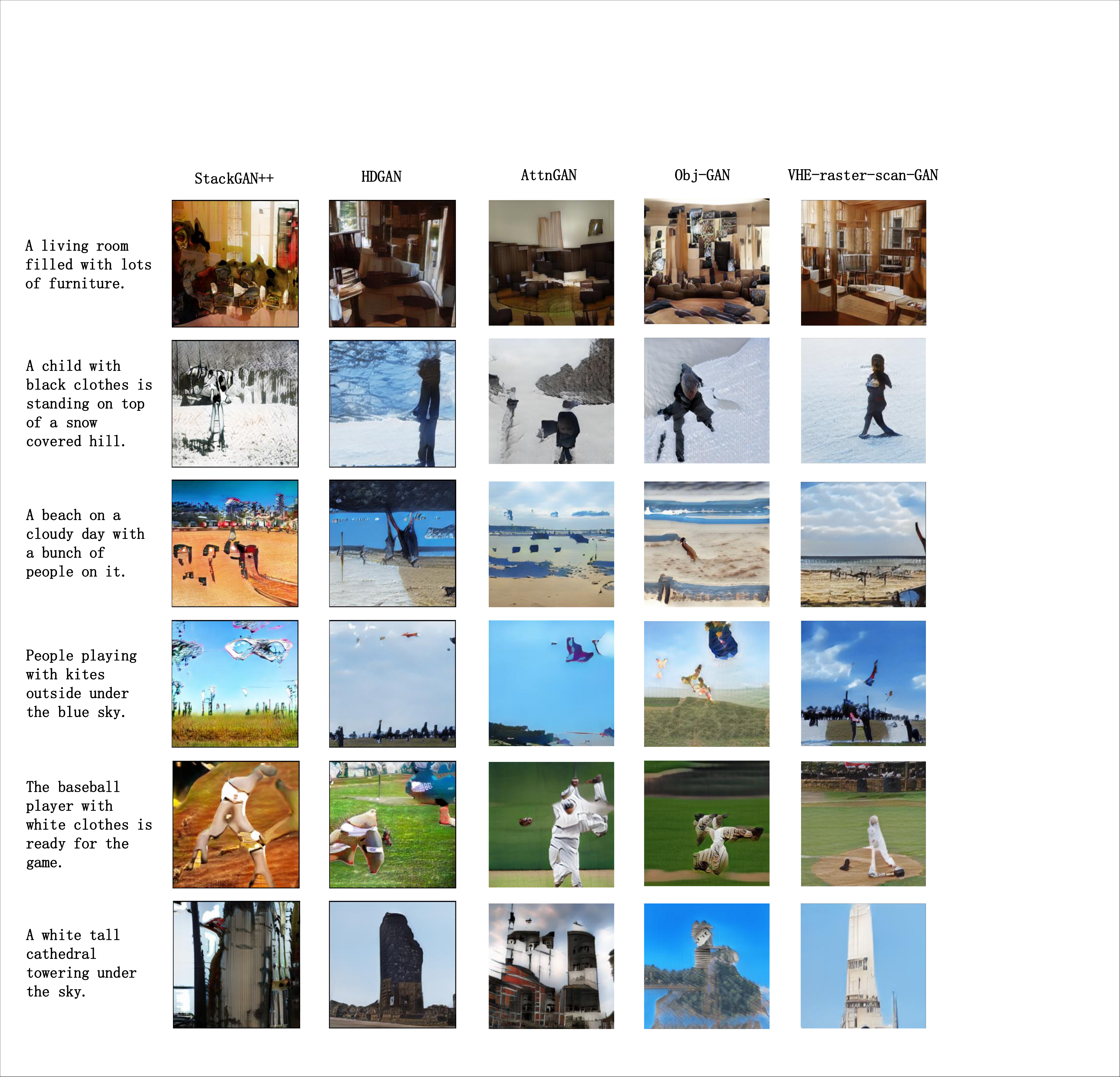}
	\caption{{Example text-to-image generation results on COCO.}}\label{Fig: coco-another-text-to-image}
\end{figure}

COCO is a more challenging dataset than CUB and Flower, as it contains very diverse objects and scenes.
We show in Fig. \ref{Fig: coco-another-text-to-image} more samples conditioned on different textural descriptions.

\clearpage

\subsection{Latent space interpolation}
\begin{figure}[h]
	\centering
	\includegraphics[scale=0.2]{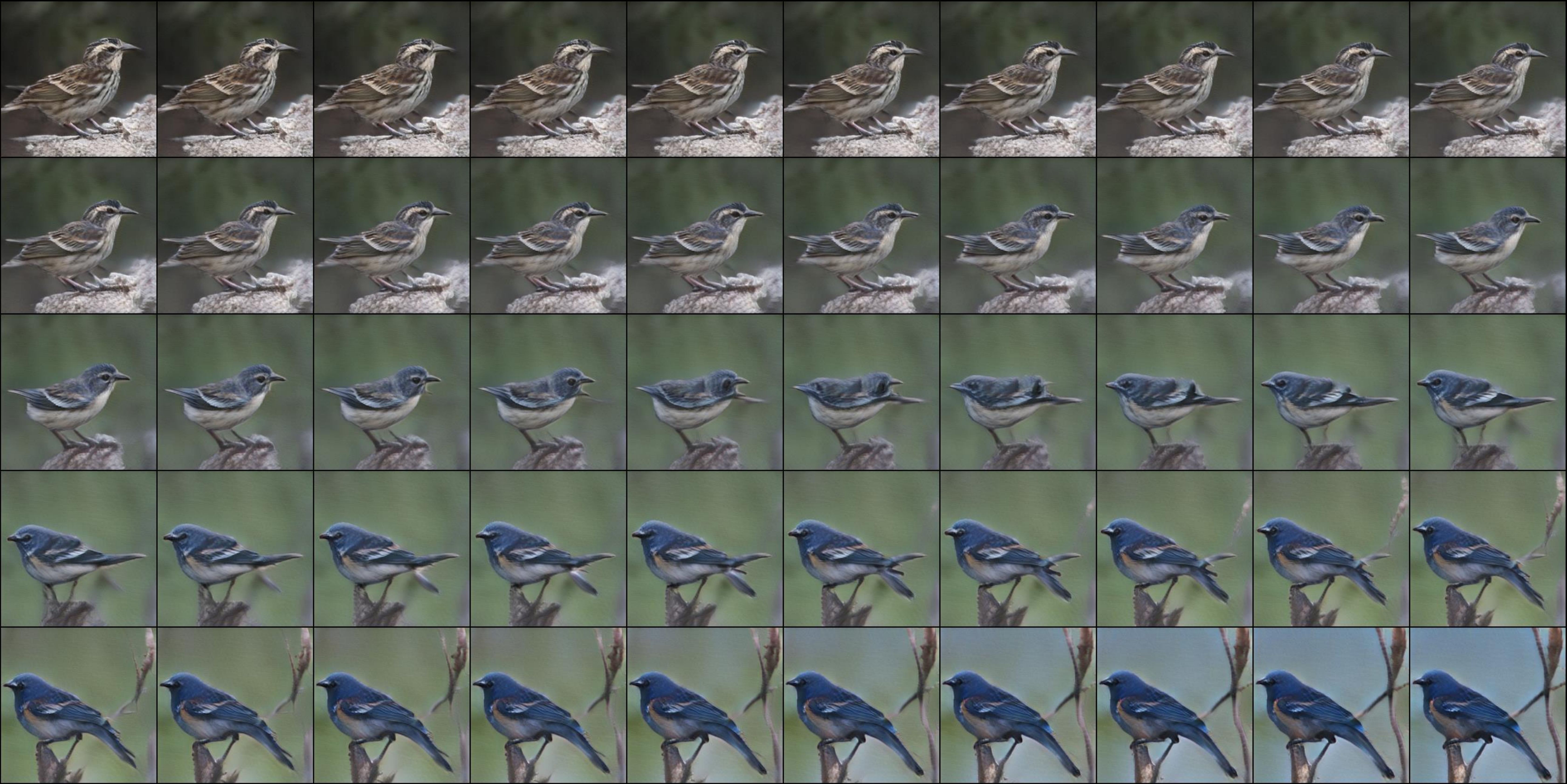}
	\vspace{-1mm}
	\caption{\small Example of latent space interpolation on CUB.}\label{Fig: appendix_interpolate_bird2}
	\vspace{5mm}
	\centering
	\includegraphics[scale=0.2]{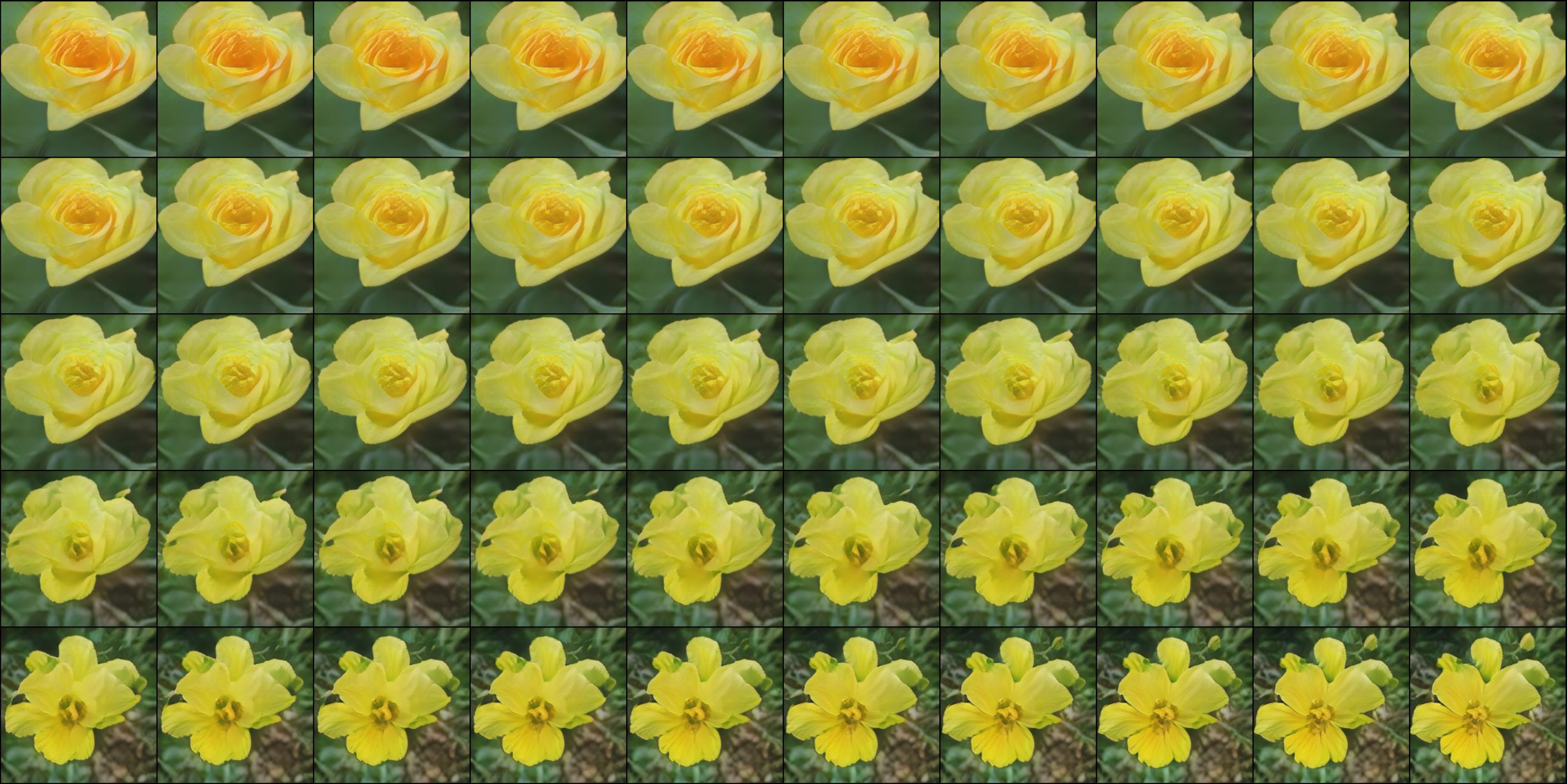}
	\vspace{-1mm}
	\caption{\small Example of latent space interpolation on Flower.}\label{Fig: appendix_interpolate_flower1}
	\vspace{5mm}
	\centering
	\includegraphics[scale=0.2]{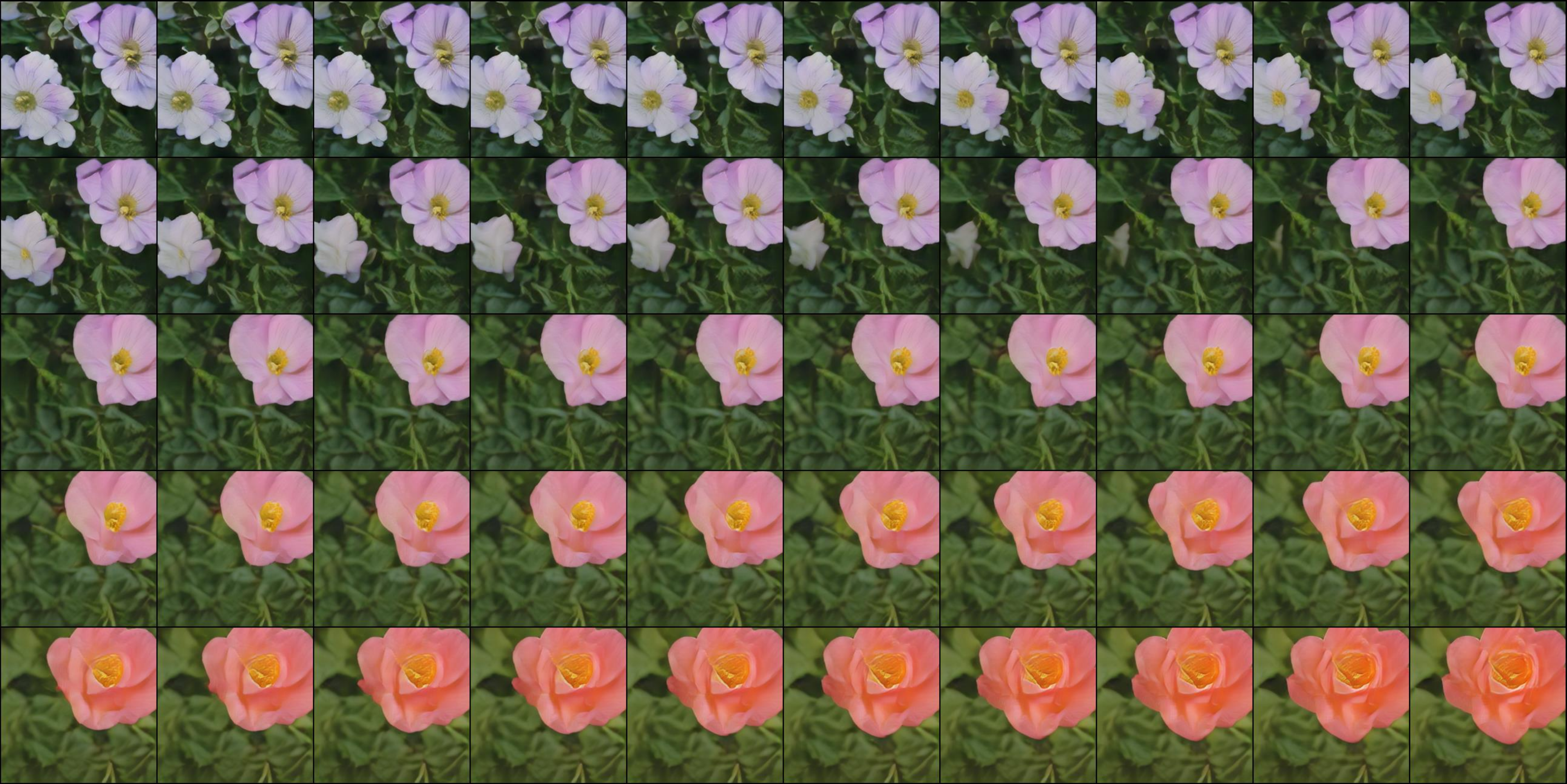}
	\vspace{-1mm}
	\caption{\small Example of latent space interpolation on Flower.}\label{Fig: appendix_interpolate_flower2}
\end{figure}

In addition to the latent space interpolation results of VHE-raster-scan-GAN in Fig. \ref{Fig: interpolate} of Section 3.1, below we provide more fine-gridded latent space interpolation in Figs. \ref{Fig: appendix_interpolate_bird1}-%
\ref{Fig: appendix_interpolate_flower2}.

\begin{figure}[h]
	\centering
	\includegraphics[scale=0.2]{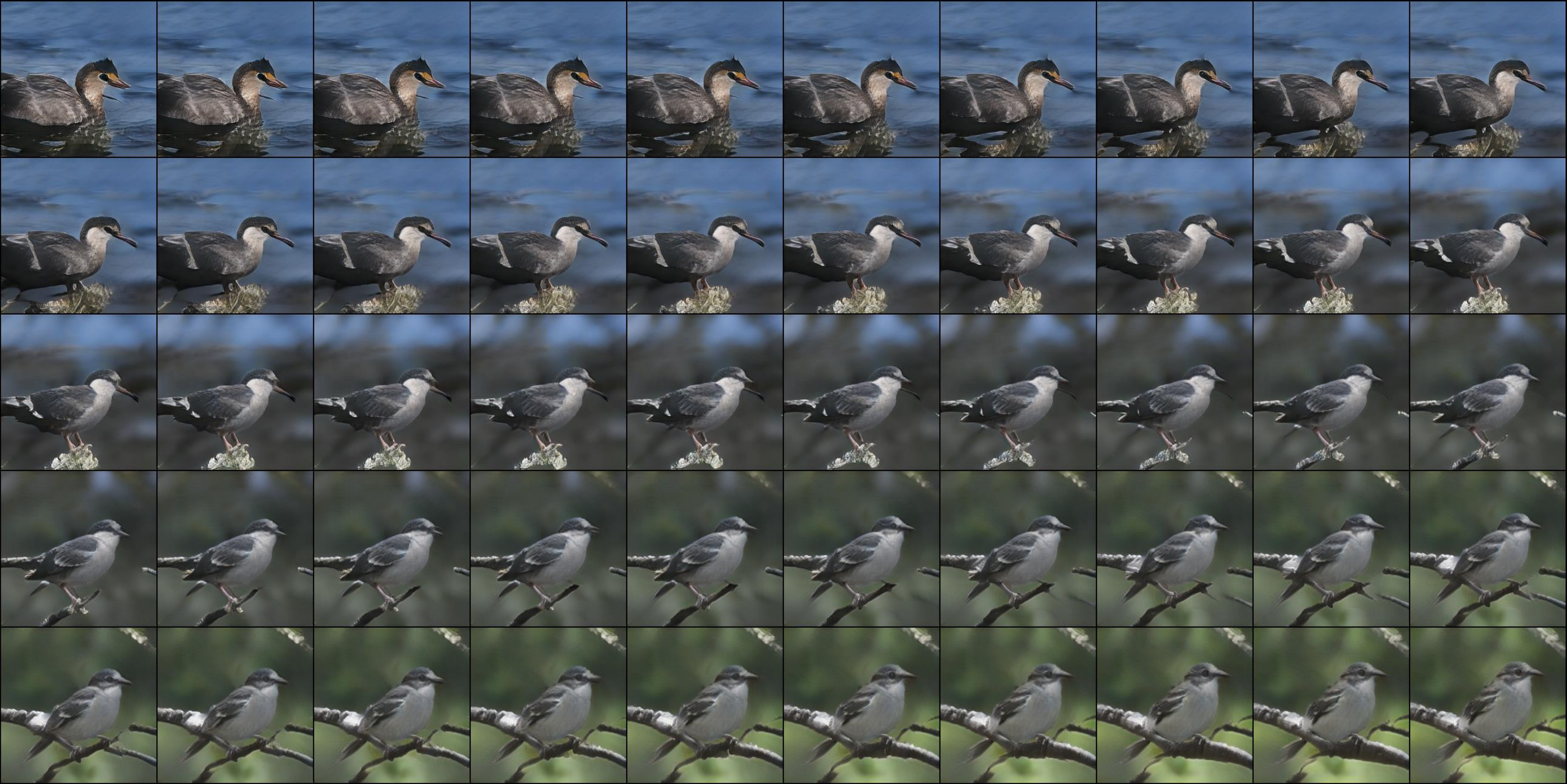}\\
	\caption{\small Example of latent space interpolation on CUB.}\label{Fig: appendix_interpolate_bird1}
\end{figure}

\begin{figure}[h]
	\centering
	\includegraphics[scale=0.2]{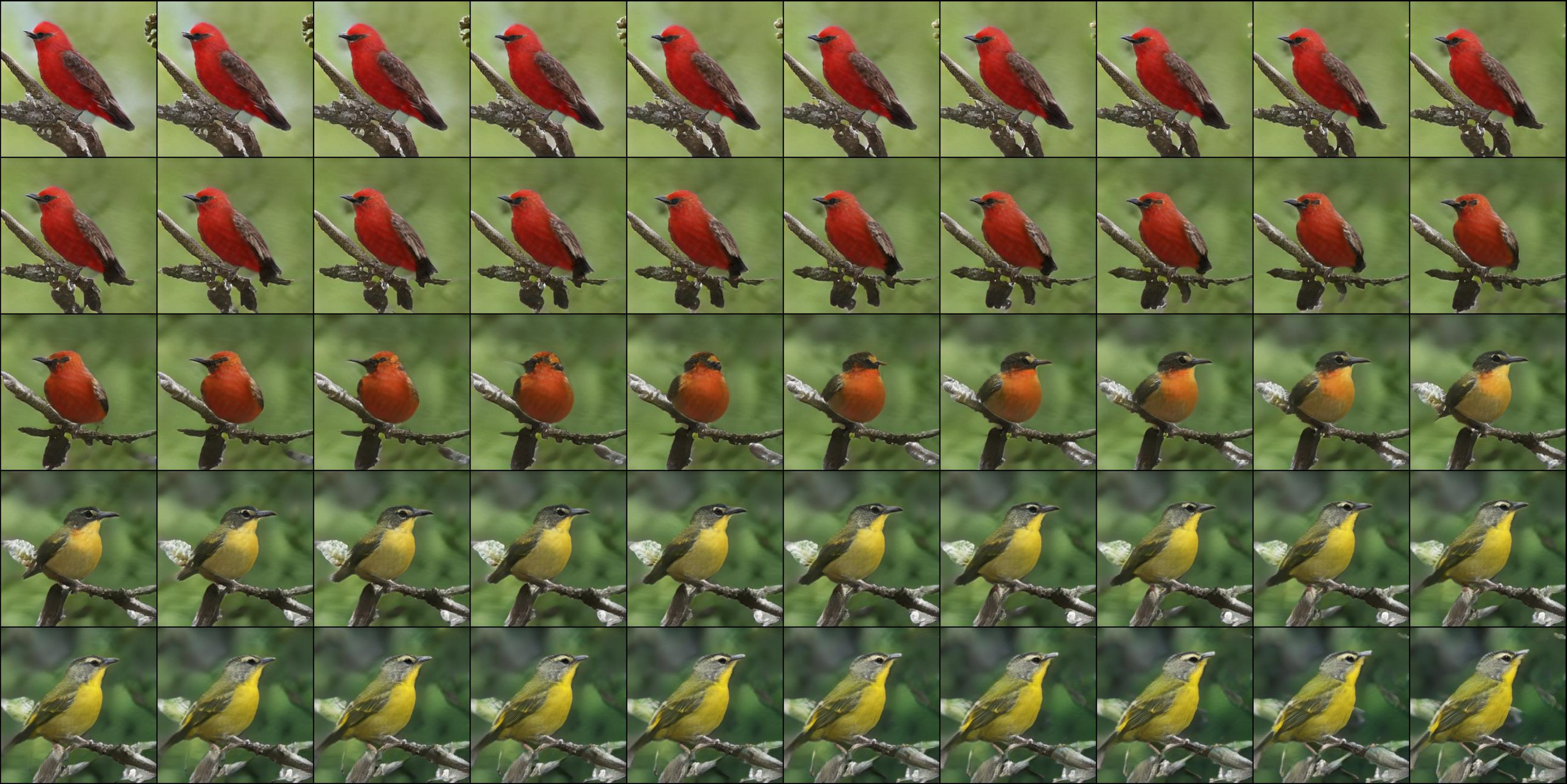}\\
	\caption{\small Example of latent space interpolation on CUB.}\label{Fig: appendix_interpolate_bird1}
\end{figure}

\clearpage
\subsection{Image retrieval given a text query}
\begin{figure}[!h]
	\centering
	\includegraphics[scale=0.34]{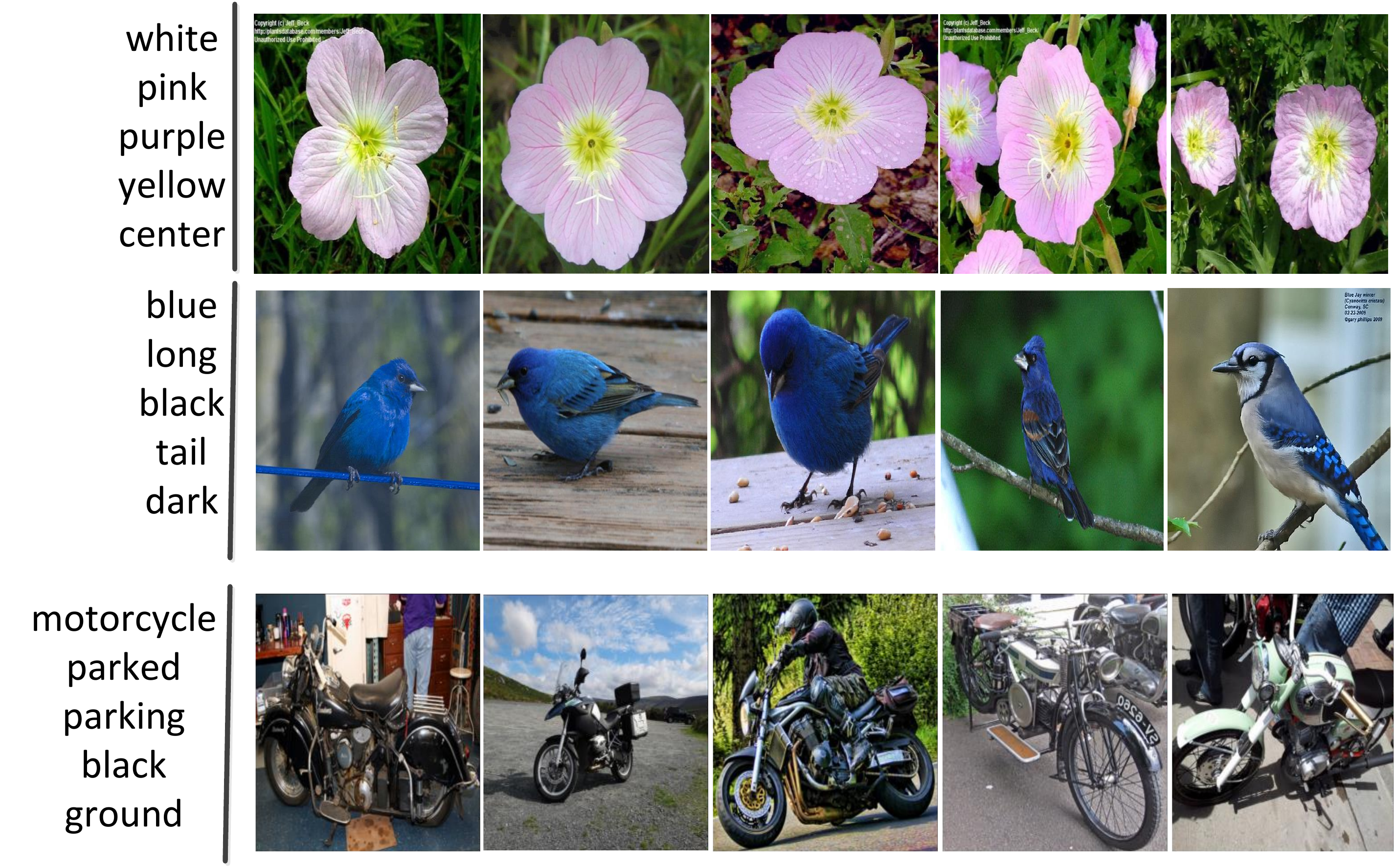}\\
	\caption{\small \small Top-5 retrieved images given a text query. Rows 1 to 3 are for Flower, CUB, and COCO, respectively.} %
	\label{Fig: image_retrival}
\end{figure}

For image $\xv_n$, we draw its BoW textual description $\hat{\tv}_n$ as %
$
\hat{\tv}_n\given \thetav_n\sim p(\tv\given \Phimat,\thetav_n),~\thetav_n\given \xv_n\sim q_{\Omegamat}(\thetav\given \Phimat,\xv_n)
$.
Given the BoW textual description $\tv$ as a text query, we retrieve the top five images ranked by the cosine distances between $\tv$ and $\hat{\tv}_n$'s.
Shown in Fig. \ref{Fig: image_retrival} are three example image retrieval
results, which suggest that the retrieved images are semantically related to their text queries in
colors, shapes, and locations.

\subsection{Image regeneration}
\begin{figure*}[ht]
	\centering
	\includegraphics[scale=0.155]{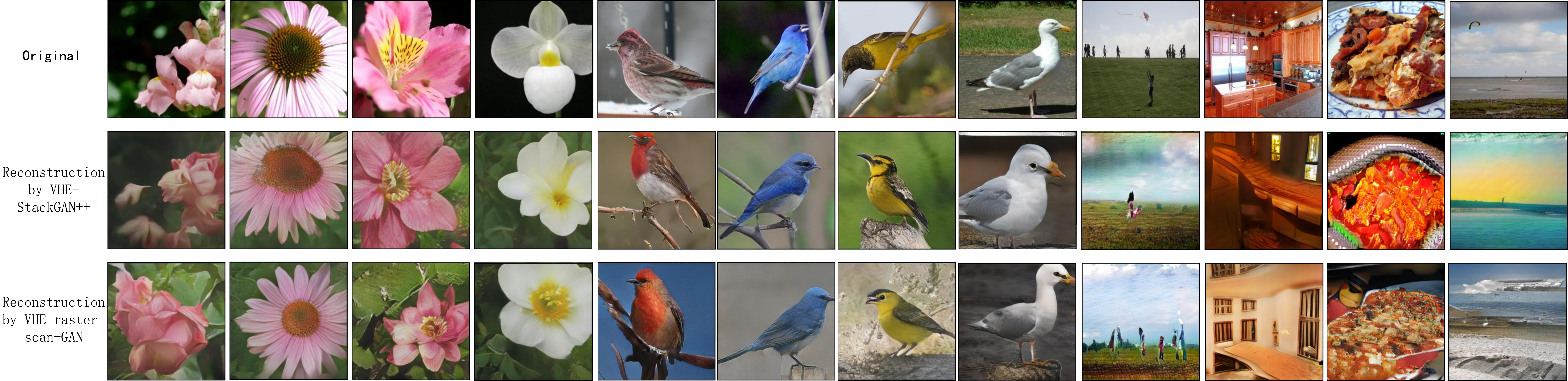}
	\caption{\small Example results of image regeneration using VHE-StackGAN++ and VHE-raster-scan-GAN. An original image is fed into the VHE image encoder, whose latent representation is then fed into the GAN image generator to generate a corresponding random image. The models in columns 1-4  are trained on Flower, columns 5-8 on CUB, and columns 9-12 on COCO.}
	\label{Fig: image_recon}
\end{figure*}

We note for VHE-GAN, %
its image encoder and GAN component together can also be viewed as an ``autoencoding'' GAN for images. More specifically, given image $\xv$, VHE-GAN can provide random regenerations using $G\left( q_{\Omegamat}\left( \thetav \given \Phimat, \xv\right) \right)$.
We show example image regeneration results by both VHE-StackGAN++ and VHE-raster-scan-GAN in  Fig.~\ref{Fig: image_recon}. These example results suggest  that the regenerated random images by the proposed VHE-GANs more of less
resemble
the original real image  fed into the VHE image encoder.

\subsection{Learned hierarchical topics in VHE}
\begin{figure}[h]
	\centering
	\includegraphics[scale=0.23]{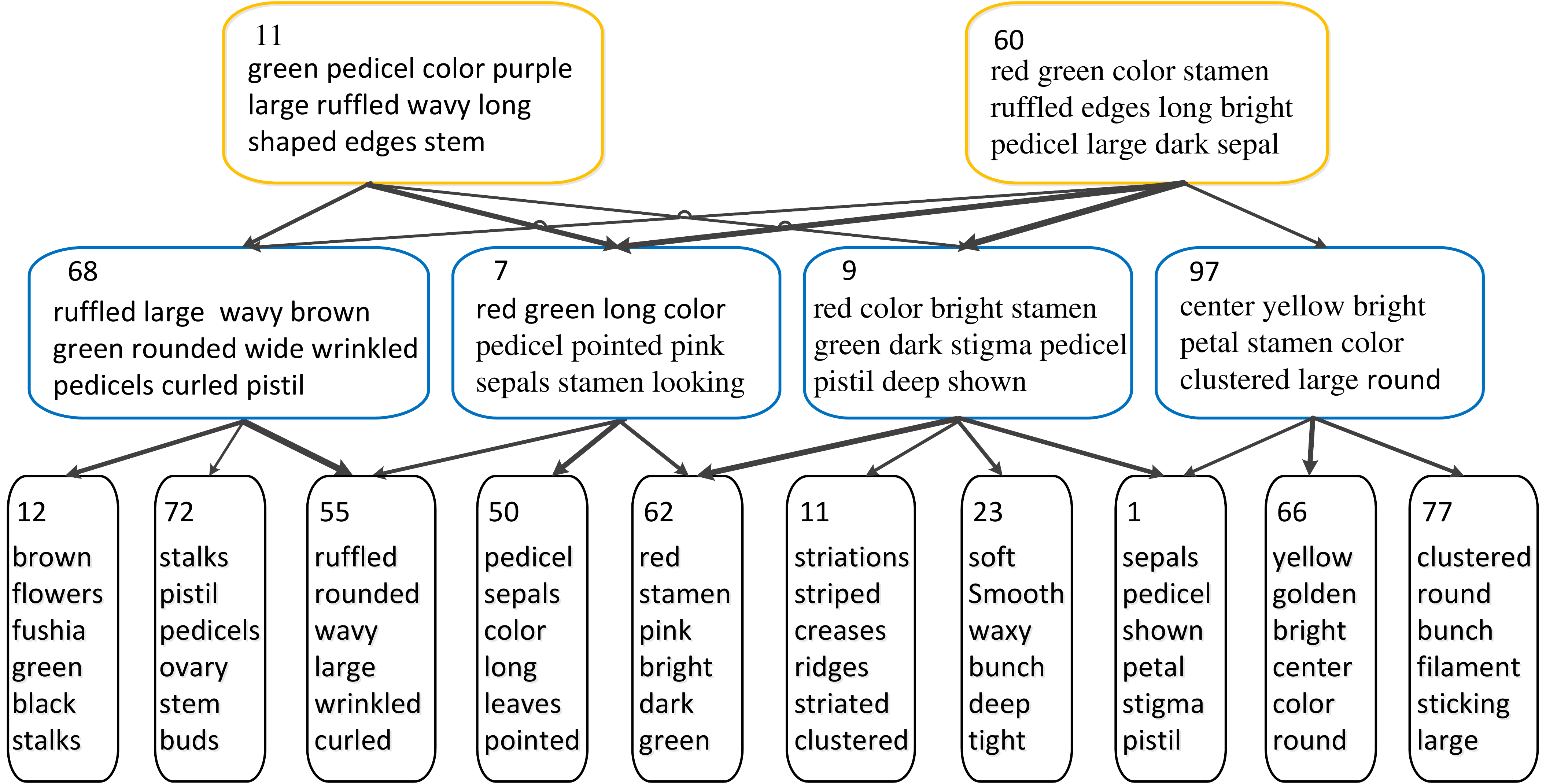}\\
	\caption{\small An example topic hierarchy taken from the directed acyclic graph learned by a three-layer VHE-raster-scan-GAN of size 256-128-64 on Flower.}\label{fig: flower_text_topic}
\end{figure}

\begin{figure}[h]
	\centering
	\includegraphics[scale=0.23]{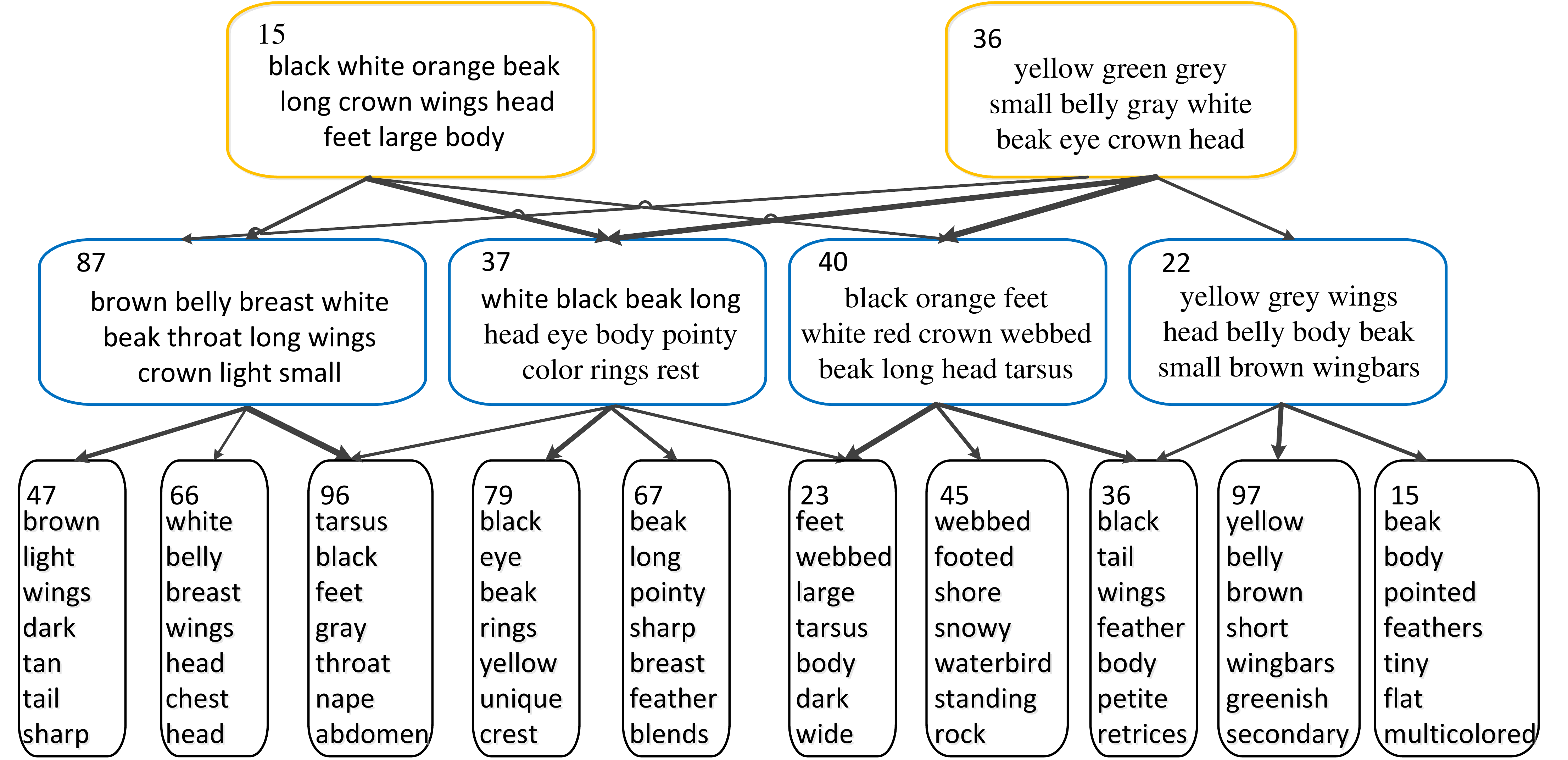}\\
	\caption{\small Analogous plot to Fig. \ref{fig: flower_text_topic} on CUB.}\label{fig: bird_text_topic}
\end{figure}

\begin{figure}[h]
	\centering
	\includegraphics[scale=0.25]{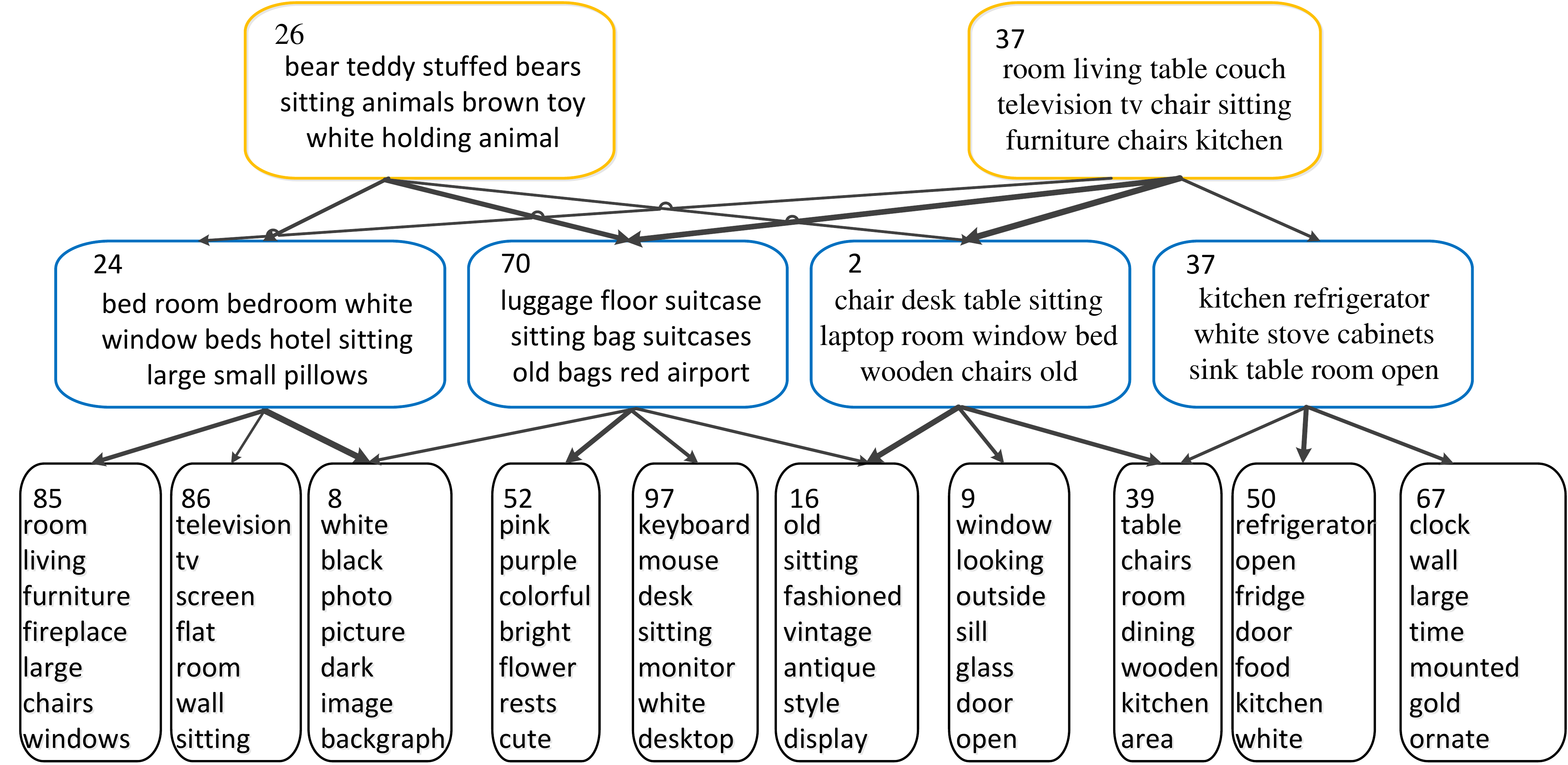}\\
	\caption{\small  Analogous plot to Fig. \ref{fig: flower_text_topic} on COCO.}\label{fig: coco_text_topic}
\end{figure}

The inferred topics at different layers and the inferred sparse connection weights between the topics of adjacent layers are found to be highly interpretable.
In particular, we can understand the meaning of each topic by projecting it back to the original data space via
$\left[ \prod_{t=1}^{l-1} \Phimat^{(t)} \right] \phiv_k^{(l)}$ and understand the relationship between the topics by arranging them into a directed acyclic graph (DAG) and choose its subnets to visualize.
We show in Figs. \ref{fig: flower_text_topic}, \ref{fig: bird_text_topic}, and
\ref{fig: coco_text_topic} example subnets taken from the DAGs inferred by the three-layer VHE-raster-scan-GAN of size 256-128-64 on Flower, CUB, and COCO, respectively.
The semantic meaning of each topic and the connection weights between the topics of adjacent layers are highly interpretable. For example,  in Figs. \ref{fig: flower_text_topic},  the topics describe very specific flower characteristics, such as special colors, textures, shapes, and parts, at the bottom layer, and become increasingly more general when moving upwards.

\clearpage
\section{Specific model structure in VHE-StackGAN++ and VHE-raster-scan-GAN}

\subsection{Model structure of VHE}
\begin{figure}[ht]
	\centering
	\includegraphics[scale=0.4]{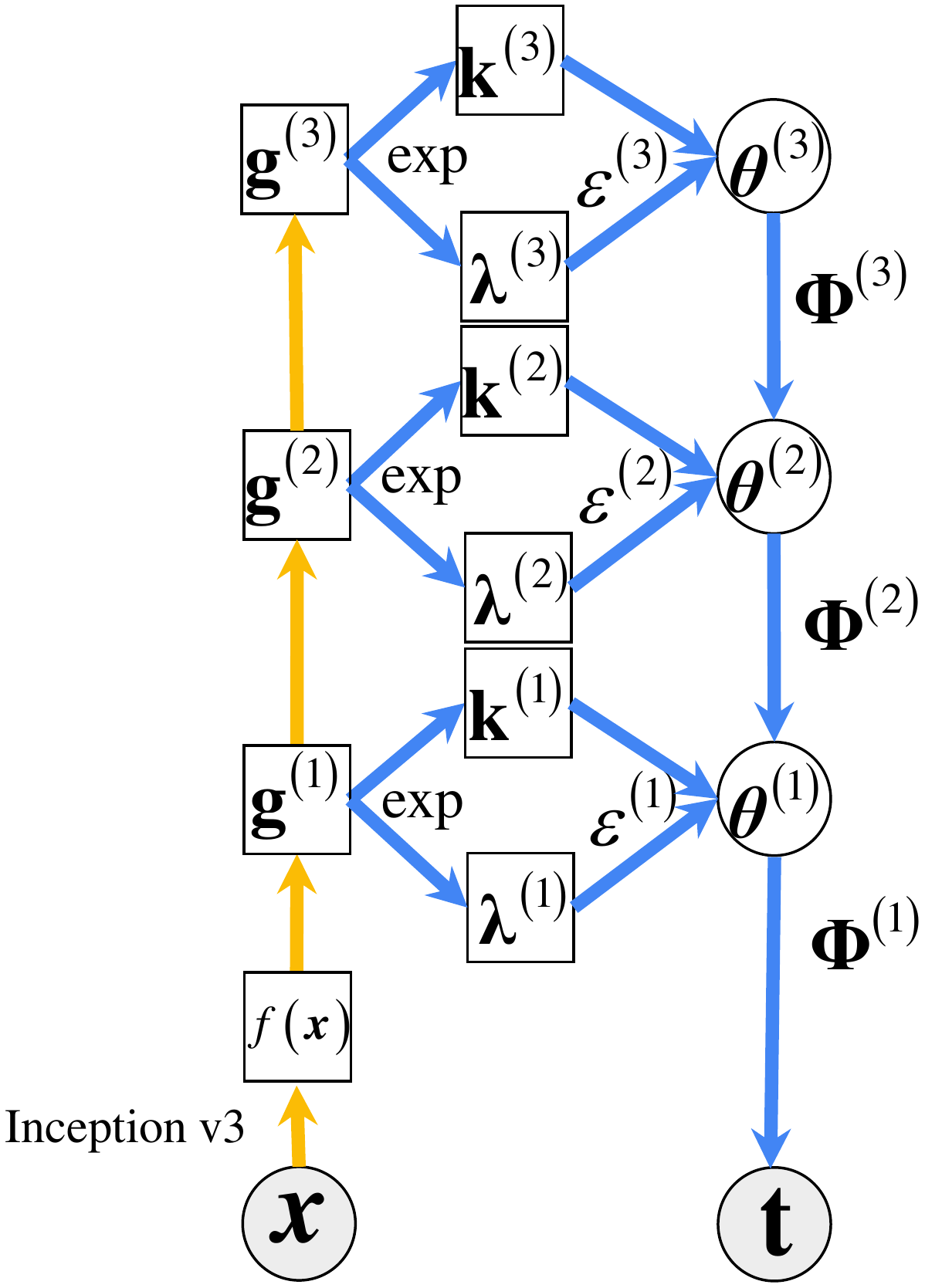}\\
	\caption{The architecture of VHE in VHE-StackGAN++ and VHE-raster-scan-GAN.}
	\label{Fig: appendix_VHAE}
\end{figure}

In Fig. \ref{Fig: appendix_VHAE}, we give the structure of VHE used in VHE-StackGAN++ and VHE-raster-scan-GAN, where $f(\xv)$ is the image features extracted by Inception v3 network and $\varepsilonv^{(l)} \sim \prod_{k=1}^{K_l}\mbox{Uniform}(\varepsilon_k^{(l)};0,1)$.
With the definition of $\gv^{(0)} = f(\xv)$, we have
\begin{align}
&\kv^{(l)} = \exp(\Wmat_1^{(l)}\gv^{(l)}+ \bv_1^{(l)}), \\
&\lambdav^{(l)} = \exp(\Wmat_2^{(l)}\gv^{(l)}+ \bv_2^{(l)}), \label{MLP2}\\
&\gv^{(l)} = \ln[1+\exp(\Wmat_3^{(l)}\gv^{(l-1)}+\bv_3^{(l)})],%
\end{align}
where $\Wmat_1^{(l)}\in\mathbb{R}^{K_l\times K_{l}}$, $\Wmat_2^{(l)}\in\mathbb{R}^{K_l\times K_{l}}$, $\Wmat_3^{(l)}\in\mathbb{R}^{K_l\times K_{l-1}}$, $\bv_1^{(l)}\in\mathbb{R}^{K_l}$, $\bv_2^{(l)}\in\mathbb{R}^{K_l}$, and $\bv_3^{(l)}\in\mathbb{R}^{K_l}$.

\subsection{Model of VHE-StackGAN++}
\begin{figure}[!h]
	\centering
	\includegraphics[scale=0.52]{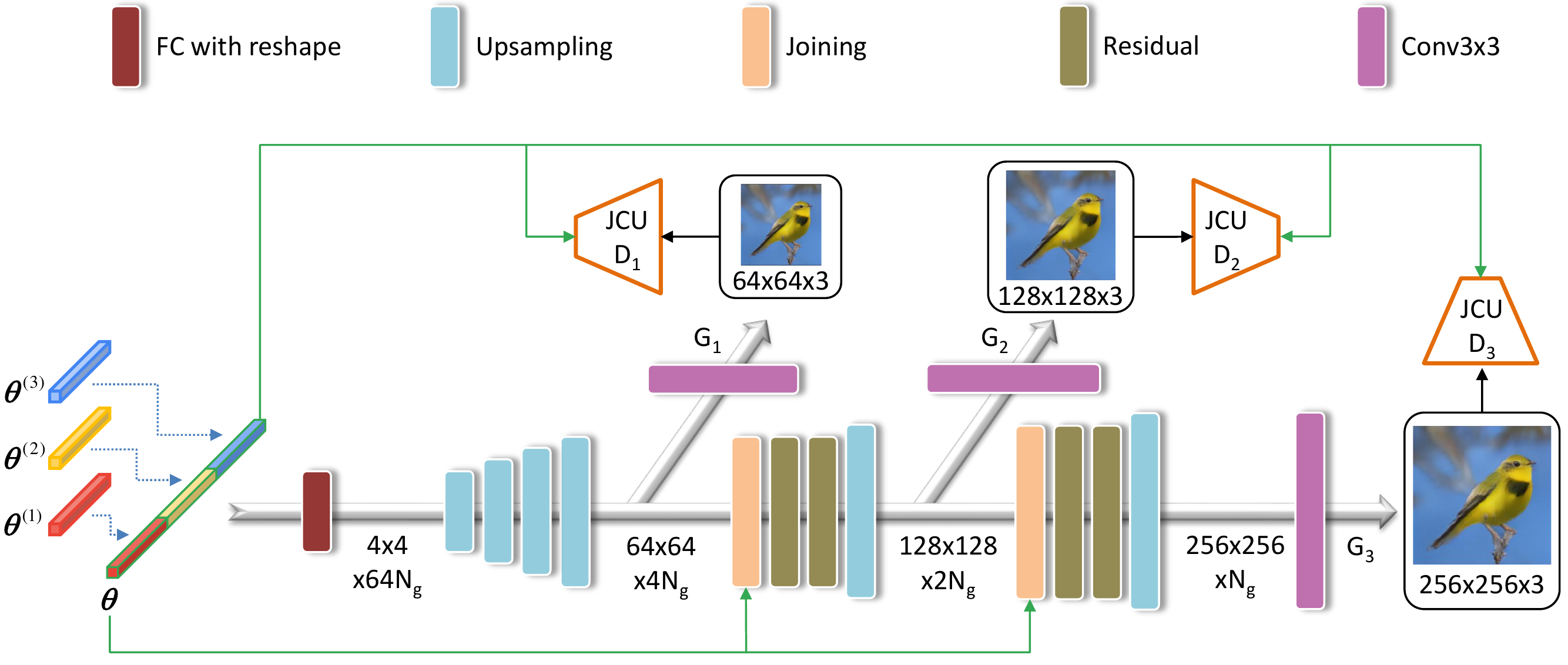}\\
	\caption{The structure of Stack-GAN++ in VHE-StackGAN++, where JCU is a type of discriminator proposed in \citet{Zhang2017StackGAN++}.}\label{Fig: appendix_StackGAN}
\end{figure}

In Section 2.2, we first introduce the VHE-StackGAN++, where the multi-layer textual representation $\{ \thetav^{(1)}, \thetav^{(2)}, \cdots, \thetav^{(L)} \}$ is concatenated as $\thetav = \left[ \thetav^{(1)}, \cdots, \thetav^{(L)} \right]$ and then fed into StackGAN++ \citep{Zhang2017StackGAN++}.
In Figs. \ref{Fig: VHEGAN} (a) and (b), we provide the model structure of VHE-StackGAN++. We also provide a detailed plot of the structure of StackGAN++ used in VHE-StackGAN++ in Fig. \ref{Fig: appendix_StackGAN}, where JCU is a specific  type of discriminator; %
see \citet{Zhang2017StackGAN++} for more details.

The same with VHE-raster-scan-GAN, VHE-StackGAN++ is also able to jointly optimize all components by merging the expectation in VHE and GAN to define its loss function as
\begin{align}\label{Eq: mvrARGAN-V1}
&\textstyle \min_{\Omegamat,\{G_i\}_{i=1}^3} \max_{\{D_i\}_{i=1}^3}
\mathbb{E}_{p_{\text{data}}(\xv_n, \tv_n)} \mathbb{E}_{\prod_{l=1}^L q(\thetav_n^{(l)}\given \xv_n, \Phimat^{(1+1)}, \thetav_n^{(l+1)})} \big\{-\log p(\tv_n\given \Phimat^{(1)}, \thetav_n^{(1)}) \nonumber\\
&~~~~~~~~~~~~~~ ~~~~~~~~~~~~~~\textstyle + \sum_{l=1}^{L} \mbox{KL}[q(\thetav_n^{(l)}\given\xv_n, \Phimat^{(1+1)}, \thetav_n^{(l+1)})\, ||\, p(\thetav_n^{(l)}\given \Phimat^{(1+1)}, \thetav_n^{(l+1)})] \nonumber\\
&~~~~~~~~~~~~~~ ~~~~~~~~~~~~~~\textstyle + \sum_{i=1}^3 [\log D_i (\xv_{n,i}, \thetav_n) + \log (1-D_i(G_i(\thetav_n),\thetav_n))]\big \}.
\end{align}

\clearpage

\subsection{Structure of raster-scan-GAN}

\begin{figure}[!h]
	\centering
	\includegraphics[scale=0.44]{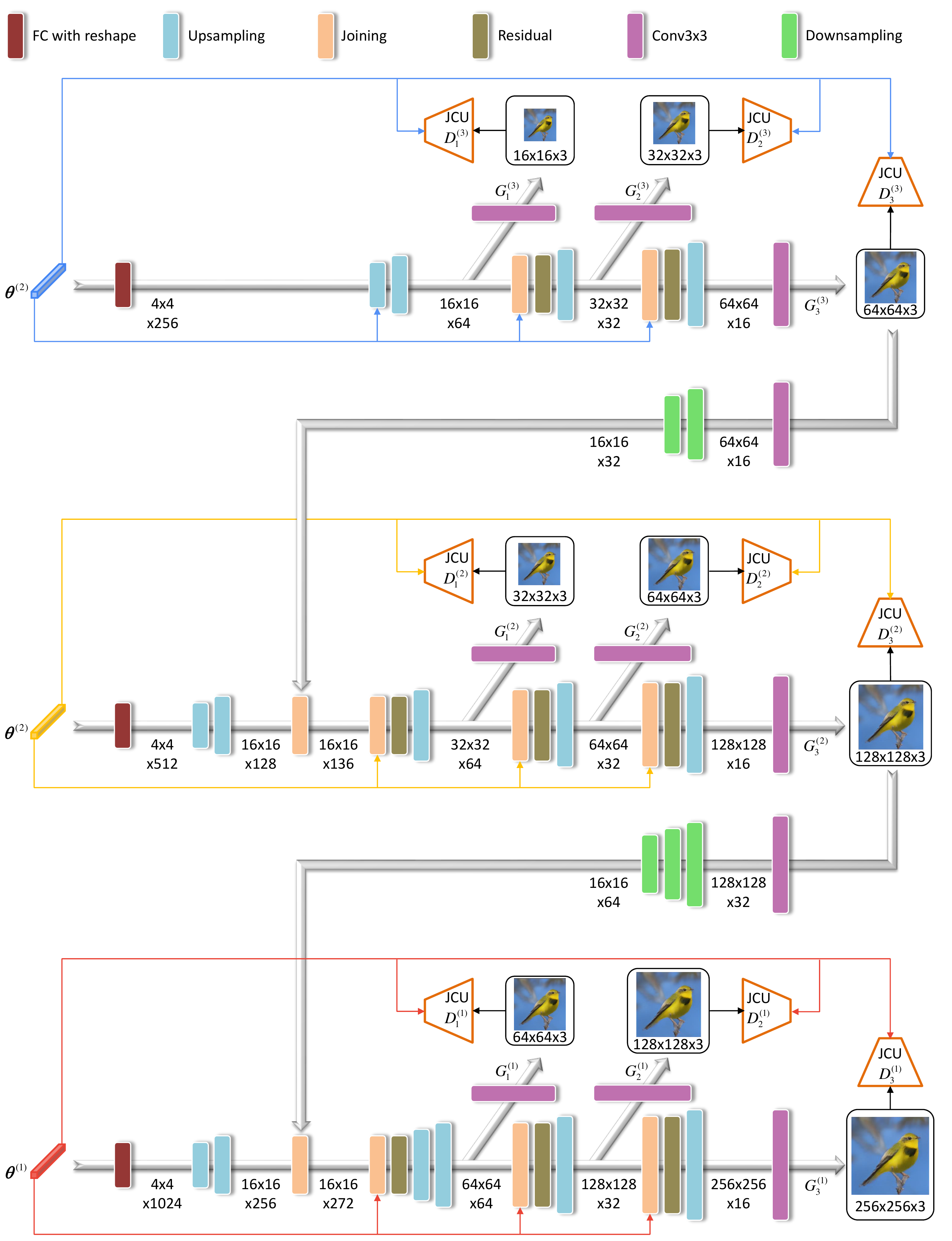}\\
	\caption{\small The structure of  raster-scan-GAN in VHE-raster-scan-GAN, where JCU is a type of discriminator proposed in \citet{Zhang2017StackGAN++}.}\label{Fig: appendix_multi-stackGAN}
\end{figure}

In Fig. \ref{Fig: appendix_multi-stackGAN}, we provide a detailed plot of the structure of the proposed raster-scan-GAN.

\clearpage

\section{Joint optimization for VHE-raster-scan-GAN}
Based on the loss function of VHE-raster-scan-GAN \eqref{Eq: mvrARGAN}, with TLASGR-MCMC \citep{cong2017deep} and WHAI \citep{Zhang2018WHAI}, we describe in Algorithm \ref{Alg: mvrAEGAN-V2} how to perform mini-batch based joint update of all model parameters.

\begin{algorithm}[ht]
	\caption{\small Hybrid TLASGR-MCMC/VHE inference algorithm for VHE-raster-scan-GAN.}
	\begin{algorithmic}
		
		\STATE Initialize encoder parameters $\Omegamat$, topic parameters of PGBN $ \{ \Phimat^{(l)} \}_{1,L}$, generator $G$, and discriminator $D$.
		
		\FOR{$iter = 1,2, \cdots$ }
		\STATE
		Randomly select a mini-batch containing $N$ image-text pairs
		$\dv = \{ \xv_n, \tv_n \}_{n=1}^{N}$;
		
		Draw random noise $\left\{ {{\varepsilonv _n^{(l)}}} \right\}_{n=1,l=1}^{N,L}$ from uniform distribution;
		
		Calculate $ {\nabla _{D }}\mathcal{L}\left( D, G, \Omegamat\, |\, \xv \right)$;
		
		Calculate $ {\nabla _{G }}\mathcal{L}\left( D, G, \Omegamat\, | \,\xv \right)$;
		
		Calculate $ \nabla _{\Omegamat }L$ by the aid of  $\left\{ {{\varepsilonv _n^{(l)}}} \right\}_{n=1,l=1}^{N,L}$;
		
		Update $D$ as $D = D + {\nabla _{D }}\mathcal{L}\left( D, G, \Omegamat \, |\, \xv \right)$;
		
		Update $G$ as $G=G-{\nabla _{G }}\mathcal{L}\left( D, G, \Omegamat \, |\, \xv \right)$;
		
		Update $\Omegamat$ as $\Omegamat=\Omegamat-\nabla _{\Omegamat }L$;
		
		Sample $\{\thetav _n^{( l )}\}_{l=1}^L$ from \eqref{Eq: posterior-completely} given $\Omegamat$ and $ \{ \Phimat^{(l)} \}_{l=1}^{L}$, and use $\{\tv\}_{n=1}^{N}$ to update topics $ \{ \Phimat^{(l)} \}_{l=1}^{L}$ according to TLASGR-MCMC;
		\ENDFOR
		
	\end{algorithmic}\label{Alg: mvrAEGAN-V2}
\end{algorithm}

\section{Data description on CUB, Flower, and COCO with training details}
In image-text multi-modality learning, CUB \citep{WahCUB_200_2011}, Flower \citep{nilsback2008automated} and COCO \citep{lin2014microsoft} are widely used datasets.

{\bf{CUB (\url{http://www.vision.caltech.edu/visipedia/CUB-200-2011.html})}}: CUB contains 200 bird species
with 11,788 images. Since 80$\%$ of birds in this dataset have object-image size ratios of less than 0.5 \citep{WahCUB_200_2011}, as a preprocessing step, we crop all images to ensure that bounding boxes of birds have greater-than-0.75 object-image size ratios, which is the same with all related work.
For textual description, \citet{WahCUB_200_2011} provide ten sentences for each image and we collect them together to form BoW vectors.
Besides, for each species, \citet{elhoseiny2017write} provide its encyclopedia document for text-based ZSL, which is also used in our text-based ZSL experiments.

For CUB, there are two split settings: the hard one and the easy one. The hard one ensures that
the bird subspecies belonging to the same super-category should belong to either the training split or test one without overlapping, referred to as CUB-hard (CUB-H in our manuscript).
A recently used split setting \citep{qiao2016less, akata2015evaluation} is super-category split, where for each super-category, except for one
subspecies that is left as unseen, all the other are used for training, referred to as CUB-easy (CUB-E in our manuscript).
For CUB-H, there are 150 species containing 9410 samples for training and 50 species containing 2378 samples for testing.
For CUB-E, there are 150 species containing 8855 samples for training and 50 species containing 2933 samples to testing.
We use both of them the for the text-based ZSL, and only CUB-E for all the other experiments as usual.

{\bf{Flower \url{http://www.robots.ox.ac.uk/~vgg/data/flowers/102/index.html}}}: Oxford-102, commonly referred to as Flower, contains 8,189 images of flowers from 102
different categories.
For textual description,  \citet{nilsback2008automated} provide ten sentences for each image and we collect them together to form BoW vectors.
Besides, for each species, \citet{elhoseiny2017write} provide its encyclopedia document for text-based ZSL, which is also used in our text-based ZSL experiments in section 4.2.2.
There are 82 species containing 7034 samples for training and 20 species containing 1155 samples for testing.

For text-based ZSL, we follow the same way in  \citet{elhoseiny2017write} to split the data.
Specifically, five random splits are performed, in each of which $4/5$ of the classes are considered as ``seen classes'' for training and $1/5$ of the classes as ``unseen classes'' for testing.
For other experiments, we follow \citet{Zhang2017StackGAN++} to split the data.

{\bf{COCO \url{http://cocodataset.org/#download}}}: Compared with Flower and CUB, COCO is a more challenging dataset, since it contains images with multiple objects and diverse backgrounds.
To show the generalization capability
of the proposed VHE-GANs, we also utilize COCO for evaluation.
Following the standard experimental setup for COCO \citep{reed2016generative,Zhang2017StackGAN++},
we directly use the pre-split training and test sets to train and evaluate our proposed models.
There are 82081 samples for training and 40137 samples for testing.

{\bf{Training details:}} we train VHE-rater-scan-GAN in four Nvidia GeForce RTX2080 TI GPUs.
The experiments are performed with mini-batch size 32 and about 30.2G GPU memory space.
We run 600 epochs to train the models on CUB and Flower,  taking about 797 seconds for CUB-E and 713 seconds for Flower for each epoch.
We run 100 epochs to train the models on COCO, taking about 6315 seconds for each epoch.
We use the Adam optimizer \citep{kingma2014adam} with learning rate $2\times 10^{-4}$, $\beta_1=0.5$, and $\beta_2=0.999$ to optimize the parameters of the GAN generator and discriminator, and use Adam with learning rate  $10^{-4}$, $\beta_1=0.9$, and $\beta_2=0.999$ to optimize the VHE parameters.
The hyper-parameters to update the topics $\Phimat$ with TLASGR-MCMC are the same with those in \citet{cong2017deep}.

\clearpage

{\section{Additional discussion on Obj-GAN}}
Focusing on the COCO dataset, the recently proposed Obj-GAN \citep{li2019object} exploits more side information, including the bounding boxes and labels of objects existing in the images, to perform text-to-image generation.
More specifically, Obj-GAN first trains an attentive 
sequence to sequence model to infer the bounding boxes given a text $\tv$:
\begin{equation}\label{Eq: box generator}
B_{1:T} = \left[ B_1, B_2, \cdots, B_T \right] = G_{\mbox{box}}(\ev),
\end{equation}
where, $\ev$ are the pre-trained bi-LSTM word vectors of $\tv$, $B_t = \left(l_t, b_t \right)$ consists of the class label of the $t$th object and its bounding box $b=(x,y,w,h) \in \mathbb{R}^4$.
Given the bounding boxes $B_{1:T}$, Obj-GAN learn a shape generator to predict the shape of each object in its bounding box:
\begin{equation}\label{Eq: shape generator}
\hat{M}_{1:T} = G_{\mbox{shape}} \left( B_{1:T}, \zv_{1:T}  \right),
\end{equation}
where $\zv_t \sim \mathcal{N} (0,1)$ is a random noise vector.
Having obtained $B_{1:T}$ and $\hat{M}_{1:T}$, Obj-GAN trains an attentive multi-stage image generator to generate the images conditioned on $B_{1:T}$, $\hat{M}_{1:T}$, and $\ev$. 

\begin{figure}[ht]
	\centering
	\includegraphics[scale=0.6]{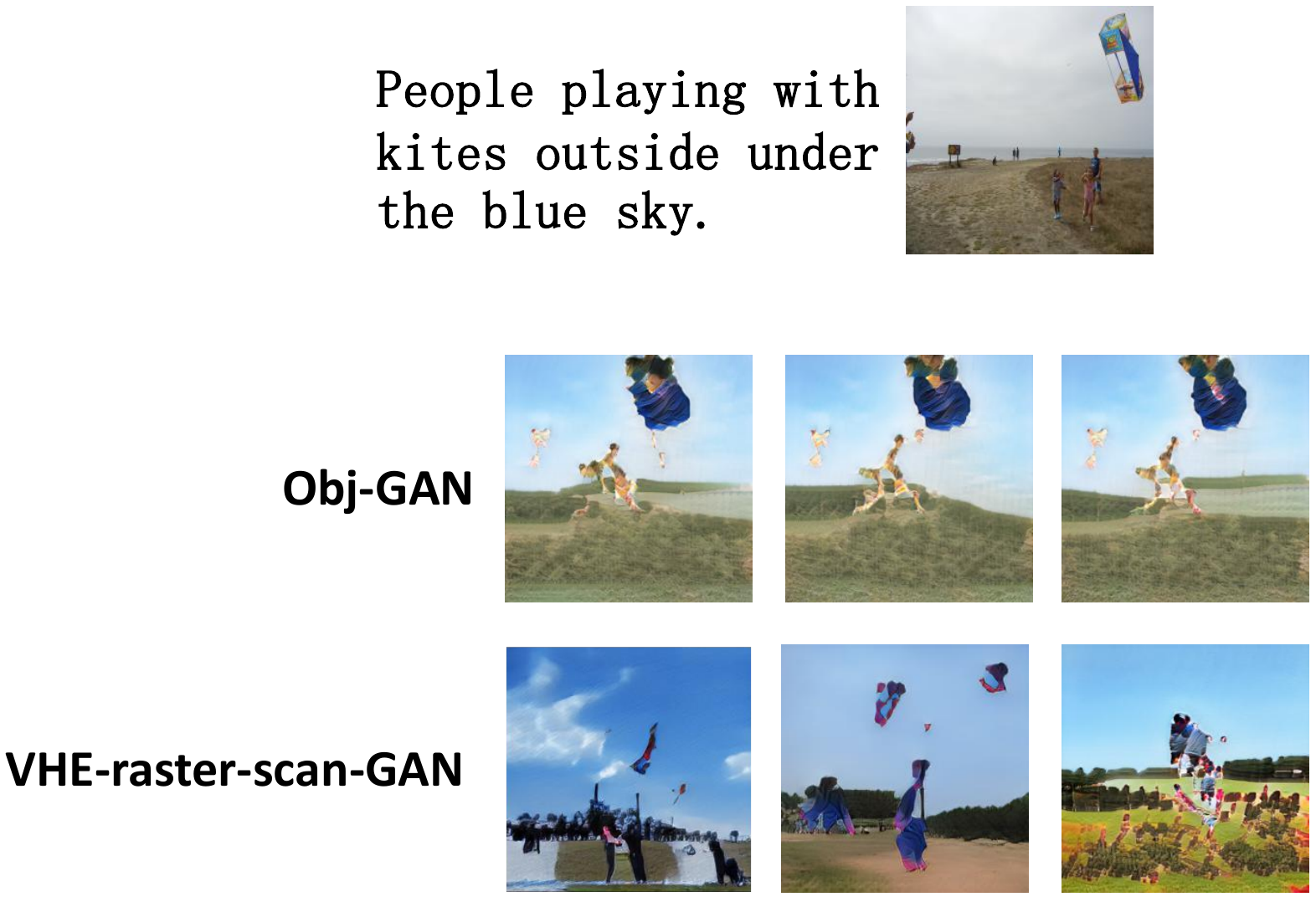}
	\caption{{The generated random images of Obj-GAN given text lack diversity.}} \label{Fig: Obj-GAN}
\end{figure}

Although Obj-GAN achieves a better FID on COCO, it has two major limitations in practice.
First, it is not always possible to obtain accurate bounding boxes and labels of objects in the image;
even they can be acquired by manual labeling, it is often time and labor consuming, especially on large datasets.
Second, 
each word  
is associated with one fixed bounding box; 
in other words, given one sentence, the locations of the objects in the generated images are fixed, which clearly hurts  the diversity of the Obj-GAN generated images, as shown in Fig. \ref{Fig: Obj-GAN}.

\end{document}